\documentclass[letterpaper, 10 pt, conference]{ieeeconf}  

\IEEEoverridecommandlockouts                              

\usepackage{graphics} 
\usepackage{epsfig} 
\usepackage{mathptmx} 
\usepackage{times} 
\usepackage{amsmath} 
\usepackage{amssymb}  
\usepackage{booktabs} 
\usepackage{multirow}
\usepackage{hyperref}
\usepackage{subcaption}

\title{\LARGE \bf
\includegraphics[width=0.6cm]{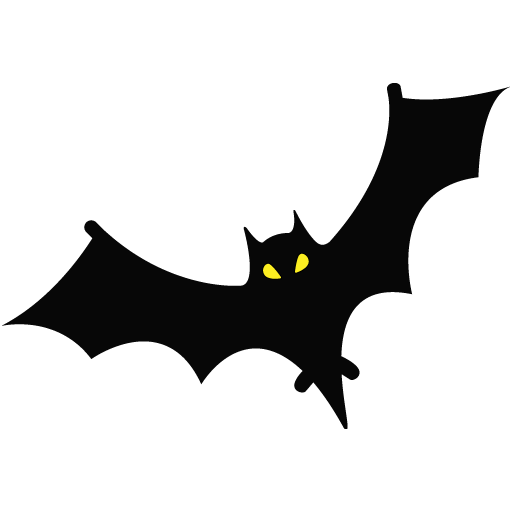}BaTCAVe: Trustworthy Explanations for Robot Behaviors
}

\author{Som Sagar$^{1*}$, Aditya Taparia$^{1*}$, Harsh Mankodiya$^{1}$, Pranav Bidare$^{1}$, Yifan Zhou$^{1}$, and Ransalu Senanayake$^{1}$
\thanks{*Equal contribution}
\thanks{$^{1}$School of Computing and Augmented Intelligence, Arizona State University, Tempe, Arizona
        {\tt\small  <ssagar6, ataparia, hmankodi, pbidare, yzhou298, ransalu>@asu.edu}}%
\thanks{This work has been accepted for publication at the IEEE/RSJ International Conference on Intelligent Robots and Systems (IROS), 2025. ©IEEE}%
}

\begin{document}

\maketitle
\thispagestyle{empty}
\pagestyle{empty}

\begin{abstract}

Black box neural networks are an indispensable part of modern robots. Nevertheless, deploying such high-stakes systems in real-world scenarios poses significant challenges when the stakeholders, such as engineers and legislative bodies, lack insights into the neural networks' decision-making process. Presently, explainable AI is primarily tailored to natural language processing and computer vision, falling short in two critical aspects when applied in robots: grounding in decision-making tasks and the ability to assess trustworthiness of their explanations. In this paper, we introduce a \textit{trustworthy} explainable robotics technique based on human-interpretable, high-level \textit{concepts} that attribute to the decisions made by the neural network. Our proposed technique provides explanations with associated uncertainty scores for the explanation by matching neural network's activations with human-interpretable visualizations. To validate our approach, we conducted a series of experiments with various simulated and real-world robot decision-making models, demonstrating the effectiveness of the proposed approach as a post-hoc, human-friendly robot diagnostic tool. Code: \href{https://github.com/aditya-taparia/BaTCAVe}{https://github.com/aditya-taparia/BaTCAVe} 

\end{abstract}

\section{INTRODUCTION}
\label{sec:intro}

A significant number of models in robotics research are now equipped with deep neural networks (DNNs), with an increasing trend towards end-to-end models. Nevertheless, only a few DNNs are deployed in real-world robots, and those that are deployed mostly focus on tasks such as object detection rather than decision-making or control. This hesitation to use DNNs in real robots is partly due to the high-stakes nature of robot decision-making, where it is unsafe to deploy systems without a clear understanding of their inner workings. Although we do not understand these black-box DNNs, we cannot simply discard them due to their remarkable performance in certain test cases. Hence, we advocate for developing new methods to explain how they work. To this end, instead of focusing on inherently interpretable white-box or gray-box models, this paper focuses on explaining black-box neural networks in robots post-hoc. 

While there are many post-hoc explainable techniques proposed by the machine learning community~\cite{ribeiro2016should, sundararajan2017axiomatic, selvaraju2017grad}, most do not focus on decision-making of a robot or a physical system. For robot decision-making, we want to explain how certain aspects of the input contribute to a particular action or a set of actions. Such explanations help engineers with debugging and legislative bodies with certifying these models. Therefore, to make explanations human-centric, we consider \emph{concepts}---defined as high-level attributes that help humans understand these black boxes~\cite{kim2018interpretability}. As an example, the concept of stripes explains why a DNN would classify an image as a zebra. Concepts~\cite{kim2018interpretability} do not necessarily need to be pixel-level geometric patterns, as in feature attribution methods~\cite{ribeiro2016should, sundararajan2017axiomatic, selvaraju2017grad, NIPS2017_7062}. For instance, in our experiments, we will show how the concept of darkness of an object can be used to explain collision avoidance decisions of a robot.

\begin{figure}[t]
    \centering
    \includegraphics[width=0.48\textwidth]{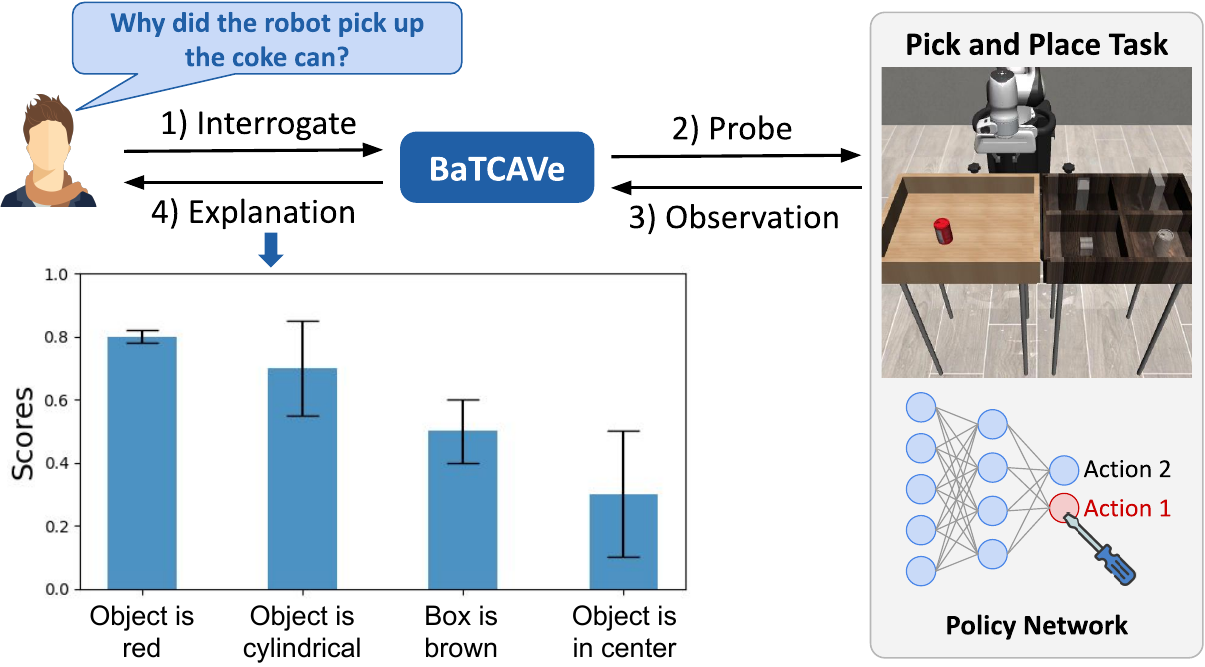}
    \caption{In this pick-and-place task, users can request a post-hoc explanation for \textit{why} the robot succeeded (or failed). BaTCAVe probes the policy network to obtain a ranked list of possible explanations, each with an associated likelihood score. In this example, the object's redness and cylindrical shape are likely contributors to the robot's actions. The uncertainty intervals indicate how much the user should trust each explanation. Explanations help with model debugging, auditing for regulatory compliance, and building trust.}
    \label{fig:intro_figure}
    \vspace{-0.5em}
\end{figure}

If an explainable AI method provides an explanation for why a robot learning algorithm made a particular decision, why should a human trust that explanation? As with any explanation, not only some explanations can be wrong but also there can be multiple explanations for the same decision. Since improving the trustworthiness of a robot explainer is crucial, this paper proposes Bayesian Testing with Concept Activation Vectors (\textbf{BaTCAVe}) to provide explanations accompanied by uncertainty. The key contributions are:
\begin{enumerate}
    \item Proposing a post-hoc explainable decision-making technique for robots equipped with DNNs.
    \item Proposing a framework to evaluate the trustworthiness of explanations using variational inference~\cite{Senanayake2024arxiv_unc}.
    \item Discovering explainable concepts for diverse robotics decision-making tasks---including, manipulation, navigation, and end-to-end driving---with images, language inputs, or proprioception data.
\end{enumerate}

Given that failures are inevitable~\cite{Sagar2024icml, sagarmystery} despite efforts to create robust models, we believe our work in developing explainability tools will assist roboticists in iteratively improving models to enhance their robustness and safety.


\section{RELATED WORK}
\label{sec:related}

\noindent \textbf{Explainability in robotics}: Since robotics is a high-stakes task with many modules that interact with each other, the majority of real-world robots are designed to be interpretable by construction. For instance, they are typically equipped with white-box planning~\cite{fox2017explainable} and control~\cite{bertsekas2019reinforcement} algorithms. Even when models are data-driven, they tend to be constructed as gray-boxes---parameters of an inherently interpretable model are estimated using data~\cite{milani2024interpretable,milani2022survey,bhattacharyya2020online}. As an example, by using a decision tree as the policy, a reinforcement learning algorithm can be made interpretable~\cite{milani2024interpretable}. Unlike these methods, our focus is on developing techniques to probe and explain inherently black-box neural networks in robot decision-making, whether they are used in a modular or an end-to-end fashion. Another line of research has explored explainability of robots for their everyday human users~\cite{silva24,das2021explainable}. In contrast, our focus is on engineers and legislators who aim to audit robot learning models and require distinct, actionable explanations.

\noindent \textbf{Explainable AI}: Unlike inherently interpretable models, the goal of explainable AI (XAI) is to develop methods for explaining black-box models. XAI techniques focus on how to extract explanations as well as how to represent them. Some techniques achieve explainability by testing input components through perturbations~\cite{sundararajan2017axiomatic} or component removal~\cite{NIPS2017_7062}, while others use local approximations of the global decision boundary~\cite{NIPS2017_7062}. XAI methods may leverage gradients~\cite{selvaraju2017grad}, weights~\cite{bau2017network}, or layer-wide insights~\cite{kim2018interpretability} to generate explanations. These explanations can be represented by highlighting specific parts of an image~\cite{selvaraju2017grad}, assigning importance scores to an image segment or pixel clusters~\cite{ribeiro2016should,NIPS2017_7062}, or by analyzing representative samples~\cite{koh2017understanding,kim2018interpretability}. In robot decision-making, highlighting certain parts of an image is not useful as it does not truly explain what aspects of the highlighted pixels is indeed important. For instance, if the XAI technique highlights a car, is it the model or the color that is important? Also, XAI methods typically do not provide uncertainty about their explanations, making them difficult to trust in high-stakes applications such as robotics. Considering the importance of trustworthiness in explanations for robots~\cite{sanneman2020trust}, our approach not only provides explanations but also quantifies their epistemic uncertainty~\cite{Senanayake2024arxiv_unc}.

\section{PRELIMINARIES}
\label{sec:preliminary}

\noindent \textbf{Concepts}: In machine learning, concepts are defined as human-interpretable, high-level attributes such as redness, stripes, young, etc. For instance, the concept of stripes is important for a classifier to identify a zebra~\cite{kim2018interpretability,ghorbani2019towards}. In a classical sense, there are the implicit features a DNN focuses on. 

\noindent \textbf{TCAV}: With the key observation of \textit{similar inputs to a neural network result in similar activation patterns at a given layer}, Testing with Concept activation vectors (TCAV)~\cite{kim2018interpretability} provided the foundation for concept-based explanations. By \textit{projecting} these activations \textit{into a human understandable representation}, humans can explain how the neural networks work in certain aspects. To make such a projection, concept activation vectors (CAVs) can be used. They measure how sensitive a classifier's output is to some activations in the direction of a concept of interest~\cite{kim2018interpretability}. For instance, if the classification of a zebra image is more sensitive to some images of stripes than images of dots (i.e., the direction in the projected space), then stripes play a more significant role in classifying a zebra than dots. However, this classical notion in computer vision cannot be directly applied to robotic decision-making for several reasons. First, robot actions often involve continuous control signals, for which CAVs are not defined. Second, traditional approaches do not provide uncertainty estimates for their explanations. Since an explainer will always return an explanation, we need an indication of whether the explanation should be trusted.


\section{METHOD}
\label{sec:method}



\begin{figure*}[t]
    \centering
    {\includegraphics[width=0.9\textwidth]{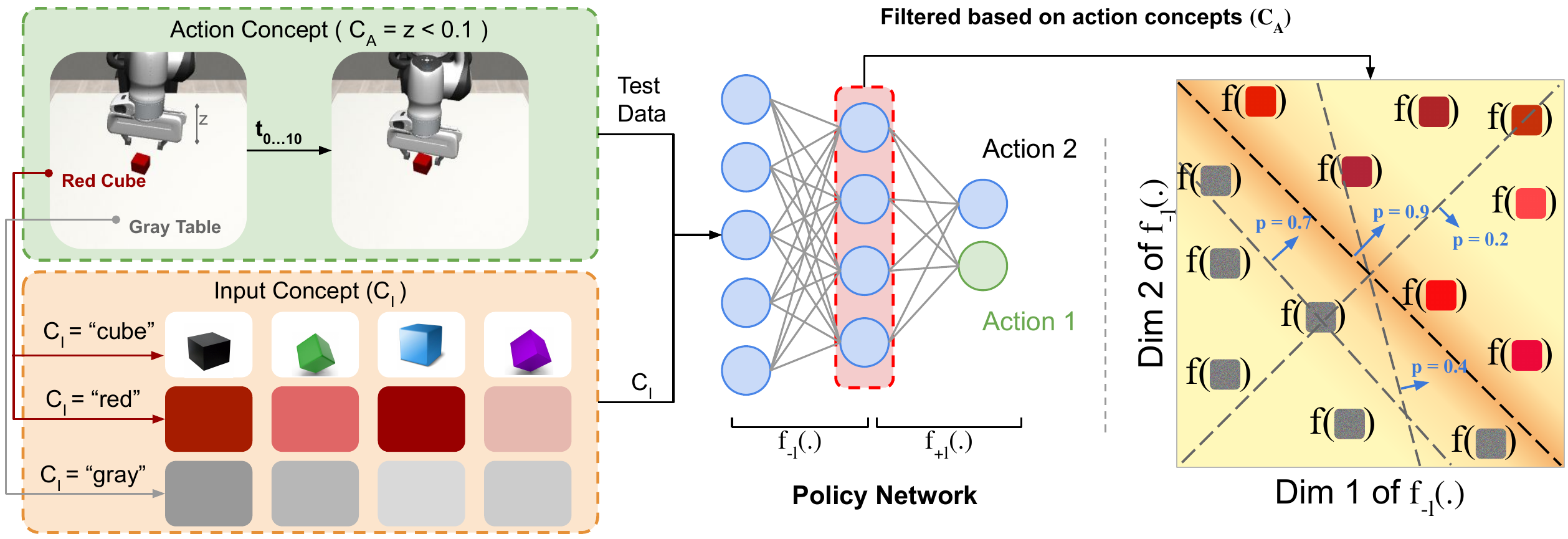}} 
    \caption{The user specifies the robot behavior of interest for analysis using action concepts, $C_A$ (e.g., when the robot is acting near the table). The user also provides input concepts, $C_I$, (e.g., potential explanations represented as a dictionary of images or proprioception data), and test data, $\mathbf{x}$ (e.g., images or proprioception data). BaTCAVe measures the similarity of activation strengths and directions between each set of input concepts and test data for the given behavior. To do this, the input concepts are linearly separated in the activation space using a Bayesian classifier, resulting in infinite number of classification boundaries, each with a different probability.}
    \label{fig:intro}
    \vspace{-0.5em}
\end{figure*}

\subsection{Defining Concepts for Robotic Decision-Making}


\noindent \textbf{Robot's Action Neural Network}: We consider a pre-trained neural network in a robot, $f(\cdot)$, that outputs actions, $\mathbf{y}=f(\mathbf{x})$, with $\mathbf{y} \in \mathbb{R}^D$, for an input $\mathbf{x}$. For instance, for a task on a 7-DOF manipulator, $f(\cdot)$ is an end-to-end neural network-based policy with $\mathbf{x}$ input images and $\mathbf{y} \in \mathbb{R}^7$ control commands to the joints.\\

\noindent \textbf{Action Concepts}: Since a robot takes a sequence of actions in the time horizon $t=0,1,\dots,T$, the entire robot decision-making process cannot be explained at once. Instead, we focus on specific robot behaviors that the user is interested in explaining (Fig.~\ref{fig:intro}). We define an action concept as \emph{a specific target robot behavior for which we seek explanations}. For instance, why does the manipulator grasp a cube, or why does the autonomous vehicle turns right at a particular road segment? We want to extract what input cues---visual or language---led to these behaviors. Is it the color, geometry, or something else? Formally, an action concept is defined as,
\begin{equation}
    C_A = \left\{ t \mid \bigwedge_{d \in D} \text{rule}(\mathbf{y}_t^{(d)}), \quad t=\{0,1,\dots\,T\}  \right\},
\end{equation}
\noindent where logical operations, $\text{rule}(\cdot)$, such as conjunction ($\wedge$), disjunction ($\vee$), and negation ($\lnot$), are applied across all dimensions of the action vector, $d \in D$, for the entire action series from $0$ to $T$. Essentially, we are extracting segments of all trajectories that exhibit a particular behavior. In a manipulation example, a user might be interested in analyzing the behavior, where the end effector moves down-right while the gripper is open. In such a case, by overloading the notation $x,y$ to represent the cardinal direction in the robot's physical space and opened gripper distance, $d$, the action concepts can be defined as the time steps that obey $\left( \frac{\mathrm{d}x}{\mathrm{d}t} >0 \right) \wedge \left( \frac{\mathrm{d}z}{\mathrm{d}t} <0  \right) \wedge \left( d > 0.5 \right)$. Action concepts help us isolate the behaviors that we are interested in explaining.\\

\noindent \textbf{Input Concepts}: An input concept is a high-level, human interpretable attribute of robot inputs that the user believes is important to explain an action concept. These can be textures, colors, sizes, distances, directions, shapes, objects, etc. Input concepts can be defined based on engineers' intuition, prior knowledge, or the test cases an engineer or a legislative body is interested in. Additionally, since input concepts can be automatically extracted~\cite{ghorbani2019towards, fel2023craft, fel2024holistic} or generated~\cite{taparia2024explainableconceptgenerationvisionlanguage} based on inputs, we consider concept discovery as out-of-scope for this paper. While action concepts are defined as rules, an input concept, $C_I$, is defined by a collection of representative inputs, $\{\mathbf{x}_{C_I}\}_{m=1}^M$. For instance, $M$ images of stripes can be used to explain why an autonomous vehicle slowed down near a crosswalk. 



\begin{figure*}[t]
    \centering
    \vspace{0.2em}
    \includegraphics[width=0.98\textwidth]{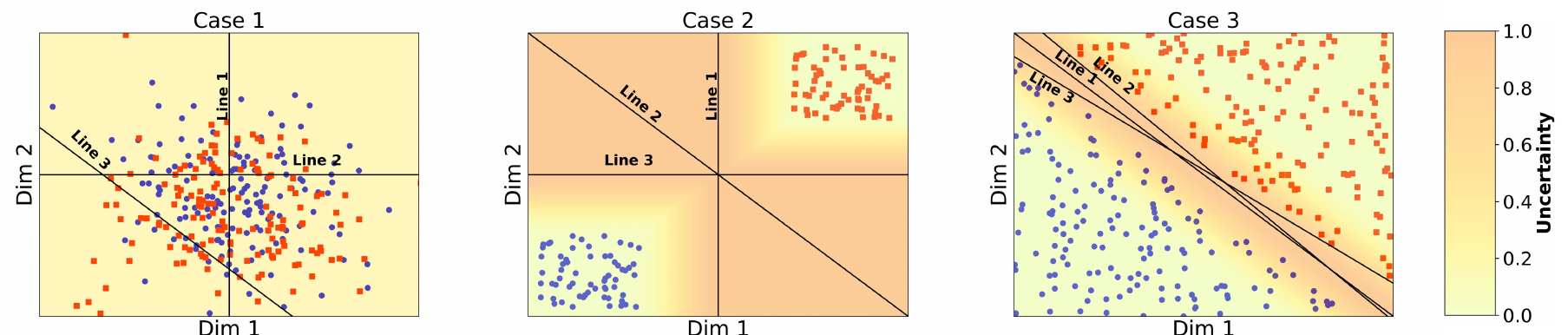}
    \vspace{-0.15em}
    \caption{Data with two classes (red and blue) are represented in the activation space. If the uncertainty is high (case 2 vs. 3), then we can sample many valid lines (i.e., many explanations). Though many lines can be sampled from case 1 as well, since the accuracy is low, the explanations cannot be trusted.}
    \label{fig:explain_cases}
    \vspace{-0.5em}
\end{figure*}

\subsection{Bayesian Testing with CAVs (BaTCAVe)}
\label{sec:btcav}

Given a collection of input concepts for an action concept, we now derive a score to measure how well each input concept explains the action concept. If an autonomous vehicle inadvertently failed to yield to another vehicle (i.e., action concept), we can test if the color or the type of the other vehicle (i.e., input concepts) influenced the decision. To derive a score for an explanation, as shown in Fig.~\ref{fig:intro}, let us decompose the neural network into two segments, $f_{-l}(\cdot)$ and $f_{+l}(\cdot)$, with one following the other, at the $l^\text{th}$ layer as, $\mathbf{y} = f(\mathbf{x}) = f_{+l}(f_{-l}(\mathbf{x}))$ for input $\mathbf{x}$. In other words, $f_{-l}(\cdot)$ are the activations of a DNN at the $l^\mathrm{th}$ layer, which are then fed to the rest of the network, $f_{+l}(\cdot)$, to obtain output $\mathbf{y}$. 

Since our objective is to build a relationship between the activations and human understandable concept inputs for some behaviors (i.e., action concepts), as shown in the plot of Fig.~\ref{fig:intro}, we work in the space of activations (i.e., the space of $f_{-l}(\cdot)$). The dimensionality of the activation space is equal to the number of nodes in the $l^\text{th}$ layer. We can obtain the sensitivity of neural network decisions, conditioned by $C_A$, w.r.t. layer activations as $\frac{\partial f^{C_A}(\mathbf{x})}{\partial f_{-l}(\mathbf{x})}$. If we want to measure this sensitivity along a particular direction, $\mathbf{v}$, in the activation space, we have to compute the \emph{directional derivative} $\frac{\partial f^{C_A}(\mathbf{x})}{\partial f_{-l}(\mathbf{x})} \cdot \mathbf{v}$. If this direction is a random variable, $V$, with its \textit{realizations}, $\mathbf{v}$, is given by the stochastic directional derivative,
\begin{equation}
\begin{split}
\mathbb{E}_V \left[ \frac{\partial f^{C_A}(\mathbf{x})}{\partial f_{-l}(\mathbf{x})} \cdot \mathbf{v} \right] = \int \frac{\partial f^{C_A}(\mathbf{x})}{\partial f_{-l}(\mathbf{x})} \cdot \mathbf{v} \cdot p(\mathbf{v}) \, \mathrm{d}\mathbf{v}.
\end{split}
\label{eq:dirdev}
\end{equation}

If we build a relationships between $V$ and $C_I$, we measure the sensitivity of the neural network decisions to human understandable input concepts. Since $V$ is a random variable, it also tells us about the uncertainty of this relationship. To build this relationship, let us estimate $\mathbf{v}$ using an external dataset containing inputs from $C_I$. To this end, similar to ~\cite{kim2018interpretability}, we collect a concept dataset: $X^{C_I} = \{(\mathbf{x}_m^{C_I},z^+)\}_{m=1}^M$ with input concepts and a different set of inputs $X^{C_I^-}=\{(\mathbf{x}_m^{C_I^-},z^-)\}_{m=1}^M$ with positive and negative classes labeled as $z^+$ and $z^-$, respectively. The negative input concept, $C_I^-$, can be just another concept or a random collection of inputs, depending on what the user is interested in comparing. How these sets are selected will be described in the Experiments section. See Fig.~\ref{fig:intro} for an illustration.

By computing $f_{-l}(\mathbf{x}_m^{C_I})$ and $f_{-l}(\mathbf{x}_m^{C_I^-})$, we obtain the activation values of the concept dataset. If we can obtain a linear separation between the two classes, then the concept is more unique in the activation space. However, as we will discuss in Section~\ref{sec:decisions}, separation in a high dimensional activation is subjected to uncertainty, we consider a probabilistic linear separation (i.e., we can draw many separation lines with different probabilities). By setting this probabilistic linear separation the same as $\mathbf{v}$, we obtain the relationship between $V$ and $C_I$. To estimate the vector distribution, $\mathbf{v}$, that separates the two classes in the activation space, we apply the Bayes theorem,
\begin{equation}
\begin{split}
    \underbrace{p(\mathbf{v}|z,X^{C_I},X^{C_I^-})}_{\text{posterior}} 
    &\propto \underbrace{p(z|\mathbf{v},X^{C_I},X^{C_I^-})}_{\text{likelihood}} \times \underbrace{p(\mathbf{v})}_{\text{prior}}.
\end{split}
\label{eq:bay}
\end{equation}

Given the positive and negative labels, the likelihood follows a Bernoulli distribution, $z|\mathbf{v} \sim \mathrm{Bernoulli}$. The prior weight distribution is considered to follow a normal distribution, $\mathbf{v} \sim \mathcal{N}$. However, because of the non-conjugate prior, the posterior distribution is not tractable~\cite{Senanayake2024arxiv_unc,bishop2006pattern}. Hence, we resort to approximate Bayesian inference. Because the activation space of the neural network can be high dimensional, rather than using an MCMC technique, we use a much faster and scalable method---variational inference~\cite{bishop2006pattern}---where we minimize the KL divergence between the true posterior and an approximate posterior distribution, $q(\mathbf{v})$. However, since the true posterior is not known, instead of minimizing the KL divergence, we maximize an evidence lower bound (ELBO),
\begin{equation}
    \mathrm{ELBO} + \mathbb{KL}[q(\mathbf{v}) || p(\mathbf{v}|z,X^{C_I},X^{C_I^-})] = \mathrm{const}.
\end{equation}
Following the locally linear approximation of the posterior in \cite{pmlr-vR1-jaakkola97a}, we learn $q(\mathbf{v})$ in an expectation-maximization-style. 

On a collection of test inputs, $X^{\text{test}}=\{\mathbf{x}_p^\mathrm{test}\}_{p=1}^P$, with relation to (\ref{eq:dirdev}), we can now compute a score, $s^{C_A, C_I, C_I^-}$,
\begin{equation}
    \begin{split}
    \frac{1}{|X^\text{test}|} \sum_{\mathbf{x}^\text{test} \in X^\text{test}} 
    &\mathbb{I} \Bigg( \Bigg( \int \frac{\partial f^{C_A}(\mathbf{x}^\text{test})}{\partial f_{-l}(\mathbf{x}^\text{test})} \cdot \mathbf{v} \cdot q(\mathbf{v}) \, \mathrm{d}\mathbf{v} \Bigg) > 0 \Bigg)
    \end{split}
    \label{eq:score}
\end{equation}
This score itself is a distribution. If we obtain samples from $q(\mathbf{v})$, each of them is a valid explanation. Since some samples are more probable than others, those with high probability are more likely concepts that explanations the decisions. The empirical mean and standard deviation of the score can be estimated easily. The mean score $\bar{s}$ is given by,
\begin{equation}
    \begin{split}
     \frac{1}{R \cdot |X^\text{test}|} \sum_{\mathbf{x}^\text{test} \in X^\text{test}} \sum_{\mathbf{v}_r \in V^{C_I}} 
    &\mathbb{I} \left( \frac{\partial f^{C_A}(\mathbf{x}^\text{test})}{\partial f_{-l}^{C_A}(\mathbf{x}^\text{test})} \cdot \mathbf{v}_r > 0 \right)
    \end{split}
    \label{eq:s_mean}
\end{equation}
where $V^{C_I} = \{\mathbf{v}_r\}_{r=1}^R$ are $R$ samples taken from $q(\mathbf{v})$. 

\noindent \textbf{Interpreting the score}: The higher the score, the better the concept $C_I$ explains the test inputs $X^\text{test}$ or the robot behavior.

\noindent \textbf{Interpreting the uncertainty}: Similarly, the empirical standard deviation reflects the epistemic nature~\cite{Senanayake2024arxiv_unc} of explanations---how much the model knows that its explanation can be wrong. Let us now discuss different variations of uncertainty. Note that, since we considered the full distribution over parameters, the uncertainty is calibrated by construction (our distributions on parameters are on a generalized linear model of the activation space, not on the neural network)~\cite{gelman1995bayesian}.

\begin{figure*}[t]
    \centering
    \vspace{0.2em}
    \includegraphics[width=0.97\textwidth]{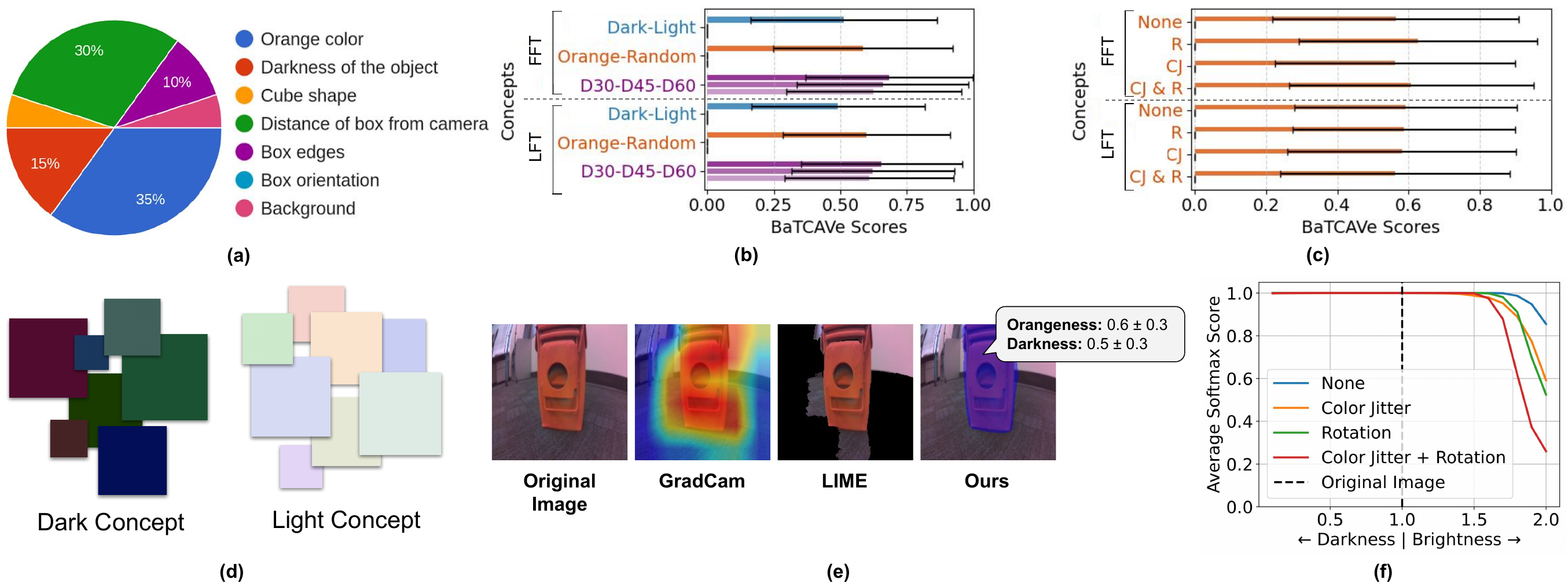}
    \vspace{-0.2em}
    \caption{(a) Shows the distribution of input concepts $C_I$ selected by the participants for object avoidance task. (b) and (c) shows the effects of fine-tuning and data augmentation, respectively. The higher the score, the better the explanation is. (d) shows samples of dark and light concepts. (e) shows while common XAI methods such as GradCam~\cite{selvaraju2017grad} and LIME~\cite{ribeiro2016should} can highlight the orange box, they do not reveal what attributes (i.e., input concepts) of the box contribute to the decision of the DNN, making it harder for the engineers to improve the DNN based on the explanations. In contrast, BaTCAVe provides semantically meaningful explanations. (f) highlights the change in confidence over modifying darkness factor in input with models trained with different data augmentation (C-Modification).}
    \label{fig:exp4.1.1_summ}
\end{figure*}

\begin{figure*}[t]
    \centering
    \includegraphics[width=0.23\textwidth]{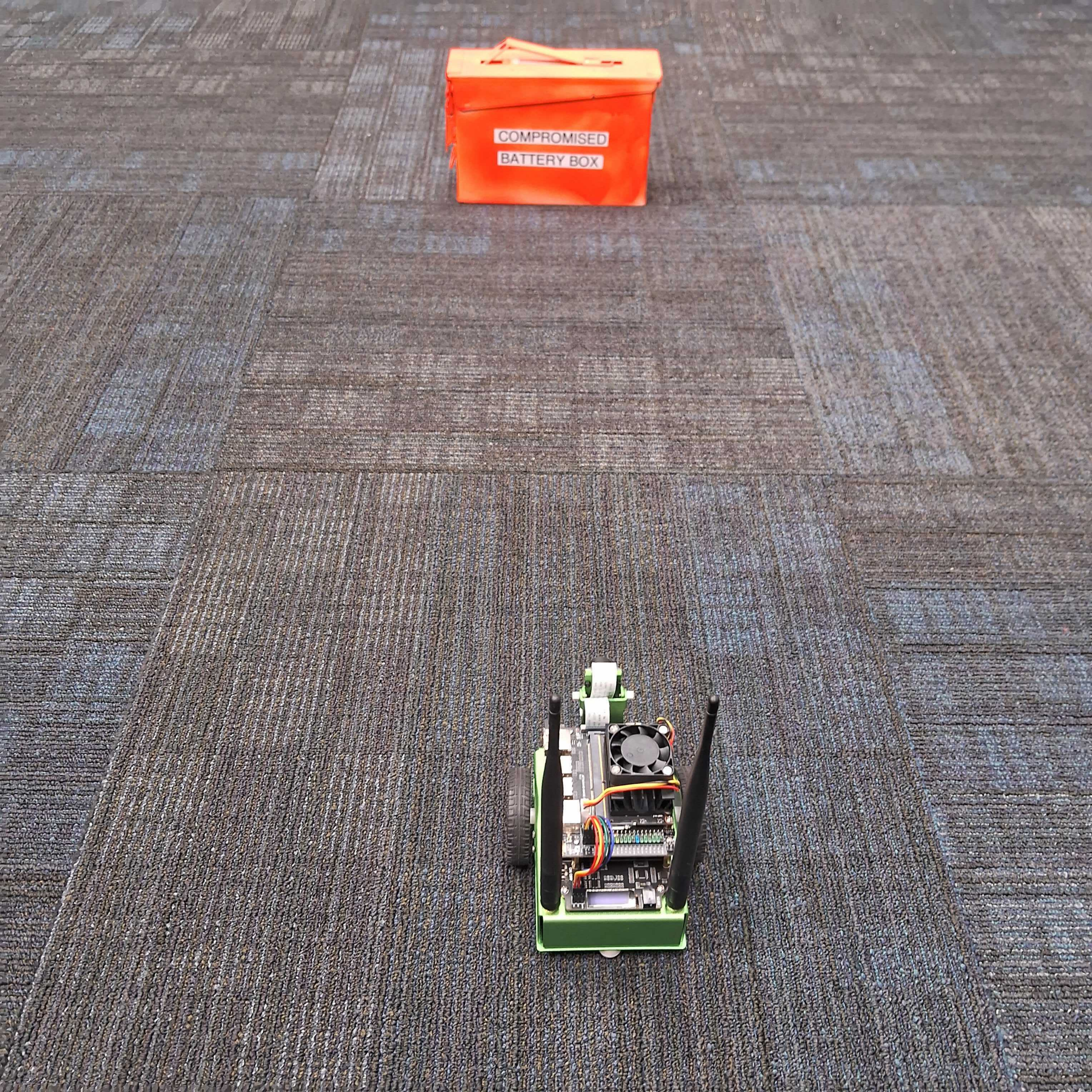}
    \includegraphics[width=0.23\textwidth]{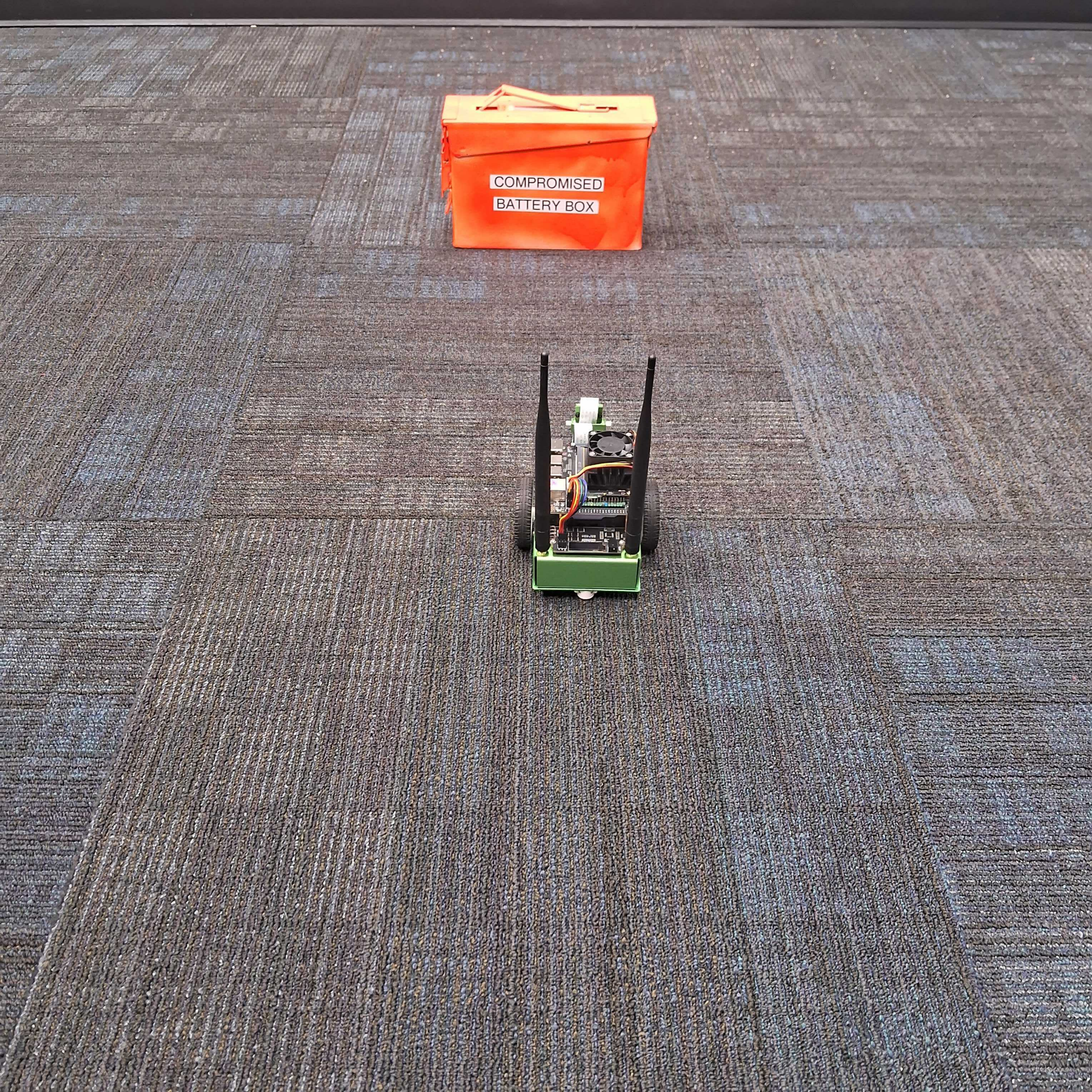}
    \includegraphics[width=0.23\textwidth]{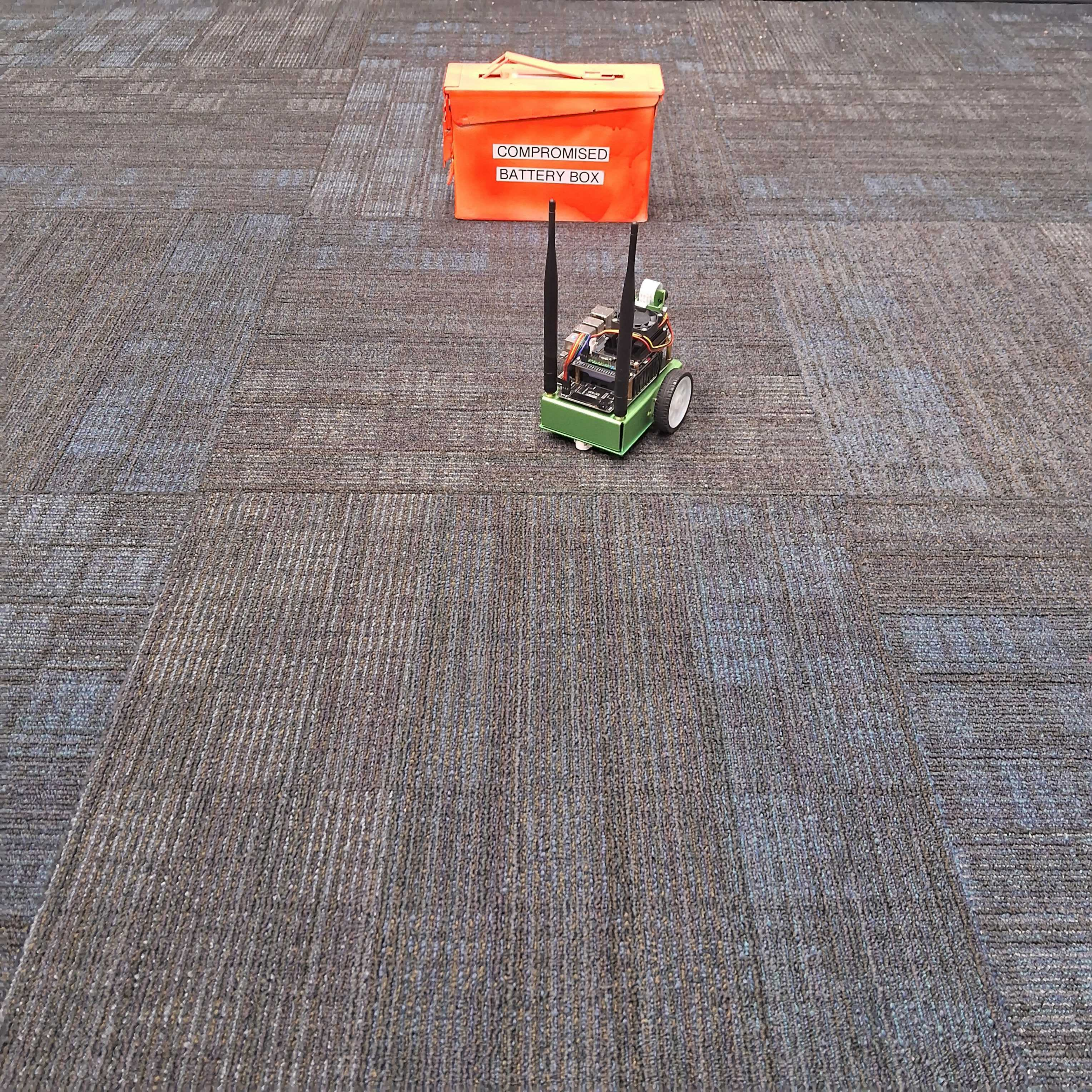}
    \includegraphics[width=0.23\textwidth]{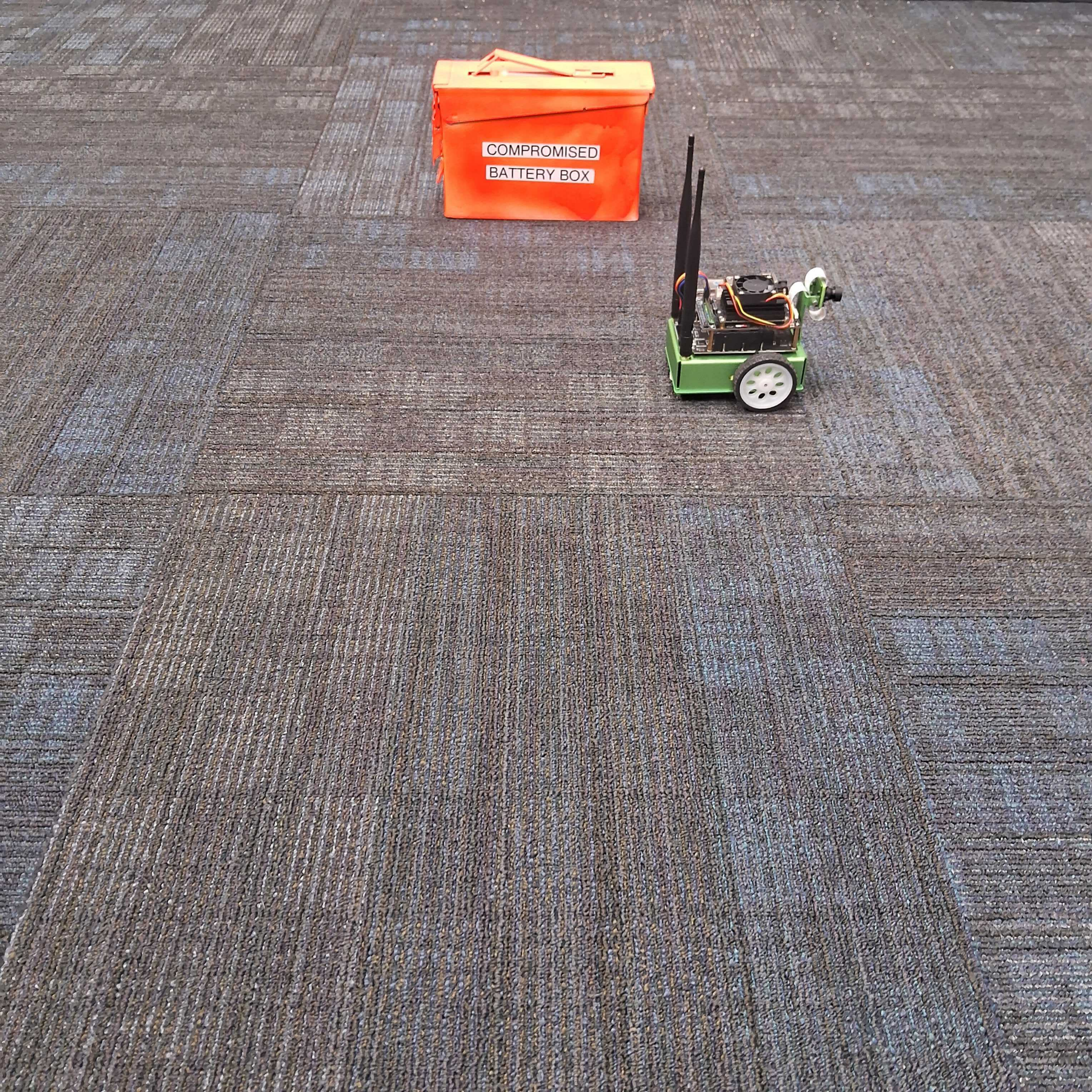}
    \caption{A JetBot rollout. We investigate what attributes (i.e., concepts) of the obstacle influenced the decision to turn.}
    \label{fig:experiments}
    \vspace{-0.5em}
\end{figure*}

\subsection{Interpreting the Trustworthiness of Explanations}
\label{sec:decisions}

To assess the trustworthiness of explanations, relying on 1) the held-out test accuracy from eq. (\ref{eq:bay}) and 2) the epistemic uncertainty of the score in eq. (\ref{eq:score}), we consider (Fig.~\ref{fig:explain_cases}):\\
{\bf Case 1} (Off-base explanations): If the accuracy is low, the concepts are not good enough to delineate the CAVs, resulting in an inaccurate explanation. Formally, in such cases, $p^{C_I} \approx p^{C_I^-}$, where the distributions are defined as $X^{C_I} \sim p^{C_I}$ and $X^{C_I^-} \sim p^{C_I^-}$ for the positive and negative classes, respectively. Such explanations should not be trusted. \\
{\bf Case 2} (Imprecise explanations): If the accuracy is high but the uncertainty is also high, multiple explanations are possible. This can be due to, 1) the two supposedly opposite concepts are not sufficiently different enough to be delineated with a low uncertainty (i.e., some conflicting information) or, most likely, 2) the test samples lack diversity in the activation space. Unfortunately, Bayesian inference cannot differentiate lack of information from conflicting information~\cite{klir1990uncertainty}.\\
{\bf Case 3} (Precise explanations): If the accuracy is high and the uncertainty is low, then the concepts we have chosen are good and the explainer is able to provide consistently good explanations. These explanations are highly trustworthy. 

The thresholds for probability should be decided by the practitioners depending on how much risk they are willing to take. For instance, if an engineer is using BaTCAVe to debug a manipulator used in an assembly line, the threshold can be selected leniently as the stakes might be relatively low. In such cases, we can obtain more valid explanations. In contrast, if a legislative body is using BaTCAVe to approve a new autonomous vehicle, then the thresholds should be strict. 
If case 1 violates, we should try new concepts to obtain a better accuracy. If case 2 violates, we can still use explanations but they might not always be the best explanations. By obtaining the mean score, we can obtain an average explanation. 


\section{Experiments}
\label{sec:result}

We consider three types of algorithms for training robot decision-making: binary supervised learning, behavioral cloning (BC), and proximal policy optimization (PPO). The experimental setup, results, and findings are detailed below.


\begin{table*}[ht]
    \centering
    \small
    \caption{BaTCAVe scores $(\uparrow)$ across different tasks, quantifying the relevance of each action concept to the task, along with uncertainty estimates. Each score measures the impact of input concepts such as the object's features, end-effector position (eef\_pos), end-effector orientation (eef\_quat in quaternion format), and gripper on task performance. (*Not Applicable)}
    \begin{tabular}{ccccccp{3.5cm}}
        \toprule
        \textbf{Tasks} & \textbf{Action Concepts} & \multicolumn{4}{c}{\textbf{Input Concepts}} & \textbf{Conclusions}\\
         & & \textbf{object} & \textbf{eef\_pos} & \textbf{eef\_quat} & \textbf{gripper}  & \\
        \midrule
        

        \multirow{7}{*}{\parbox{2.5cm}{\centering \textbf{Pick \& Place} \\ \vspace{2mm} \includegraphics[width=2cm]{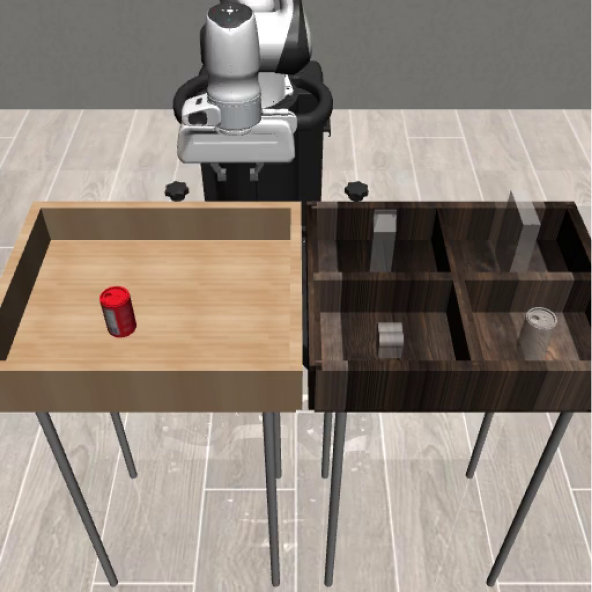}}} 
        
        & $\Delta$ X & $\mathbf{1.00 \pm 0.00}$ & $0.54 \pm 0.49$ & $\mathbf{0.96 \pm 0.16}$ & $0.09 \pm 0.09$ & \multirow{7}{3.5cm}{Object features are most important, with $\Delta X$ (1.00) and $\Delta Z$ (0.90) scoring highest. End-effector orientation ($0.96$) aids alignment and gripper is relevant for $\Delta Z$ ($0.91$).} \\
        & $\Delta$ Y  & $0.00 \pm 0.00$ & $0.7 \pm 0.41$ & $0.24 \pm 0.43$ & $0.00 \pm 0.00$ \\
        & $\Delta$ Z  & $\mathbf{0.90 \pm 0.17}$ & $0.85 \pm 0.35$ & $0.77 \pm 0.41$ & $\mathbf{0.91 \pm 0.27}$ \\
        & $\Delta$ Roll & $0.44 \pm 0.49$ & $0.53 \pm 0.49$ & $0.43 \pm 0.49$ & $0.49 \pm 0.49$ \\
        & $\Delta$ Pitch & $0.86 \pm 0.33$ & $0.55 \pm 0.49$ & $0.45 \pm 0.49$ & $0.46 \pm 0.49$ \\
        & $\Delta$ Yaw  & $0.84 \pm 0.35$ & $0.51 \pm 0.49$ & $0.50 \pm 0.49$ & $0.58 \pm 0.49$ \\
        & Gripper & $0.10 \pm 0.00$ & $0.64 \pm 0.46$ & $0.56 \pm 0.48$ & N.A* \\
        
        \midrule

        \multirow{7}{*}{\parbox{2.5cm}{\centering \textbf{Lift} \\ \vspace{2mm} \includegraphics[width=2cm]{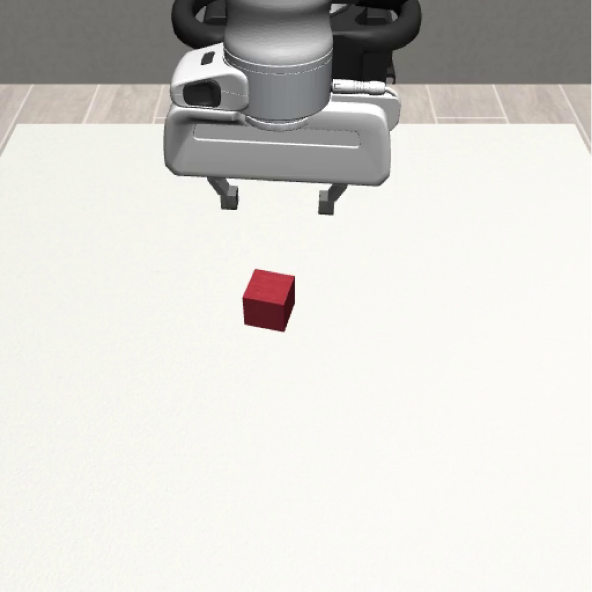}}}

        & $\Delta$ X  & $ 0.84 \pm 0.31$ & $0.42 \pm 0.40$ & $0.42 \pm 0.49$ & $0.42 \pm 0.49$ & \multirow{7}{3.5cm}{End-effector positioning is critical, with $\Delta Y$ ($0.99$) and $\Delta Z$ ($0.99$) dominating. The gripper's role ($0.95$) suggests a strong grasp is essential.} \\
        & $\Delta$ Y  & $0.85 \pm 0.32$ & $\mathbf{0.99 \pm 0.07}$ & $0.56 \pm 0.49$ & $0.49 \pm 0.40$ \\
        & $\Delta$ Z  & $0.98 \pm 0.10$ & $\mathbf{0.99 \pm 0.04}$ & $0.42 \pm 0.49$ & $0.39 \pm 0.48$ \\
        & $\Delta$ Roll  & $0.46 \pm 0.49$ & $0.41 \pm 0.49$ & $0.48 \pm 0.48$ & $0.51 \pm 0.49$ \\
        & $\Delta$ Pitch  & $0.54 \pm 0.49$ & $0.28 \pm 0.73$ & $0.43 \pm 0.47$ & $0.48 \pm 0.49$ \\
        & $\Delta$ Yaw  & $0.73 \pm 0.39$ & $0.28 \pm 0.45$ & $0.47 \pm 0.49$ & $0.48 \pm 0.49$ \\
        & Gripper & $\mathbf{0.95 \pm 0.18}$ & $0.004 \pm 0.06$ & $0.31 \pm 0.39$ & N.A* \\

        \midrule
        
        \multirow{7}{*}{\parbox{2.5cm}{\centering \textbf{Nut Assembly} \\ \vspace{2mm} \includegraphics[width=2cm]{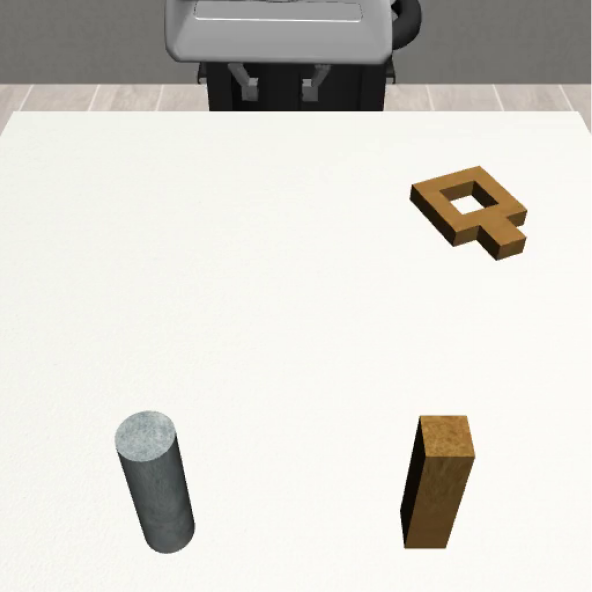}}} 
        
        & $\Delta$ X & $0.64 \pm 0.47$ & $0.59 \pm 0.49$ & $0.44 \pm 0.49$ & $\mathbf{0.87 \pm 0.32}$ & \multirow{7}{3.5cm}{Gripper control is key, influencing $\Delta X$ ($0.87$) and $\Delta Z$ ($0.85$). Object and end-effector positioning contribute moderately, indicating precise handling.}  \\
        & $\Delta$ Y  & $\mathbf{0.87 \pm 0.32}$ & $0.59 \pm 0.48$ & $0.46 \pm 0.49$ & $0.37 \pm 0.48$ \\
        & $\Delta$ Z  & $0.01 \pm 0.07$ & $0.54 \pm 0.49$ & $0.51 \pm 0.49$ & $\mathbf{0.85 \pm 0.35}$ \\
        & $\Delta$ Roll & $0.52 \pm 0.49$ & $0.48 \pm 0.49$ & $0.48 \pm 0.49$ & $0.45 \pm 0.49$ \\
        & $\Delta$ Pitch & $0.66 \pm 0.46$ & $0.51 \pm 0.49$ & $0.45 \pm 0.49$ & $0.56 \pm 0.49$ \\
        & $\Delta$ Yaw & $0.58 \pm 0.49$ & $0.44 \pm 0.49$ & $0.45 \pm 0.49$ & $0.52 \pm 0.49$ \\
        & Gripper & $0.10 \pm 0.00$ & $0.88 \pm 0.31$ & $0.53 \pm 0.47$ & N.A* \\
        \bottomrule
    \end{tabular}
    \label{table:com_val}
    \vspace{-1.2em}
\end{table*}

\subsection{Experiment 1: Vision-based Mobile Robot Navigation}
\label{sec:exp1}

\emph{Setup}: As a real-world platform, we used a JetBot with a NVIDIA Jetson Nano (Fig~\ref{fig:experiments}). It uses a pretrained AlexNet~\cite{krizhevsky2012imagenet} with the last layer replaced with two nodes. Using 175 ``obstacle'' and 175 ``free'' images, we fine-tune the DNN so that the robot can learn to avoid obstacles. With the objective of avoiding orange obstacles, we fine-tuned the full DNN with images of an orange box, shown in Fig.~\ref{fig:experiments}, until we achieve a validation accuracy of 100\%. We analyze how the decision of the network is affected by different data pre-processing and fine-tuning techniques for $C_A=\{\text{avoid obstacle}\}$ and $C_I=\{\text{orangeness, darkness, distance}\}$. 


\emph{Results}: 

\emph{Q1. Are explanations useful to the engineers?}
We surveyed 20 engineering students who have at least 1 year of experience training DNNs to get their opinion on what the DNN has learned just by reading our description. We described them the architecture and showed fine-tuning images without telling them that our objective is avoiding orange obstacles. 

First, we asked participants to describe what attributes the DNN should have learned and then we gave them a list of potential concepts to narrow down their choices. As shown in Figs.~\ref{fig:exp4.1.1_summ}(a), many participants (35\%) speculated that the model would have learned to distinguish orange from the rest. In contrast, showcasing the BaTCAVe's ability to provide true explanations that engineers might not even think of, as shown in Fig.~\ref{fig:exp4.1.1_summ}(b), BaTCAVe revealed that the model had learned to distinguish dark from light  objects (or the {\emph{value} or brightness in HSV scale). The orange box is merely a shade of the broader ``dark'' concept. Thus, we conclude that BaTCAVe is capable of providing explanations of inner understanding of the neural network, which, without explanations, is not intuitive to engineers.

\emph{Q2. Are BaTCAVe explanations correct?}
Similar to C-deletion in XAI literature~\cite{ghorbani2019towards, fel2023craft, fel2024holistic}, we verified that the darkness is an important concept by gradually varying the brightness of the orange object, as shown in Fig.~\ref{fig:exp4.1.1_summ}(f). Based on Fig.~\ref{fig:exp4.1.1_summ}, BaTCAVe made the following explanations:
\begin{enumerate}
    \item The score is proportional to the concept of ``distance between the robot and obstacle,'' verifying that the DNN has learned the distance.
    \item When the final layer is fine-tuned (LFT) instead of the full DNN (FFT), the prominence of the orange concept becomes higher compared to the dark concept, which is what we originally intended, demonstrating how engineers can use concepts to debug robots.
    \item As shown in Fig~\ref{fig:exp4.1.1_summ}(c), the importance of the dark concept remains consistent across different data augmentation methods---adding color jitter (CJ) and/or image rotation (R)---for LFT while it varies for FFT.
\end{enumerate}

Since XAI methods are hard to quantitatively benchmark, similar to other work, Fig.~\ref{fig:exp4.1.1_summ}(e) qualitatively compares why BaTCAVe is a better choice than feature attribution methods for robot learning.

\begin{figure}[h]
    \centering
    \includegraphics[width=0.45\textwidth]{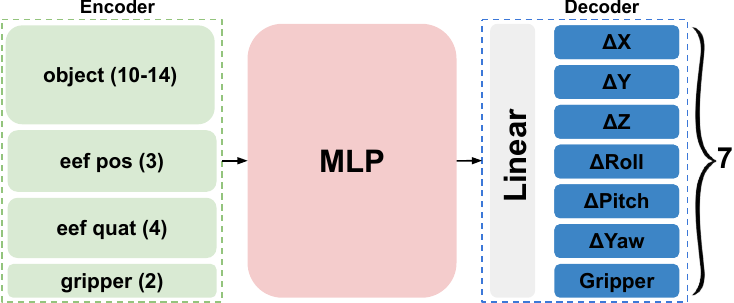}
    \includegraphics[width=0.45\textwidth]{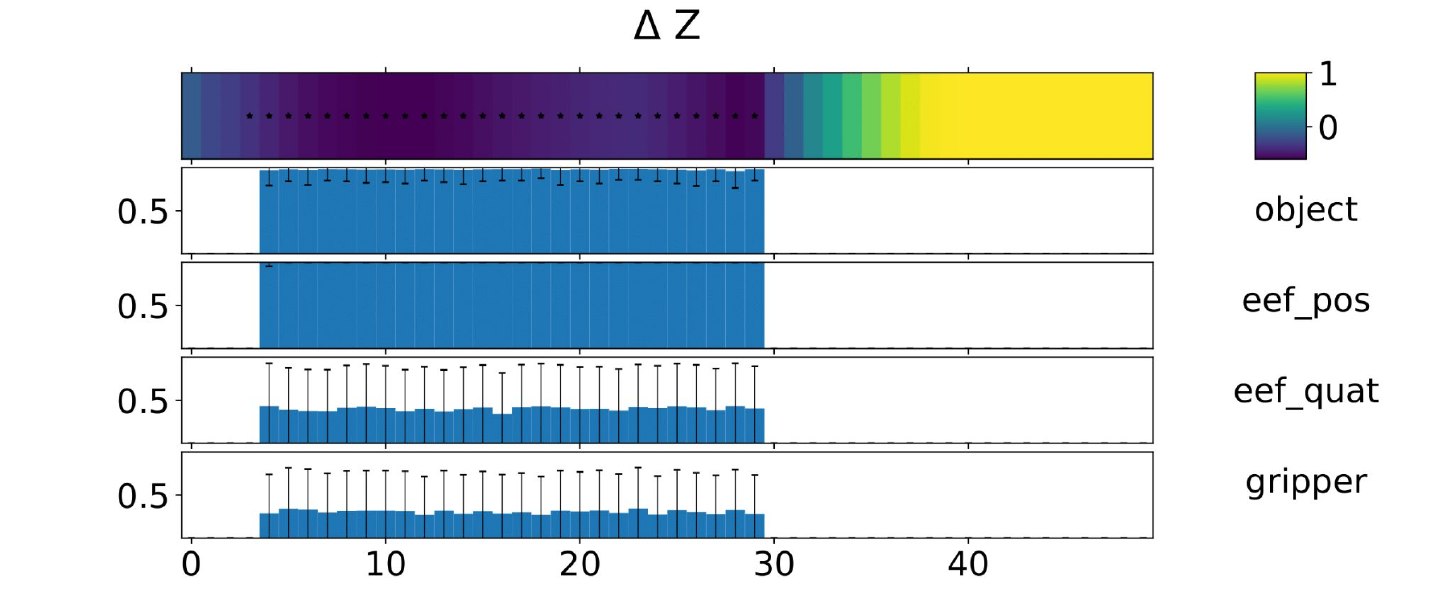}
    \caption{a) Architecture. b) Task : Lift, $C_{A} = \Delta Z$. The colorbar indicates the $\Delta$ actions and the dots on them indicate the actions we consider by following the rule $C_A=${top 25\% of moving down}. The three bar plots indicate the BaTCAVe scores with their uncertainties for three input concepts EEF position, quaternion, and gripper status.}
    \label{fig:Local explaination non image}
    \vspace{-0.5em}
\end{figure}

\subsection{Experiment 2: Lift Cube, Pick \& Place, and Nut Assembly with Proprioceptive Sensors}
\label{sec:exp2}


\emph{Setup}: We use a Panda, a 6-DOF robotic arm with a gripper, to complete the three tasks shown in Table~\ref{table:com_val} on RoboSuite~\cite{robosuite2020}. The setup includes various proprioceptive sensors to monitor the arm's movements and positions. By using the DNN shown in Fig~\ref{fig:Local explaination non image}a and trained on the low-dimensional dataset from robomimic ~\cite{robomimic2021}, we developed an agent based on behavioral cloning~\cite{robomimic2021}. Robot's actions are $\Delta$ differences. Our objective is to explain which concepts of the inputs, $C_I$=\{object, EEF position measurement, EEF quaternion measurement,gripper open width\}, is responsible when the robot is taking a set of actions defined by $C_A = \mathrm{top~75\%~of~each~} \Delta~\text{action}$.

\emph{Results}: We obtain notably high BaTCAVe score with a low uncertainty, corroborating that accurate object information from proprioceptive sensors, unlike in vision-based settings, helps with precisely performing the task. Table~\ref{table:com_val} shows sample scores of $C_A$'s tested across different $C_I$'s. Further, an analysis of per time step explanations for the lifting cube task, depicted in Fig~\ref{fig:Local explaination non image}b, explains that only the object pose and EEF position measurements matters when the EEF is moving down.

\subsection{Experiment 3: Lift Cube with Vision-Language Inputs}
\label{sec:exp3}
\emph{Setup}: As a complex SOTA task, this experiment replicates the setup and conditions of Experiment~\ref{sec:exp2}, but proprioceptive sensors are replaced by a high-resolution RGB images (224$\times$224$\times$3) from the camera and language instructions describing the task. The model was train for 100,000 epochs on 300 demonstrations with a batch size of 64 using a Huber loss function and Adam optimizer. The architecture of the network is shown in Fig~\ref{fig:iamge_robosuite}.

\begin{figure}[t] 
    \centering
    \includegraphics[width=0.17\textwidth]{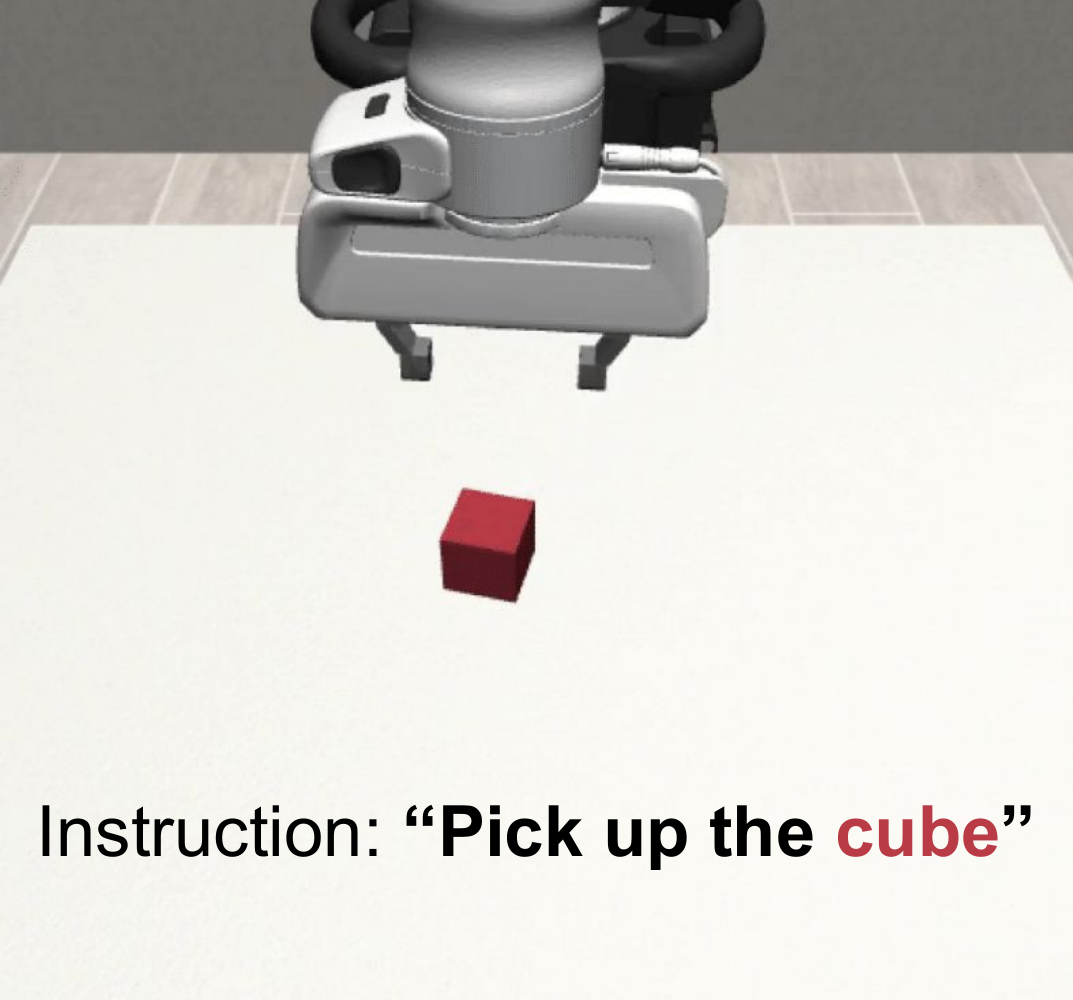}
    \includegraphics[width=0.28\textwidth]{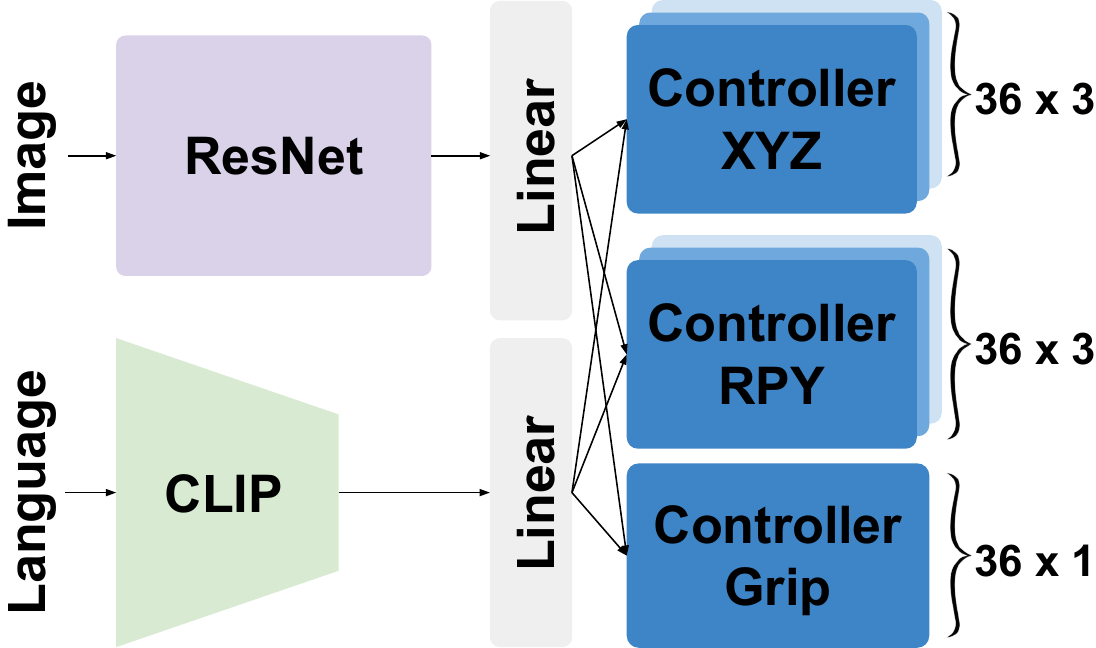}
    \caption{Model architecture for robot control used in Sec~\ref{sec:exp3}, consisting of a ResNet-based Visual Encoder, a Visual Narrower for dimensionality reduction, a CLIP-based Task Encoder. The final linear layer outputs a 360-dimensional action vector, mapping the robot’s motion over 36 timesteps.}
    \label{fig:iamge_robosuite}
    \vspace{-0.5em}
\end{figure}

\emph{Results}: We evaluated $C_{I}$=({images=\{cube, gripper, table\}}, language=\{proper language commands, gibberish, verbs\}). Interestingly, language concepts reveals that the verb in the sentence matters more than the rest of the sentence at the time it lifts the object as seen in Fig~\ref{fig:Local explaination image}(a) where verbs have high score with low uncertainty towards the end of the episode for $C_A = X$. 
\begin{figure}[t]
    \centering
    \includegraphics[width=0.48\textwidth]{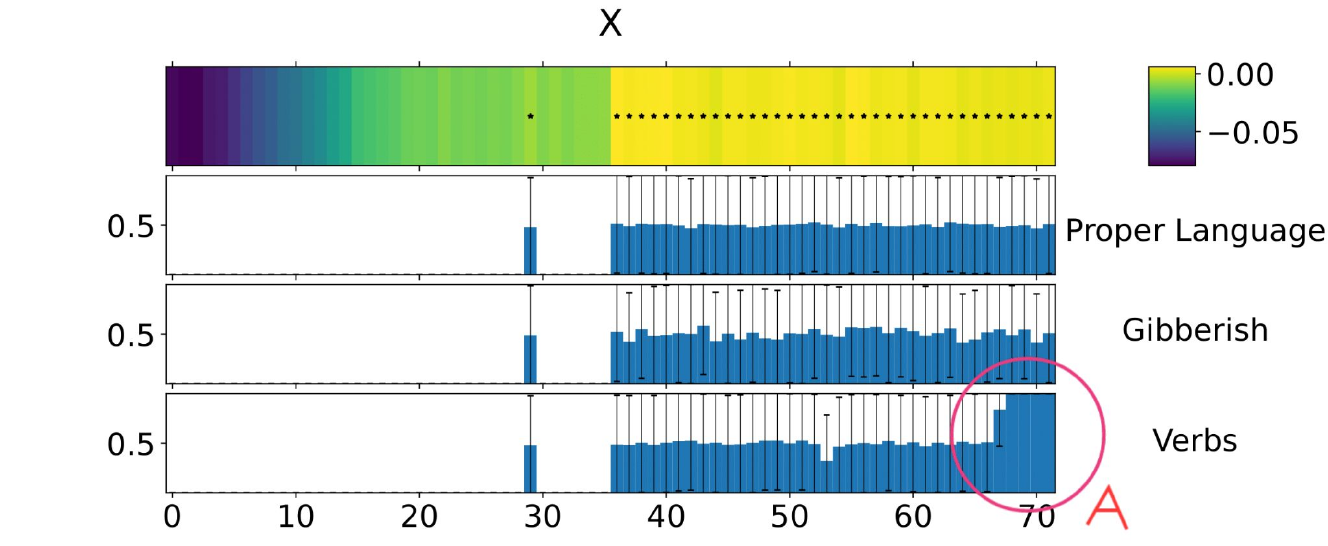}
    \includegraphics[width=0.48\textwidth]{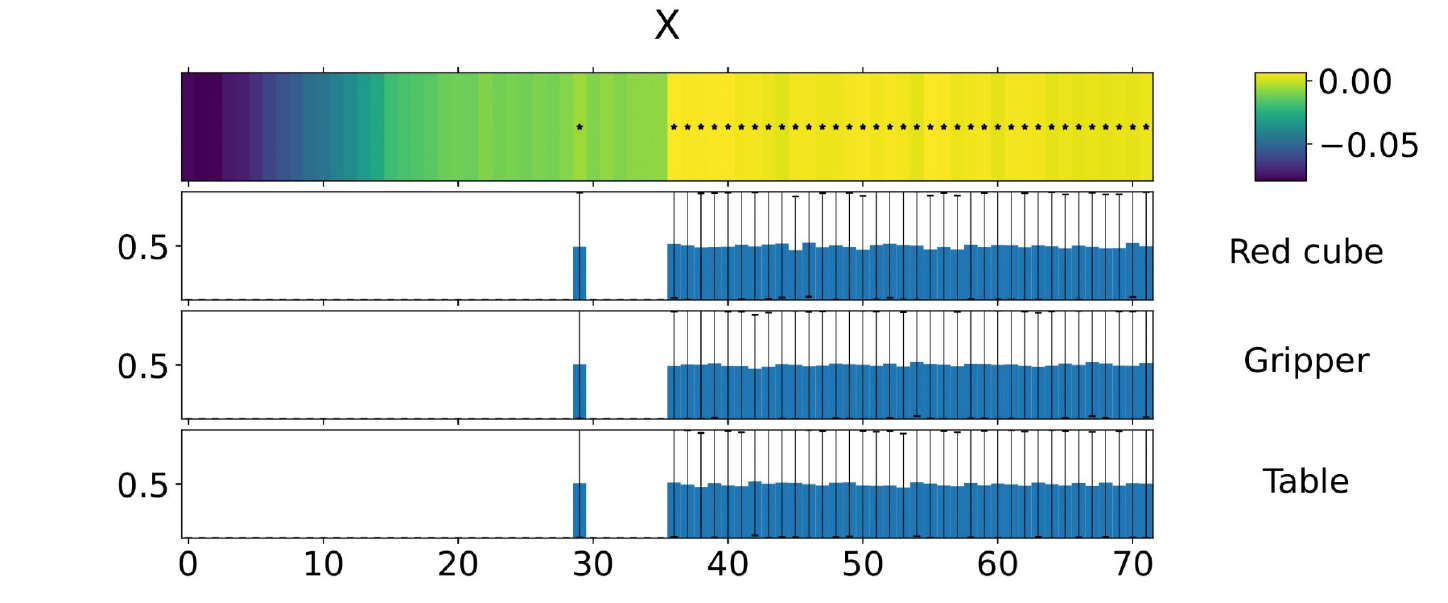}
    \vspace{-0.1em}
    \caption{(a) Indicates verbs in a language command matter most when it is lifting. (b) Indicates the images concepts are highly uncertainy.}
    \vspace{-0.5em}
    \label{fig:Local explaination image}
\end{figure}

\begin{figure}[t]
    \begin{minipage}{0.5\textwidth}
        \centering
        \begin{tabular}{c}
            \toprule
            \textbf{BaTCAVe ratio} \\
            \midrule
            Black/ Orange = 1.05 \\
            Black / Green = 1.03 \\
            Green / Orange = 2.41\\
            Random / Cones = 1.93 \\
            \bottomrule
            
        \end{tabular}
        \label{table:NA_GE}
    \end{minipage} \hfill
    \begin{minipage}{0.5\textwidth}
        \centering
        \includegraphics[width=0.98\textwidth]{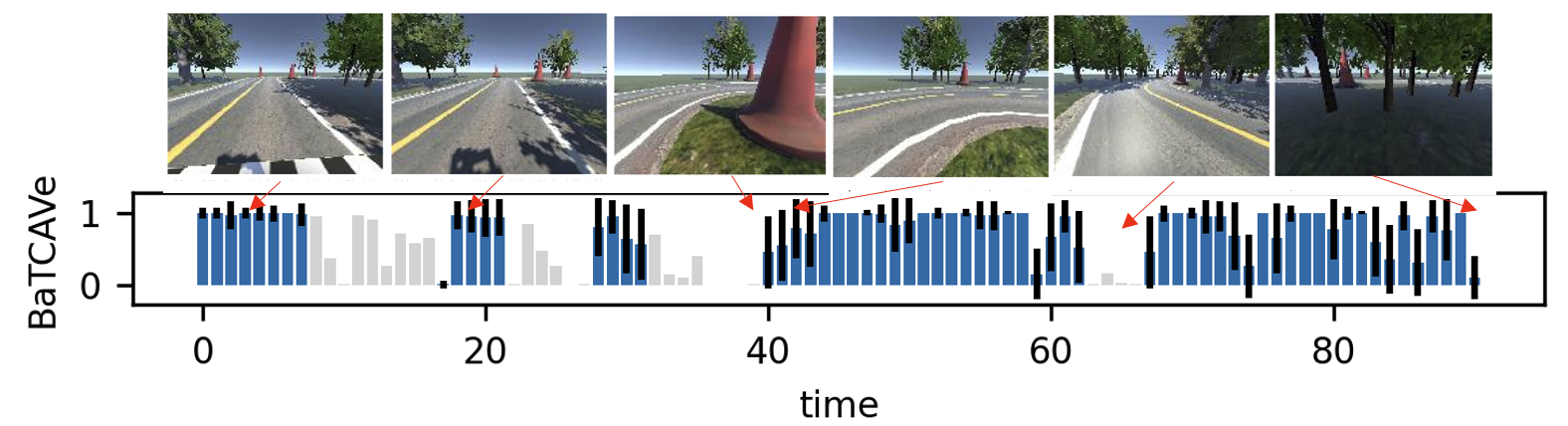}
        \label{fig:donkey_time}
    \end{minipage}
    \caption{(a) The relative importance of concepts indicates that the agent relies on the black concept ($\sim$ road) than the orange concept ($\sim$ cones). (b) Temporal change of scores.}
    \label{fig:donkey}
    \vspace{-0.5em}
\end{figure}

\subsection{Experiment 4: Autonomous Driving with Vision}
\label{sec:exp4}
\emph{Setup}: To simulate autonomous driving, we trained an end-to-end deep reinforcement learning policy using proximal policy optimization (PPO)~\cite{schulman2017proximal} to steer and throttle a vehicle on Donkey Simulator. 
The network is a CNN policy followed by an MLP, which was trained for a total of 100,000 timesteps with learning rate of 0.0003, a rollout of 2048 steps per update, and a batch size of 64.

\emph{Results}: We analyzed how various input concepts affect steering decisions ($C_A = \text{steering angle} > 2$). Fig.~\ref{fig:donkey} shows how the input concept of ``black shades,'' which correlates with roads, explains steering decisions. Around $t=35$, the car points out of the road, and when it steers back we observe that its BaTCAVe for ``black shades'' increases around $t=41$, indicating that it was able to recover by focusing on the road. To improve the policy, it is important to identify what part of the neural network learns incorrect information. By applying BaTCAVe on the last CNN layer ($\bar{s}=0$) and last MLP layer ($\bar{s}=0.822$), we found that errors do not originate in the CNN, which implicitly acts as an image encoder. Therefore, instead of improving the encoder, the policy needs to be trained more.



\section{Conclusion}
\label{sec:conclusion}

We proposed a task-agnostic explainable robotics technique. We demonstrated how our method can be used to explain various robot behaviors across a variety of domains with different decision networks. The actionable insights provided by BaTCAVe helps engineers identify vulnerabilities in various components of robot training---data augmentation, fine-tuning, domain-shift analysis, verification, etc. We show how the uncertainty provided by BatCAVe helps in distinguishing between reliable and ambiguous explanations, which is crucial for high-stakes applications. Future work will focus on explaining sequential decisions rather than specific behaviors and developing methods to amplify activations that have limited signal (e.g., only a few pixels cover ``farther away cones'' in experiment 4). 



\bibliographystyle{IEEEtran}
\bibliography{bibli}

\onecolumn
\section*{Appendix}
\setcounter{section}{0}

\section{Experiment 1: Mobile Robot Navigation}
\label{app:exp1}

In this experiment we considered two distinct setups. In the initial setup we trained the obstacle classifier specifically on an orange box, varying in different orientation and distance, as shown in Fig~\ref{fig:jet_app}. We then compare the variation in importance of concepts caused by the fine-tuning method used and the pre-processing steps involved during training. In the second setup, we generalized the classifier by introducing different objects as obstacles in the training dataset. We test the model with color, darkness and distance as a concept. These criteria were selected based on a human study(Appendix~\ref{app:human}) performed where we ask them to describe important attributes the model might have learned given the dataset. The dataset used to train the classifier model which is used for decision making in the JetBot is collected from the onboard camera attached on top the JetBot. In both the setups, we have 350 images in training and 150 in testing, split equally among both classes as shown in Fig~\ref{fig:orange-dataset}.

\begin{figure}[h]
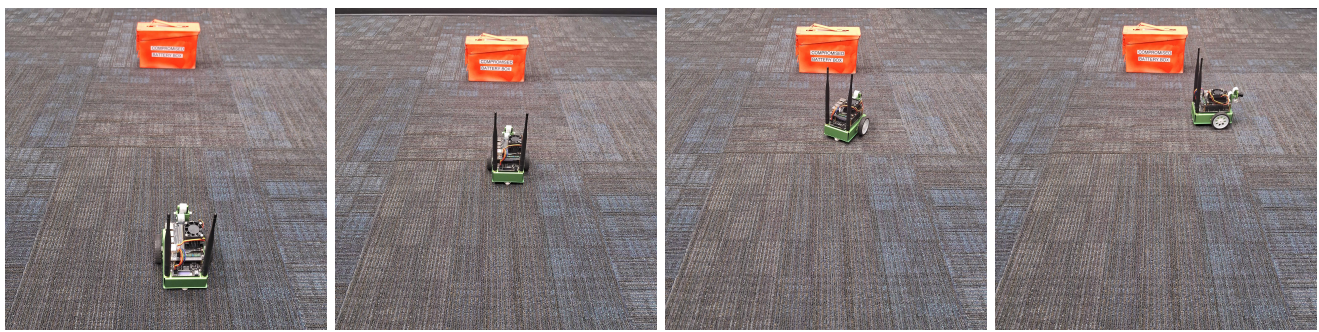

    \centering
    \includegraphics[width=0.24\textwidth]{figs/j11.pdf}
    \includegraphics[width=0.24\textwidth]{figs/j12.pdf}
      \includegraphics[width=0.24\textwidth]{figs/j13.pdf}
      \includegraphics[width=0.24\textwidth]{figs/j14.pdf}
    \caption{Snapshot of JetBot rollout}
    \label{fig:jet_app}
\end{figure}

\begin{figure}
    \centering
    \includegraphics[width=0.85\textwidth]{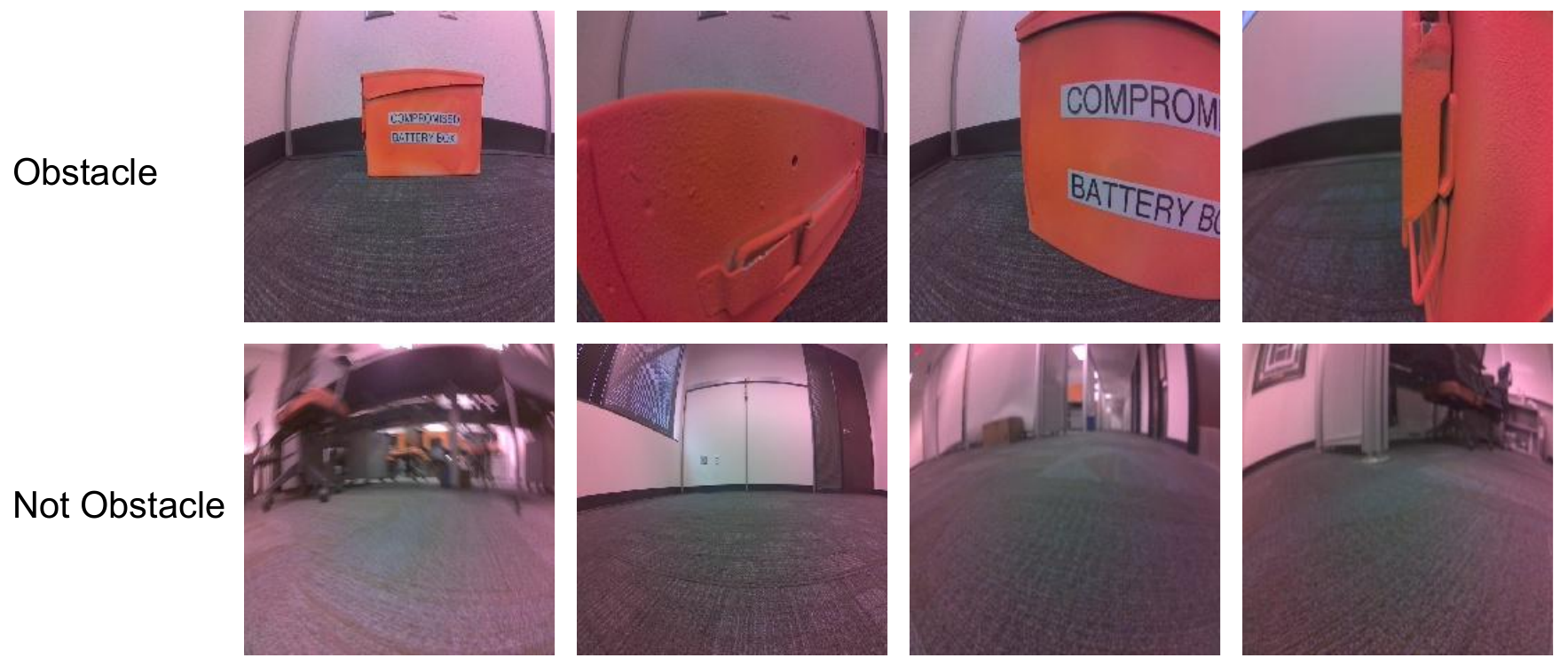}
    \caption{Orange box Obstacle and Not obstacle dataset sample}
    \label{fig:orange-dataset}
\end{figure}

\textbf{Jetbot configuration}: The JetBot is a compact robot built on the NVIDIA Jetson Nano platform, designed for AI and robotics applications. Key components of the JetBot include: The Jetson Nano board which has a 4GB LPDDR4 RAM, a 128-core NVIDIA Maxwell GPU, and a quad-core ARM Cortex-A57 MPCore processor. An 8 MP wide-angle camera is attached on top. It also includes two motors and a motor driver for precise control of the robot's wheel movement.

\textbf{AlexNet}~\cite{krizhevsky2012imagenet}:  AlexNet is the classifier model we use for decision making in JetBot. AlexNet was used due its high processing speed which is critical in robotics. We change the architecture by replacing the final layer with a layer with 2 nodes.

\textbf{Training}: We finetuned the AlexNet model with orange box obstacle dataset and general object obstacle dataset. We finetuned the models with FFT and LFT for 100 epochs with cross-entropy loss and SGD optimizer with learning rate of 0.001 and momentum 0.9, while applying color jitter (CJ) and/or image rotation (R). The training loss across different experiments are shown in Fig~\ref{fig:jetbot train loss}.

\subsection{Concepts}
\label{app:jetbot_concepts}

Fig~\ref{fig:concept-jetbot} shows the concepts which were used across different experiments. Random concepts were creating by randomly picking a unique color for each pixel value in the image. We evaluated the models with dark-light concept, orange-random concept, distance 30-random concept, distance 45-random concept, and distance 60-random concept sets.

\begin{figure}[h]
    \centering
    \includegraphics[width=0.75\textwidth]{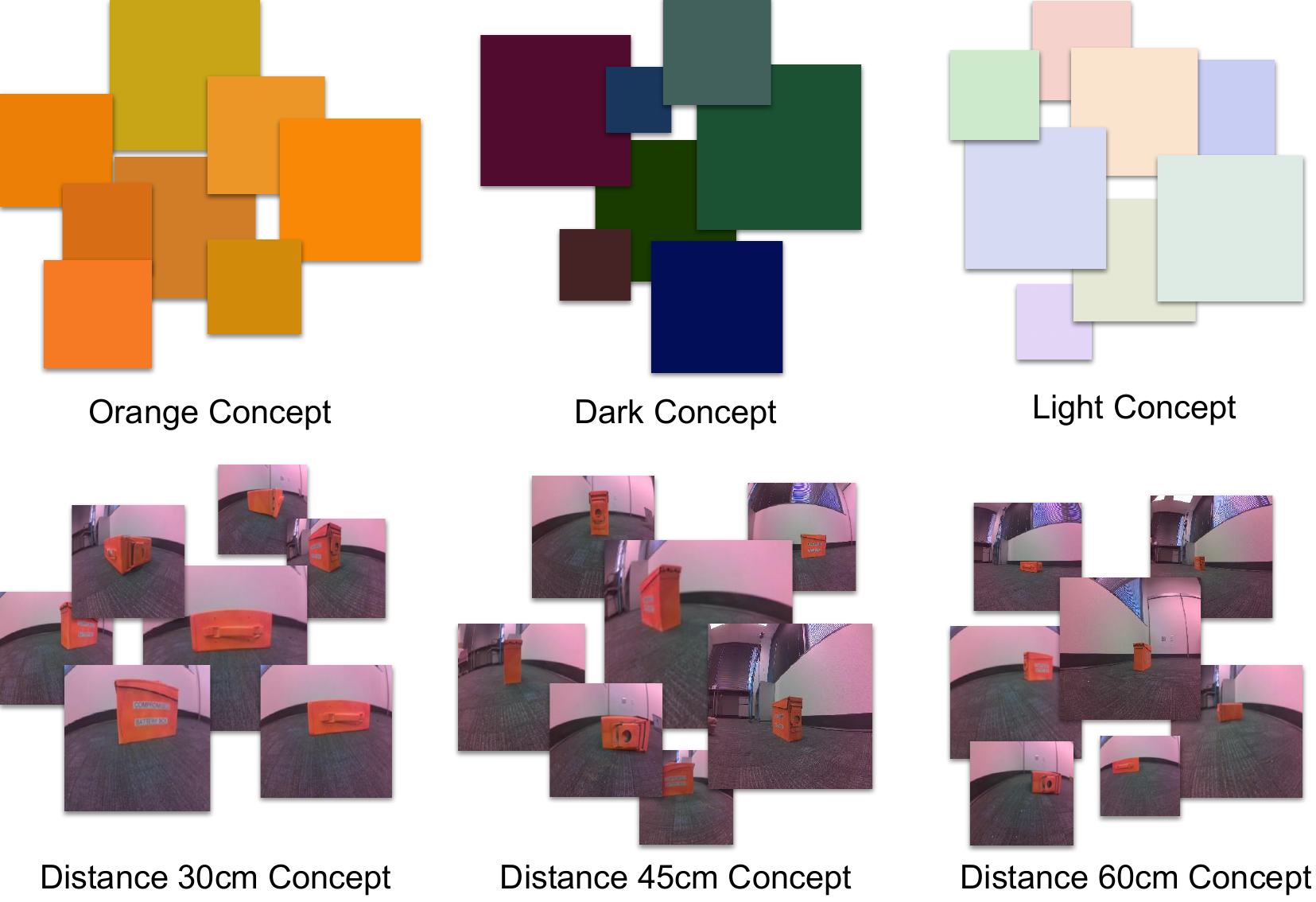}
    \caption{Different concepts sets used in BaTCAVe for Experiment 1.}
    \label{fig:concept-jetbot}
\end{figure}

\subsection{Additional Results}
\label{app:jetbot_addresults}

\begin{figure}[h]
    \centering
    \includegraphics[width=0.25\textwidth]{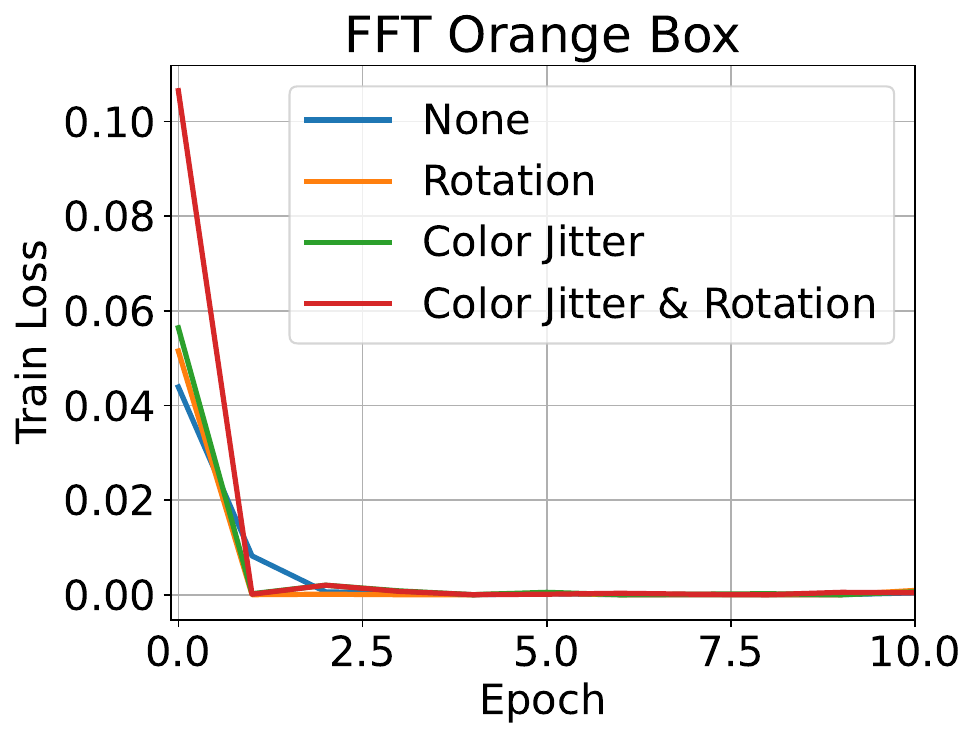}
    \includegraphics[width=0.25\textwidth]{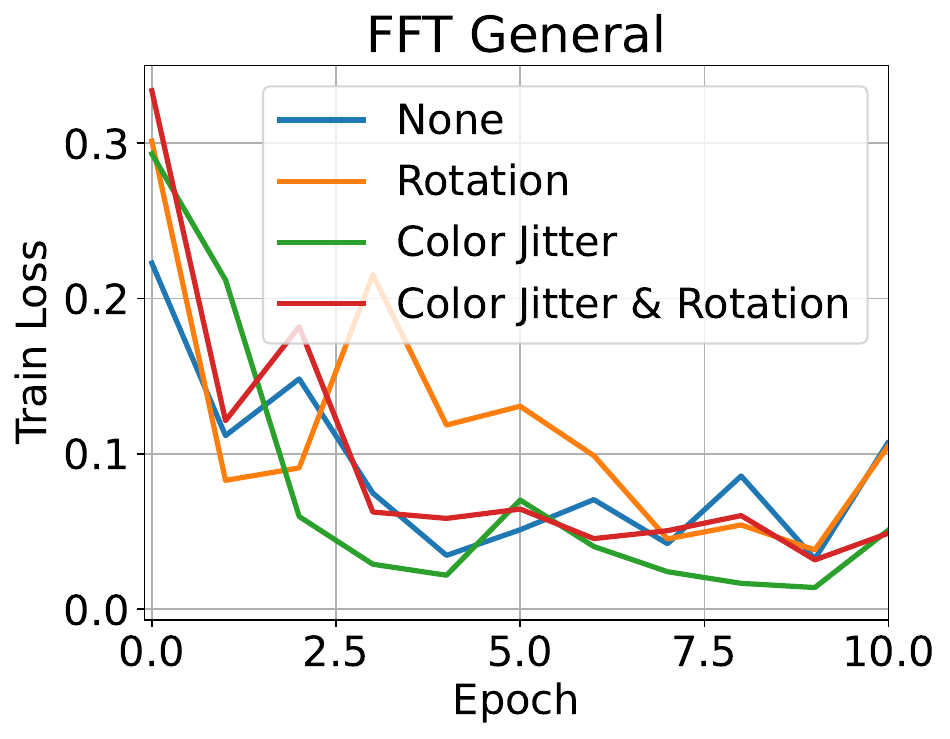}
    \includegraphics[width=0.25\textwidth]{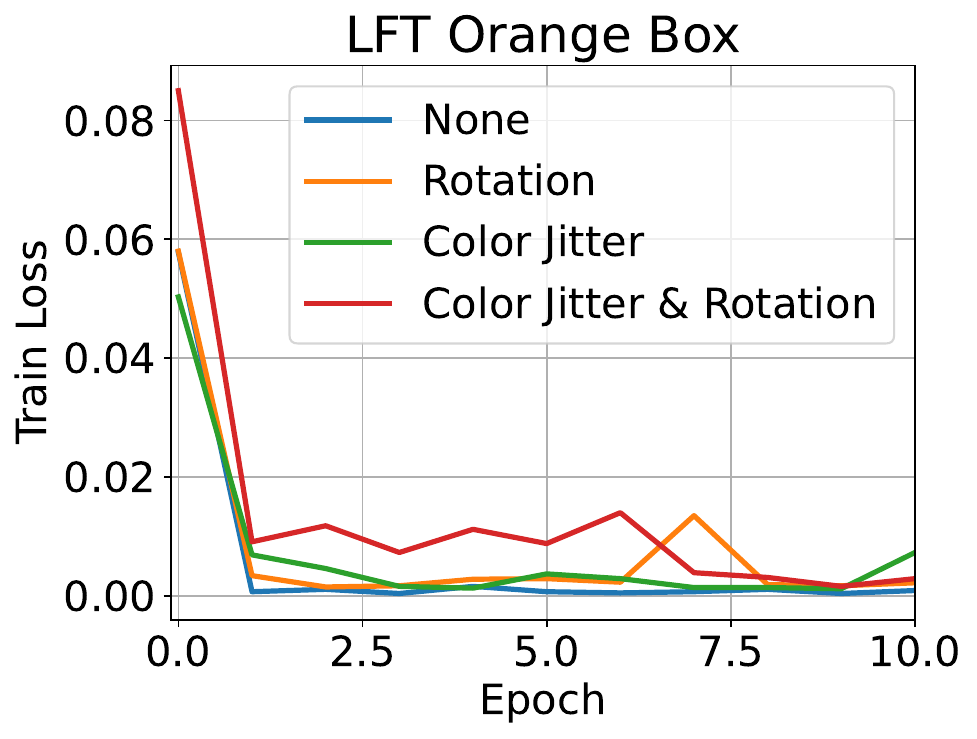}
    \includegraphics[width=0.25\textwidth]{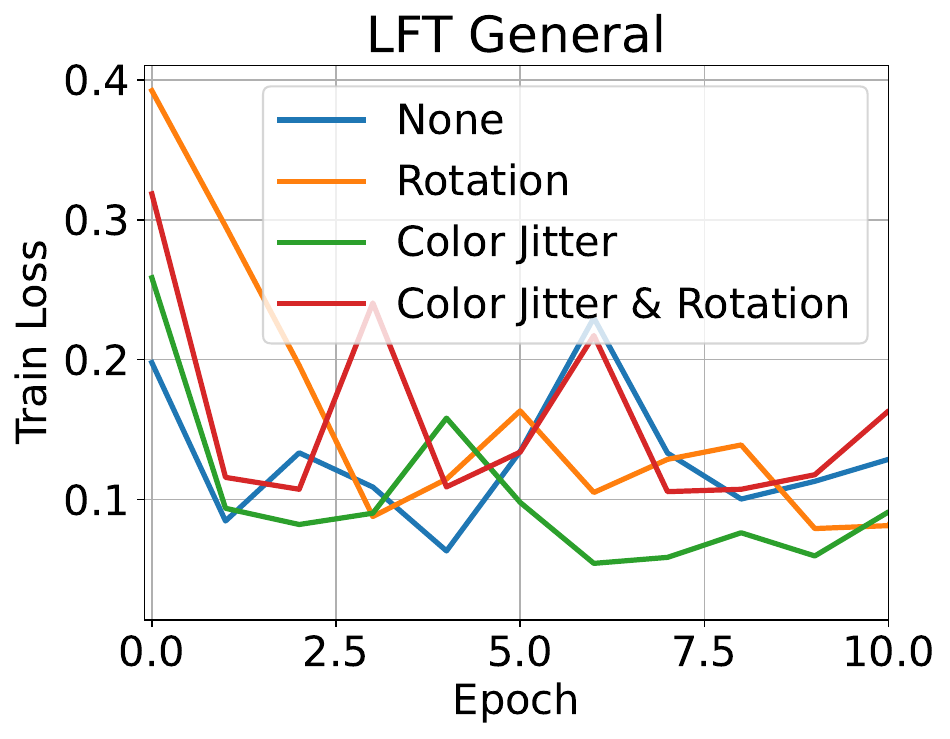}
    \caption{Training loss of FFT (left) and LFT (right) over orange obstacle and general obstacle dataset and different augmentations.}
    \label{fig:jetbot train loss}
\end{figure}

Fig~\ref{fig:concept-test-general-fft} and Fig~\ref{fig:concept-test-general-lft} shows the results of general model, finetuned with all parameters and final layer parameters, respectively. They are compared over different concepts and augmentation techniques. We see that in both the cases models were not able to understand dark difference between dark and light concept.

\begin{figure}[h]
    \centering
    \includegraphics[width=1\textwidth]{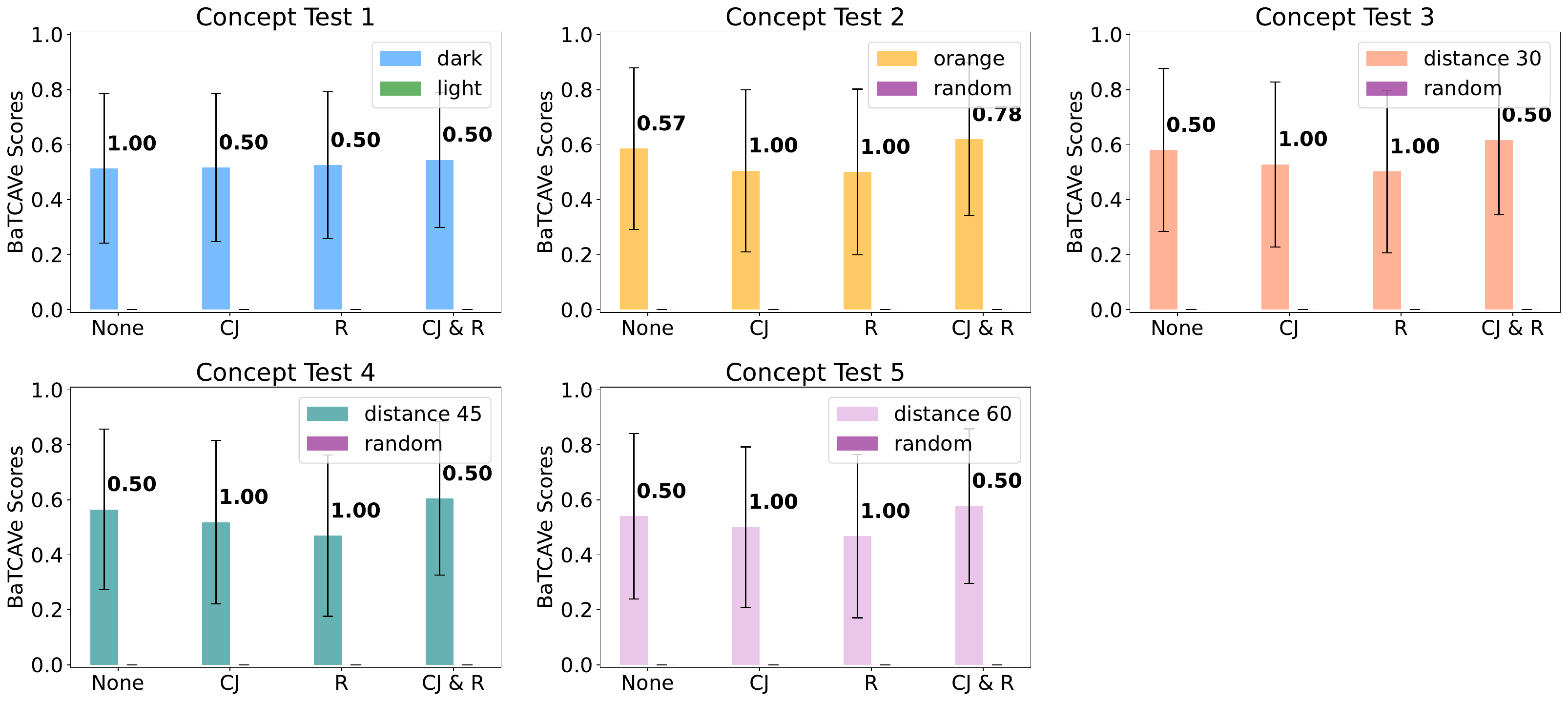}
    \caption{Different concepts test on FFT model trained on general dataset over different augmentations. (BaTCAVe's classifier accuracy is shown near the bar graph for each augmentation.)}
    \label{fig:concept-test-general-fft}
\end{figure}

\begin{figure}[h]
    \centering
    \includegraphics[width=1\textwidth]{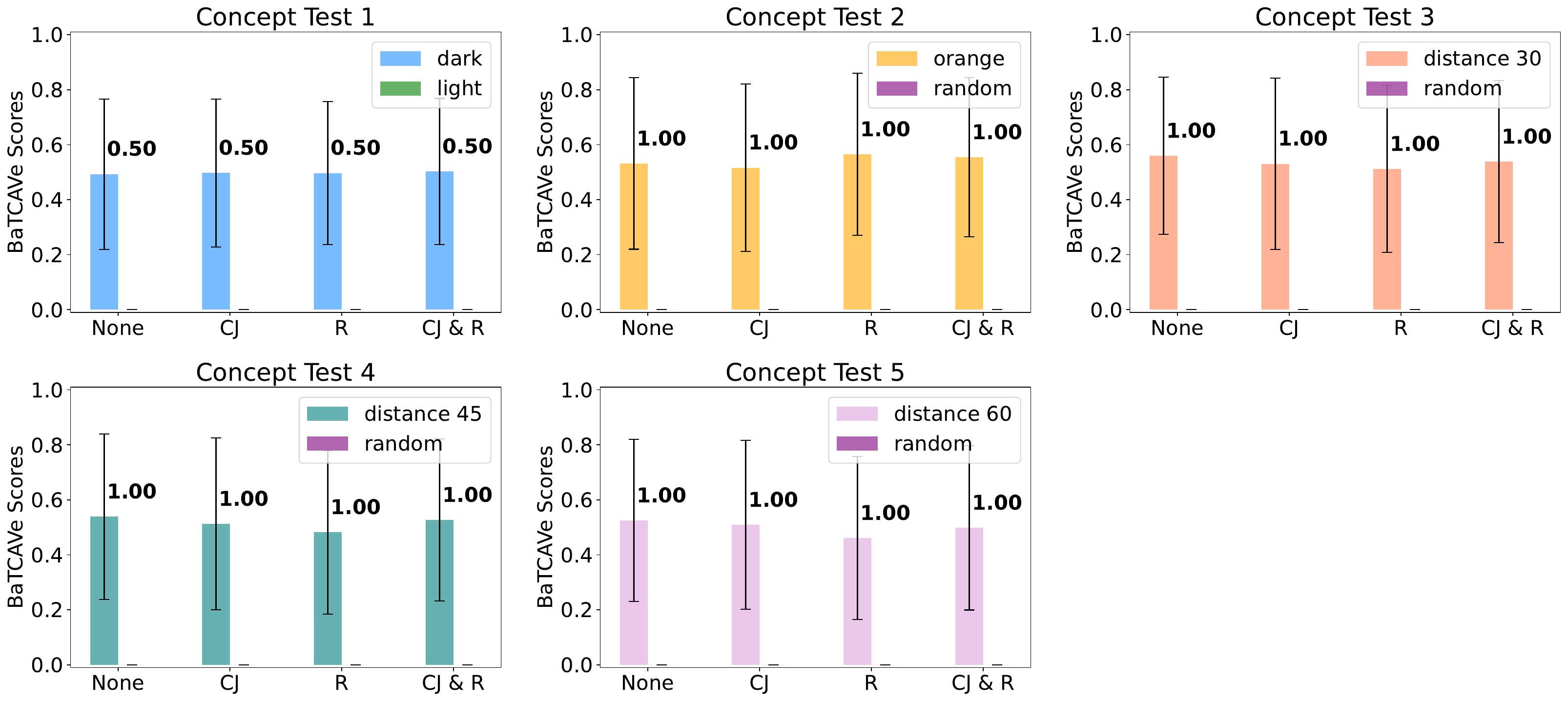}
    \caption{Different concepts test on LFT model trained on general dataset over different augmentations. (BaTCAVe's classifier accuracy is shown near the bar graph for each augmentation.)}
    \label{fig:concept-test-general-lft}
\end{figure}


\subsection{Human survey}
\label{app:human}
Fig~\ref{human-survey}(a) shows the screenshot of the survey that was circulated. Fig~\ref{human-survey}(b) is the distribution of $C_I$ chosen by the participants (20). Fig~\ref{human-survey}(c) shows the word cloud representation of all the replies gotten by the participants. 

\begin{figure}[t]
\vskip 0.2in
\begin{center}
    \includegraphics[width=1\columnwidth]{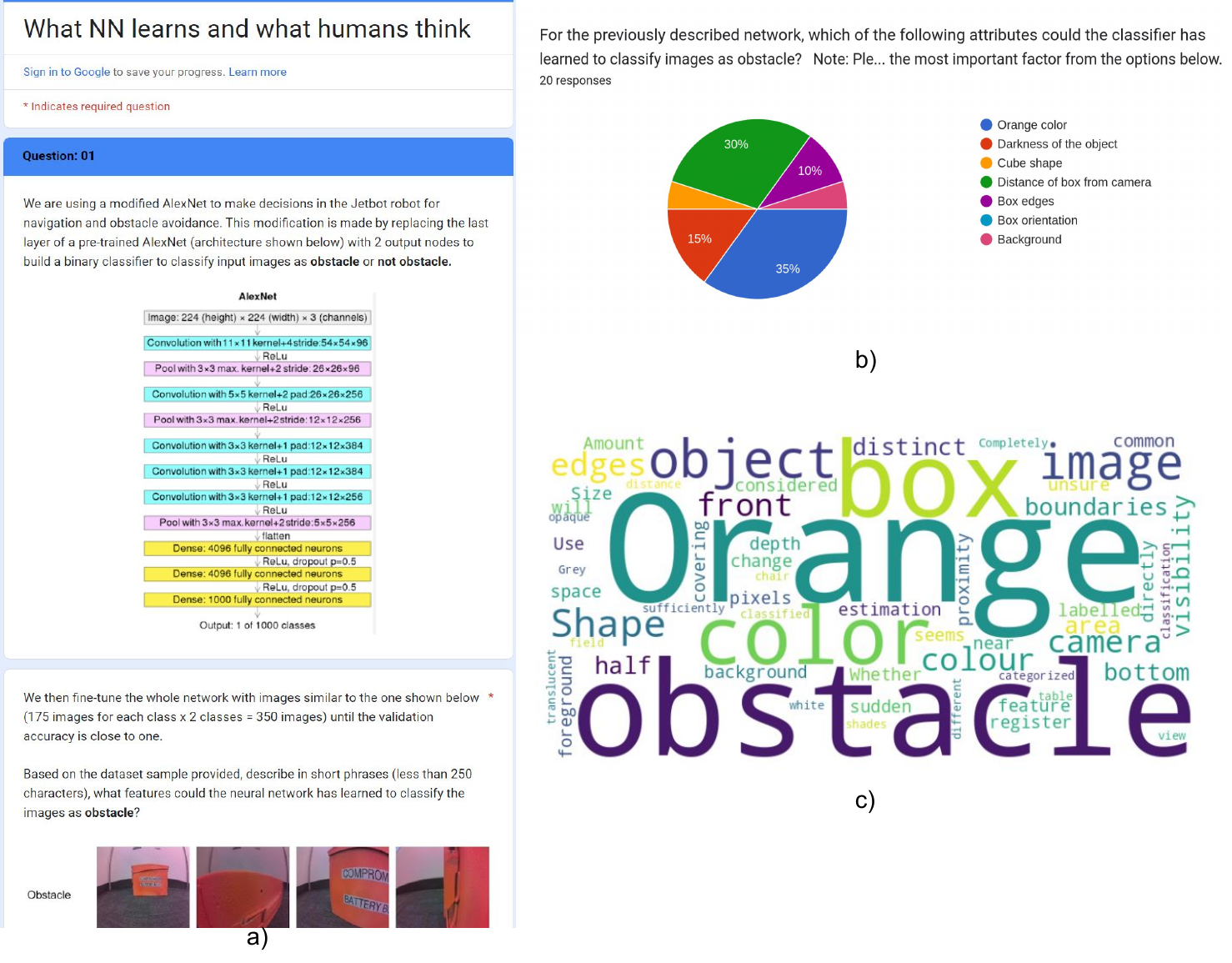}
    \caption{Human Survey: a) Survey screenshot b) Percentage of concept chosen c) Wordcloud of response given}
    \label{human-survey}
    \end{center}
    \vskip -0.2in
\end{figure}

\section{Experiment 2: Tasks with Proprioceptive Sensors}
\label{appendix:3.1}

\subsection{Model details}
\label{app:exp2_MT}
\begin{figure}[h]
    \centering
    \includegraphics[width=0.32\textwidth]{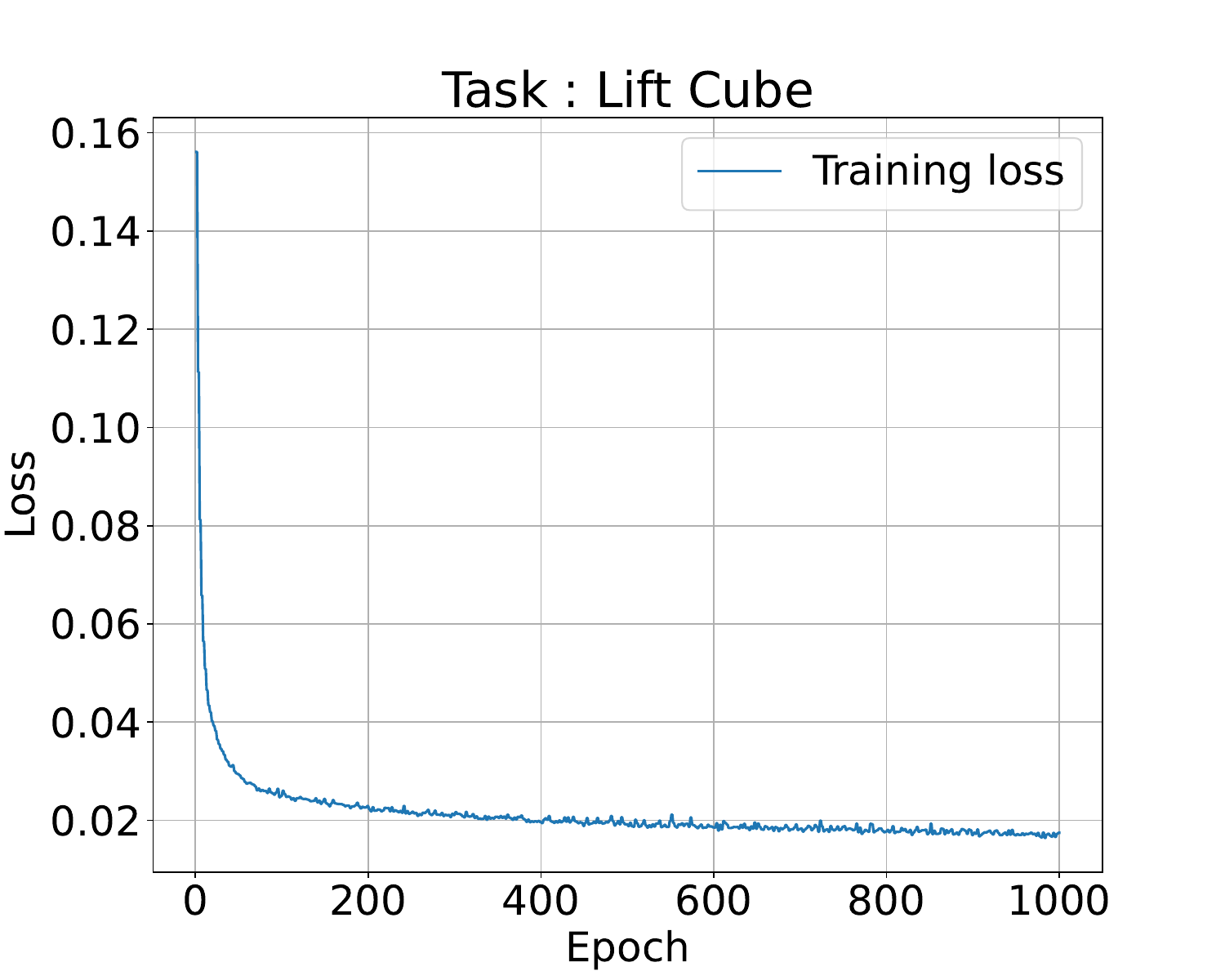}
      \includegraphics[width=0.32\textwidth]{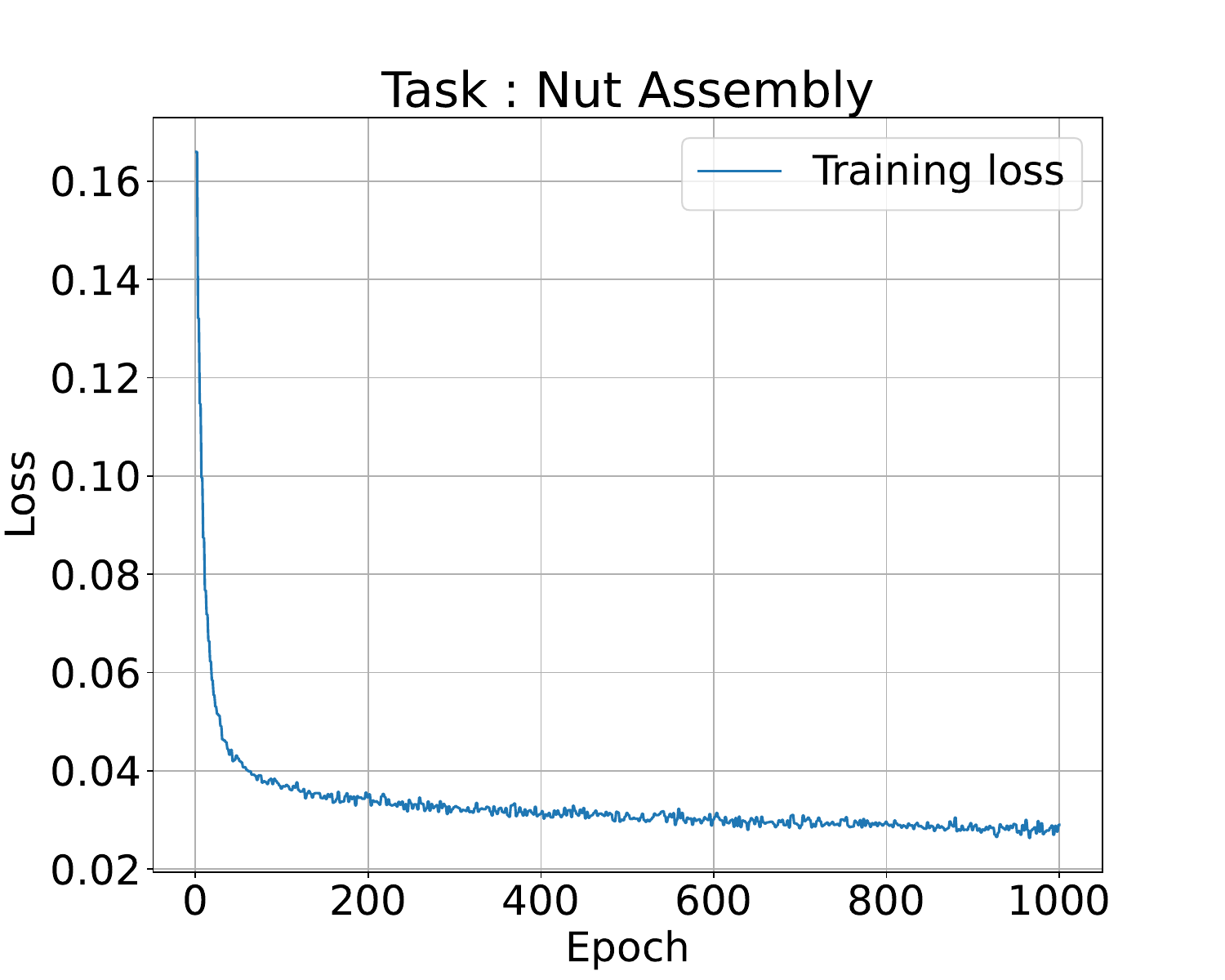}
      \includegraphics[width=0.32\textwidth]{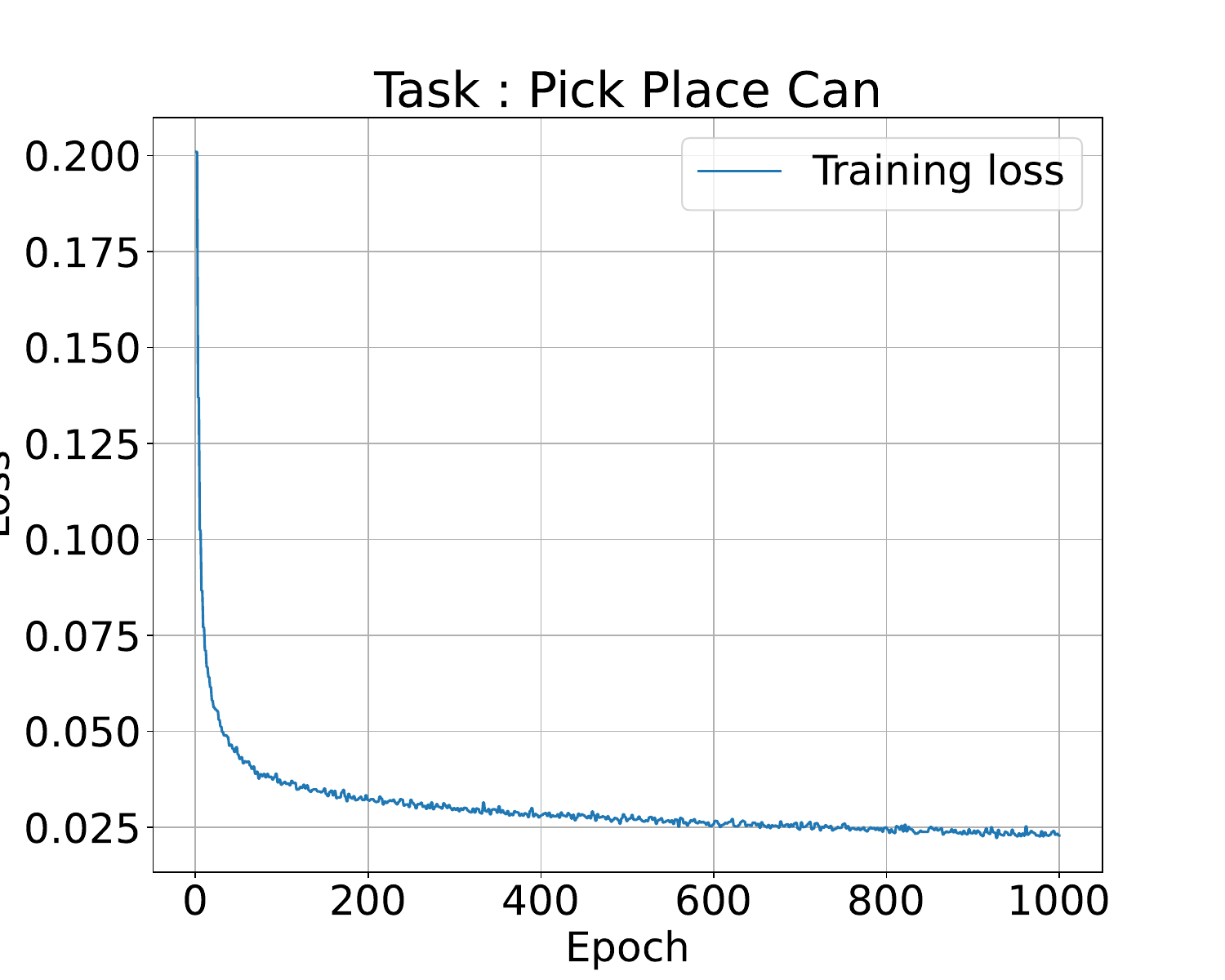}
    \caption{Training loss of model across tasks}
    \label{fig:ex2 model losses}
\end{figure}

\textbf{Architecture}: The model architecture, depicted in Fig~\ref{fig:Local explaination non image}(a), employs a low-dimensional observation modality that integrates multiple sensor inputs: gripper position (2-vector), end-effector position (3-vector), object characteristics (10-14 vector), and end-effector orientation (4-vector). The network outputs a 7-dimensional action vector which includes delta values for the robot's movements: three for end-effector position (x, y, z), three for orientation (roll, pitch, yaw), and one for gripper force. 

The model has 3 main components: Observation Encoding, Multi-Layer Perceptron(MLP), Observation Decoding.
\begin{enumerate}
    \item Observation Encoding: Prior to processing, observed states undergo encoding through an Observation Group Encoder. This encoder handles observation modalities, which are object information, robot end effector position and orientation, and gripper state. The encoded observations are concatenated into a single vector representation.
    \item MLP: The concatenated vector is then fed into an MLP comprising a single hidden layer with ReLU activation, facilitating the extraction of high-level features from the encoded observations. The output dimensionality of the MLP is 1024.
    \item Observation Decoding: Following feature extraction, the output of the MLP is decoded to produce the desired actions. This decoding process involves a linear transformation, mapping the MLP output to the action space. The actions consist of seven elements, corresponding to the changes in position (x, y, z) and orientation (roll, pitch, yaw) of the robot end effector, as well as the gripper state.
\end{enumerate}

\textbf{Training}: We train the model using the ph low\_dim dataset from robomimic~\cite{robomimic2021} on task can,lift and sqare. The training process begins with data normalization to ensure consistent input scaling. The training loop spans 500 epochs, each consisting of 100 gradient steps. During each step, the model calculates the mean squared error (MSE) loss between predicted and ground truth actions. This loss is then used to update the model parameters using the Adam optimizer.

This model was trained across three different task:
\begin{enumerate}
    \item Pick Place Can(PPC): The primary goal of this task is for the robot to accurately pick up a can from one location using a gripping mechanism and then move it to a specified area to place it down. Fig~\ref{fig:task_appppc} shows the state across different timesteps and Fig~\ref{fig:app-ppc} shows BaTCAVe performed across multiple $C_A$ on PPC.
    \item Nut Assembly(NA): This task requires the robot to pick up a hollow object and accurately pick its handle and fit it in the object which fits the hollow area. Fig~\ref{fig:task_appna} shows the state across different timesteps and Fig~\ref{fig:app-na} shows BaTCAVe performed across multiple $C_A$ on NA.
    \item Lift Cube(LC): In the LC task, the robot's goal is to lift a cube from the table. Fig~\ref{fig:task_applift} shows the state across different timesteps and Fig~\ref{fig:app-lift} shows BaTCAVe performed across multiple $C_A$ on LC.
\end{enumerate}

Table~\ref{table:com_val_2} shows BeTCAVe scores across all tasks and $C_A$'s.

\begin{figure}[h]
    \centering
    \includegraphics[width=0.24\textwidth]{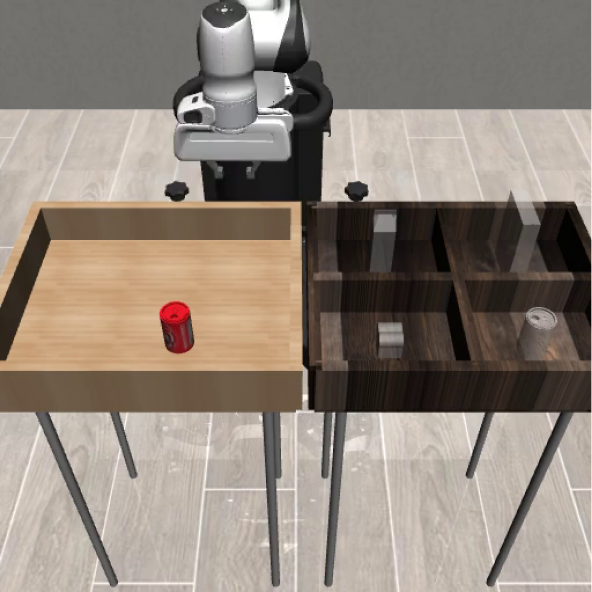}
      \includegraphics[width=0.24\textwidth]{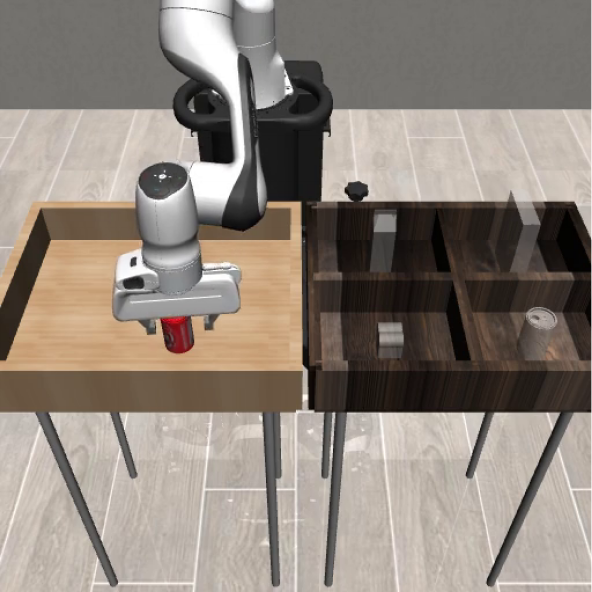}
      \includegraphics[width=0.24\textwidth]{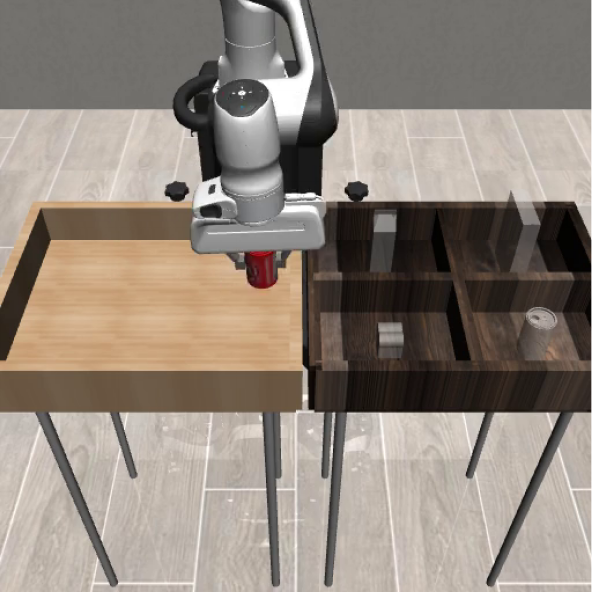}
      \includegraphics[width=0.24\textwidth]{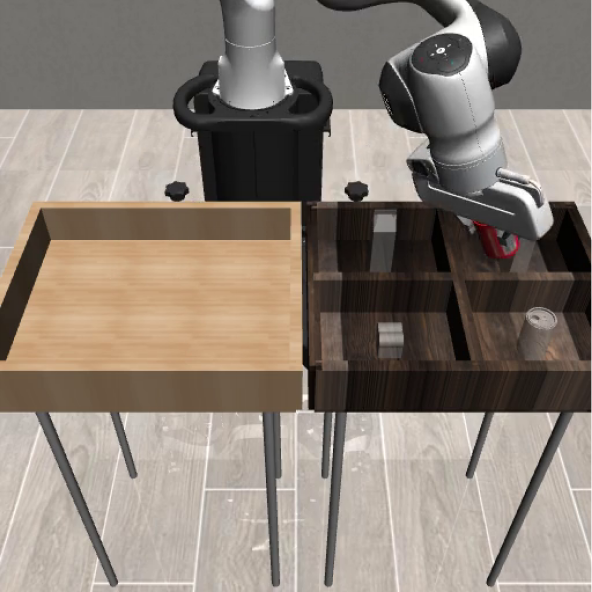}
    \caption{Snapshot of PPC task rollout}
    \label{fig:task_appppc}
\end{figure}

\begin{figure}[h]
    \centering
    \includegraphics[width=0.49\textwidth]{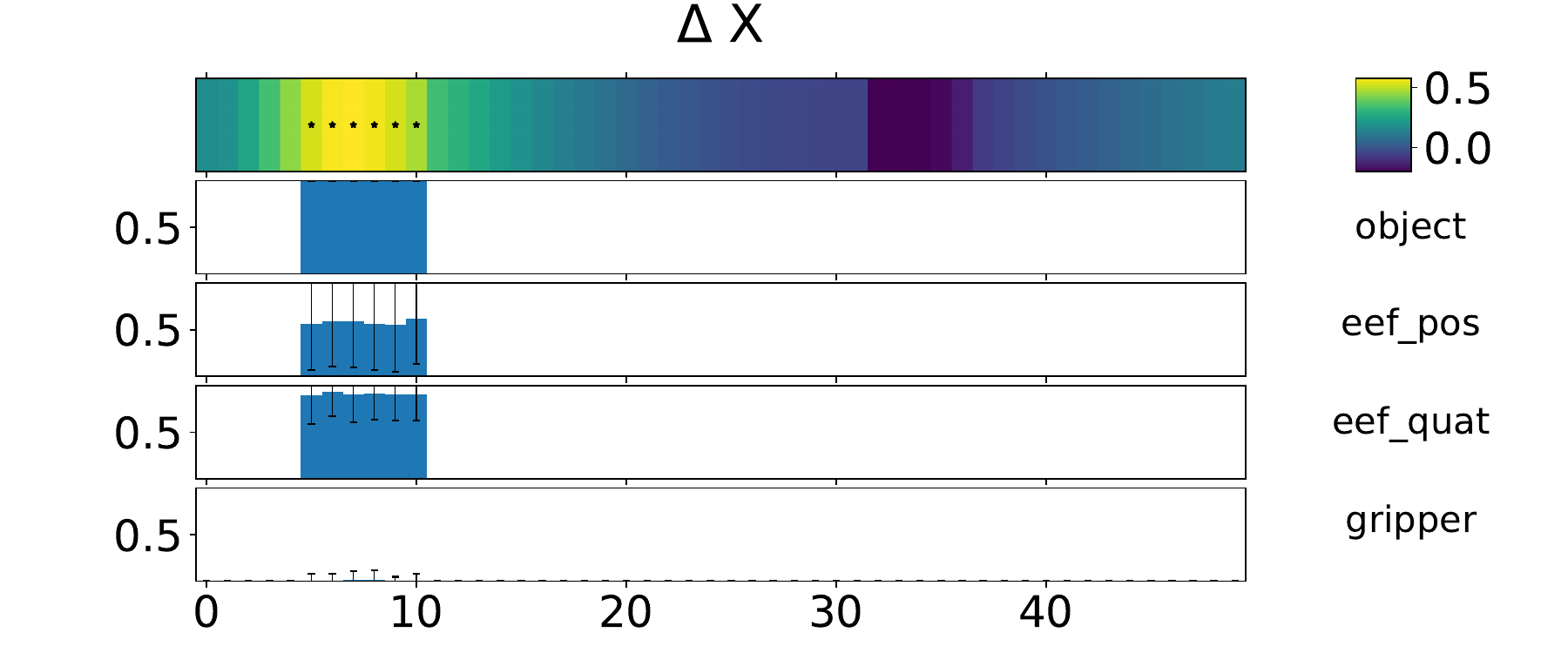}
    \includegraphics[width=0.49\textwidth]{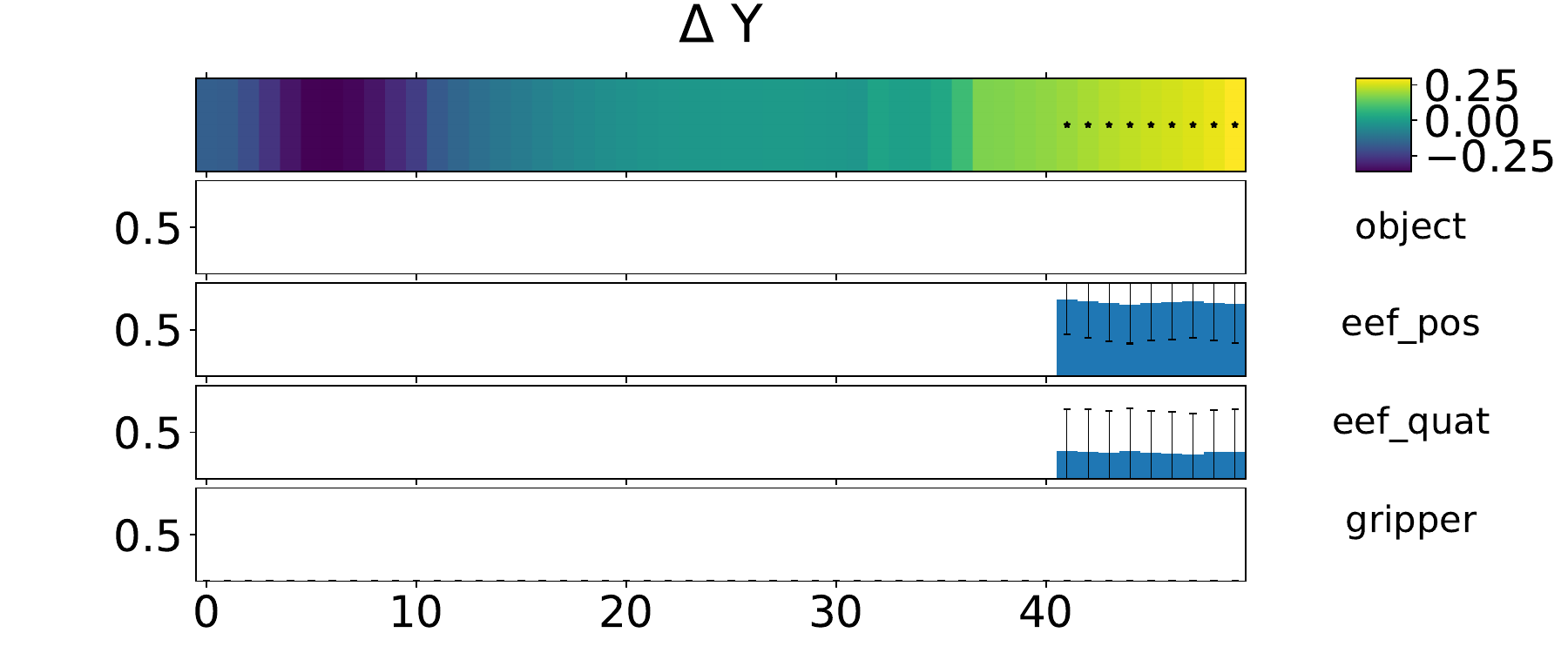}
    \includegraphics[width=0.49\textwidth]{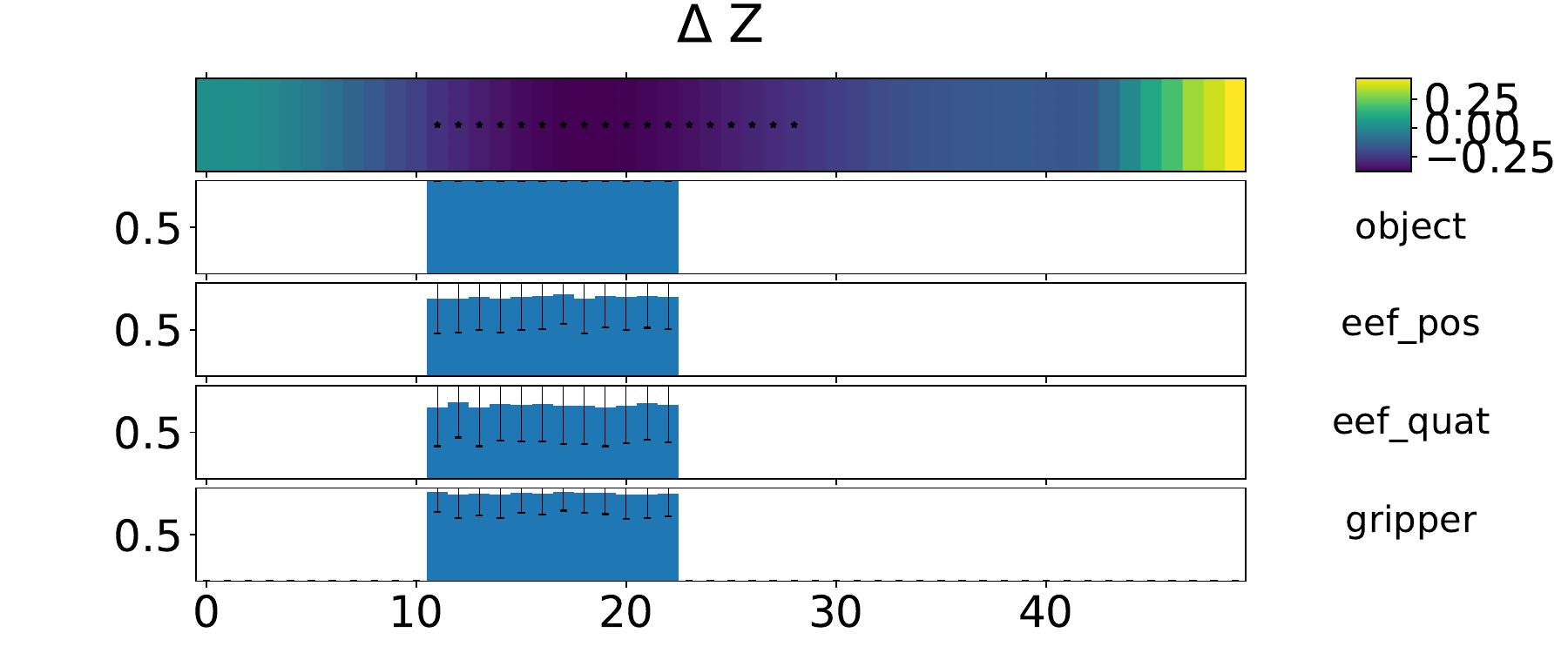}
    \includegraphics[width=0.49\textwidth]{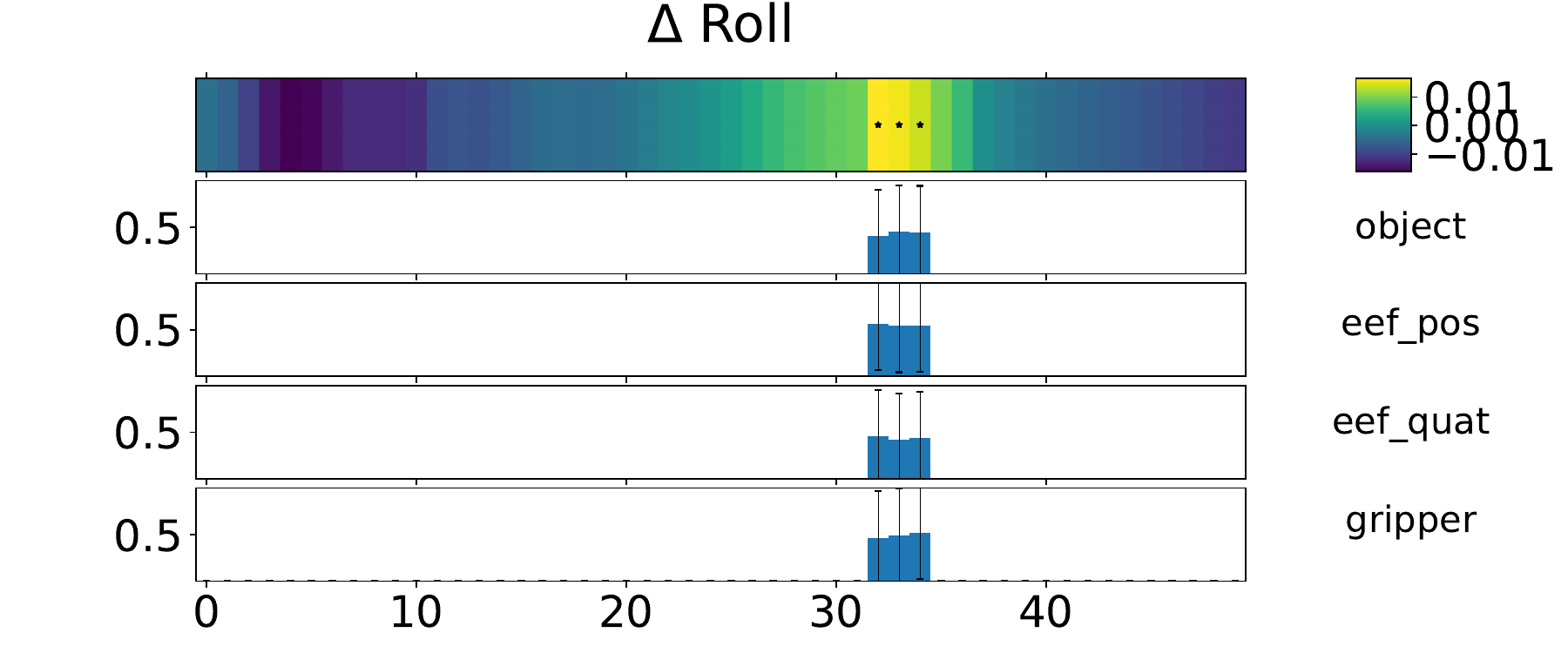}
    \includegraphics[width=0.49\textwidth]{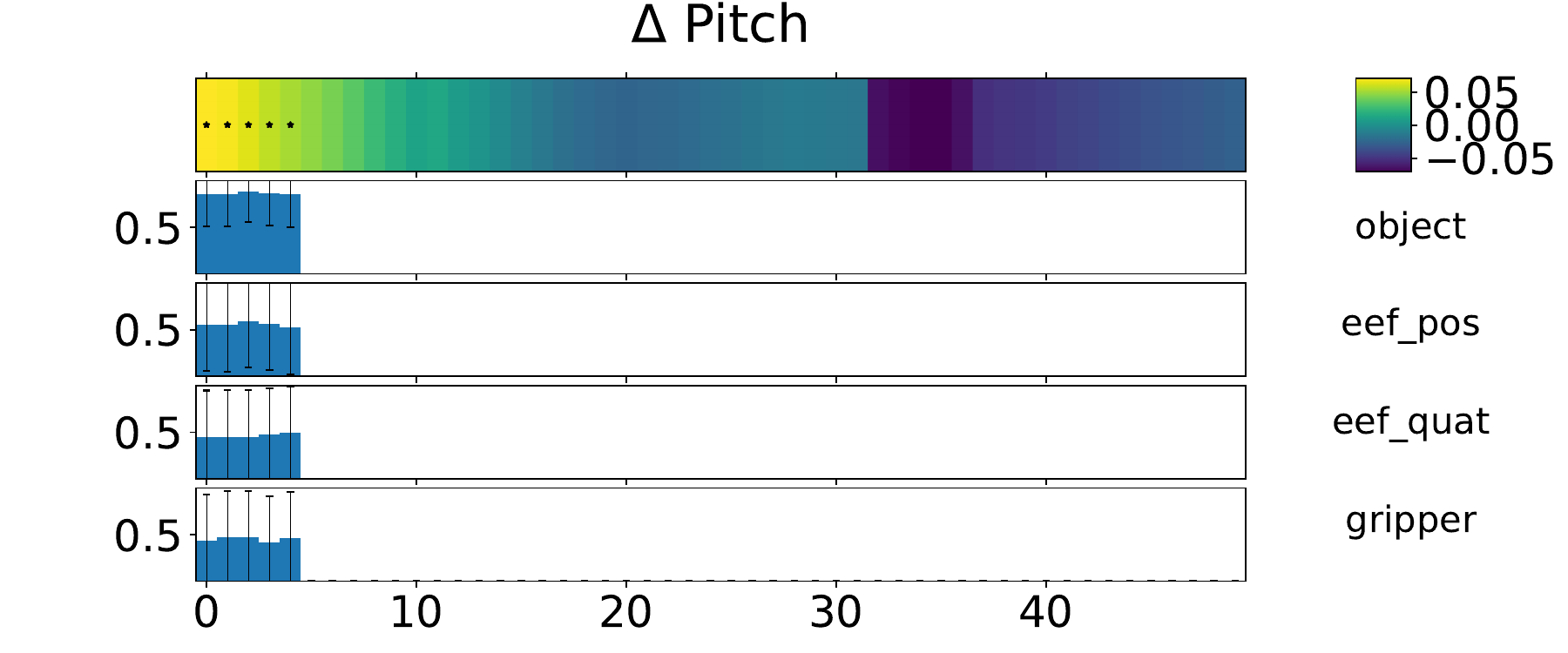}
    \includegraphics[width=0.49\textwidth]{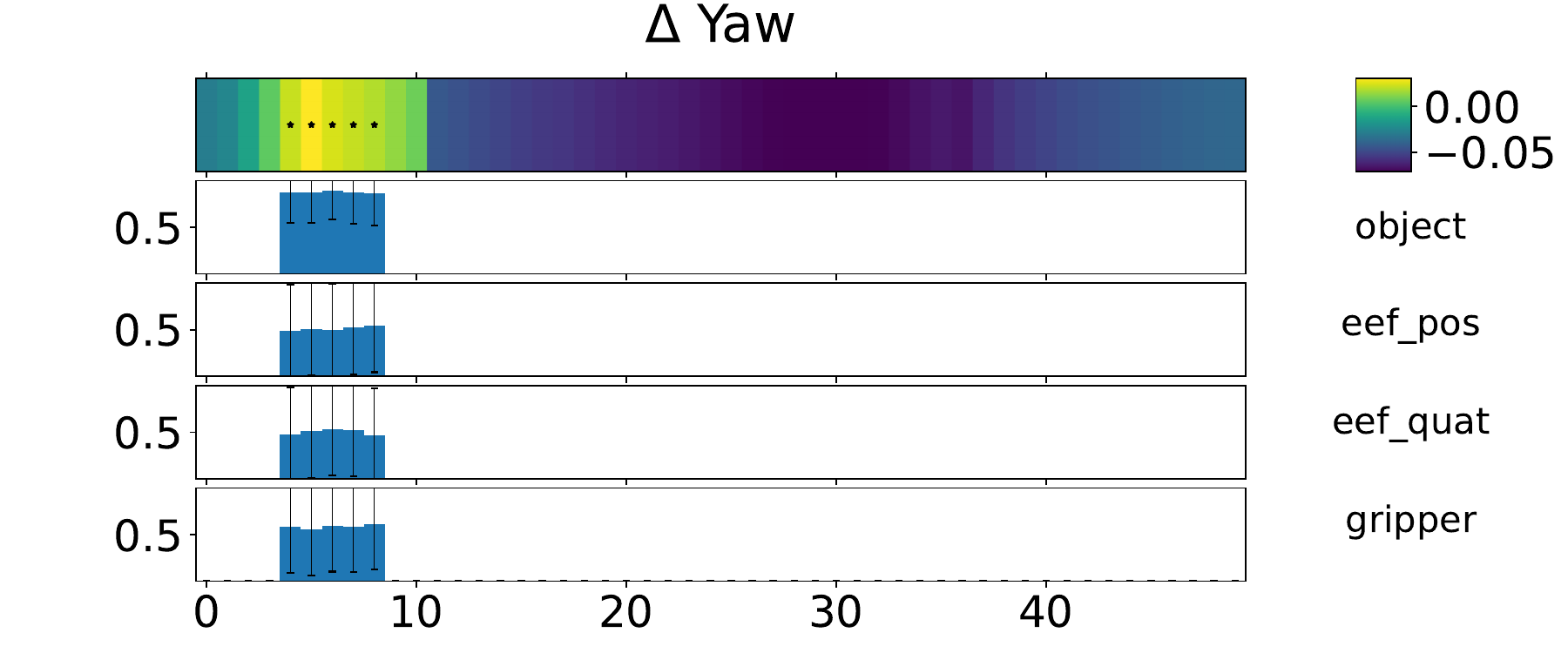}
    \includegraphics[width=0.49\textwidth]{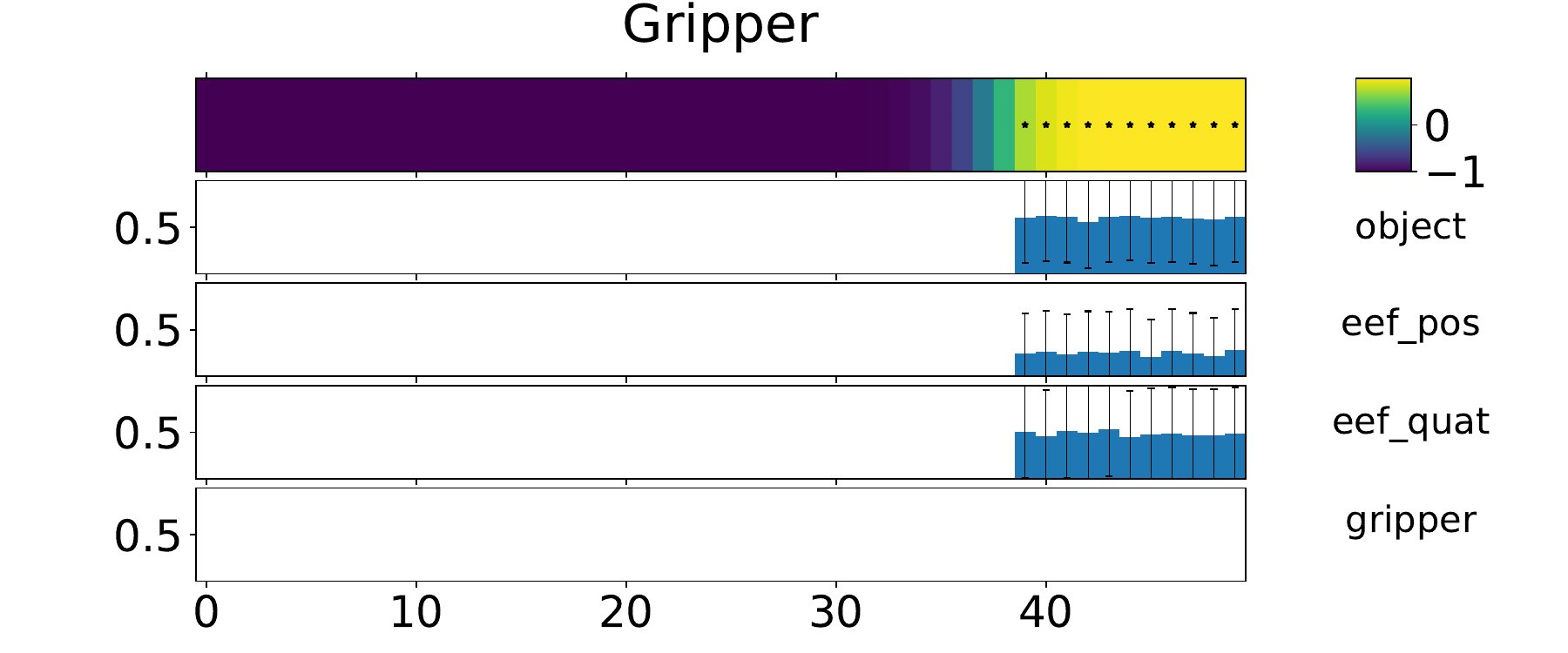}
    \caption{BeTCAVe across multiple $C_A$ on task PPC}
    \label{fig:app-ppc}
\end{figure}

\begin{figure}[h]
    \centering
    \includegraphics[width=0.24\textwidth]{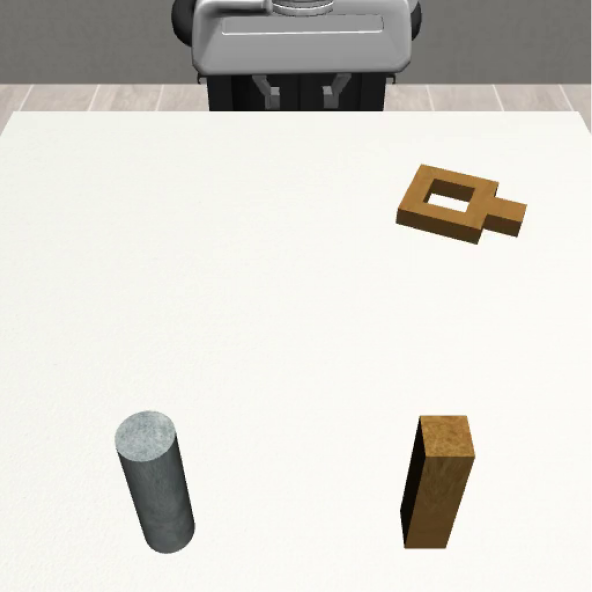}
      \includegraphics[width=0.24\textwidth]{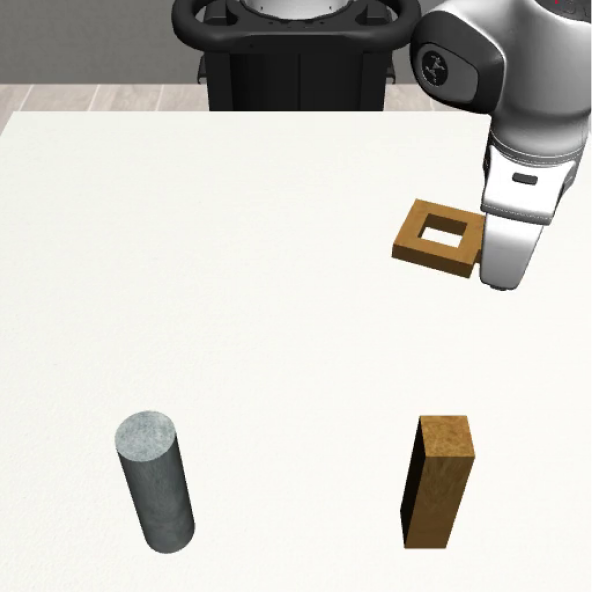}
      \includegraphics[width=0.24\textwidth]{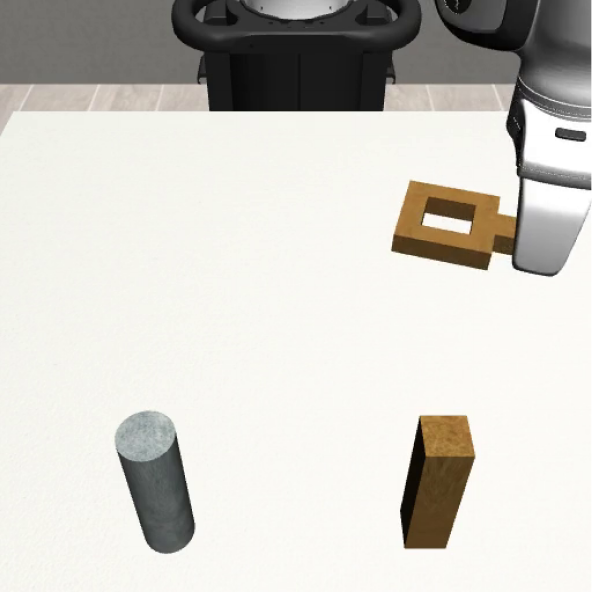}
      \includegraphics[width=0.24\textwidth]{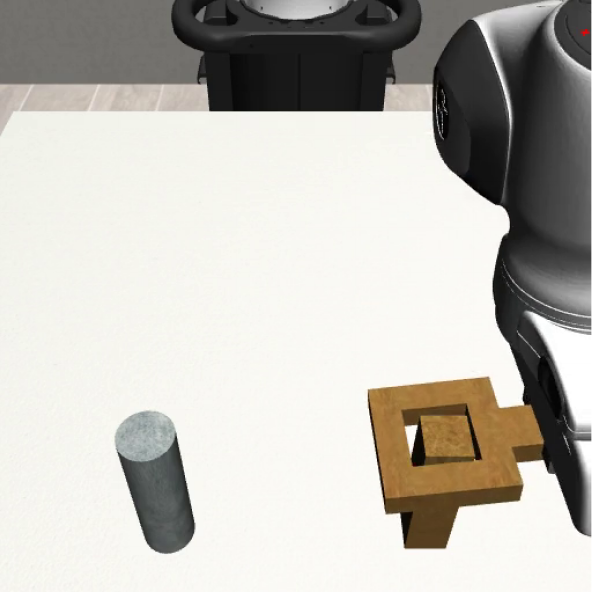}
    \caption{Snapshot of NA task rollout}
    \label{fig:task_appna}
\end{figure}

\begin{figure}[h]
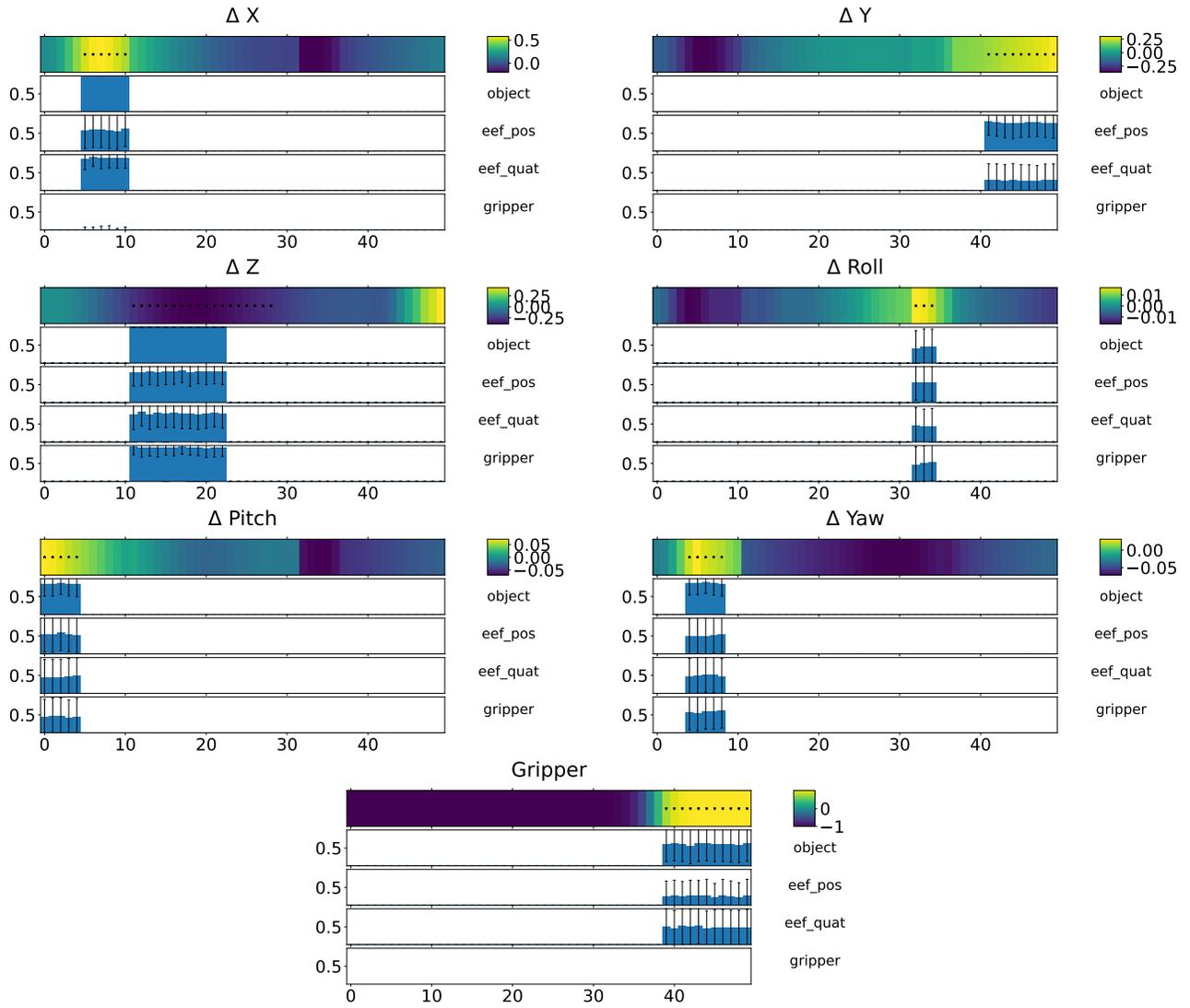

    \centering
    \includegraphics[width=0.49\textwidth]{figs/delta_XPPC_1.pdf}
    \includegraphics[width=0.49\textwidth]{figs/delta_YPPC_1.pdf}
    \includegraphics[width=0.49\textwidth]{figs/delta_ZPPC_1.pdf}
    \includegraphics[width=0.49\textwidth]{figs/delta_RollPPC_1.pdf}
    \includegraphics[width=0.49\textwidth]{figs/delta_PitchPPC_1.pdf}
    \includegraphics[width=0.49\textwidth]{figs/delta_YawPPC_1.pdf}
    \includegraphics[width=0.49\textwidth]{figs/GripperLIFT_1.pdf}
    \caption{BeTCAVe across multiple $C_A$ on task NA}
    \label{fig:app-na}
\end{figure}

\begin{figure}[h]
    \centering
    \includegraphics[width=0.24\textwidth]{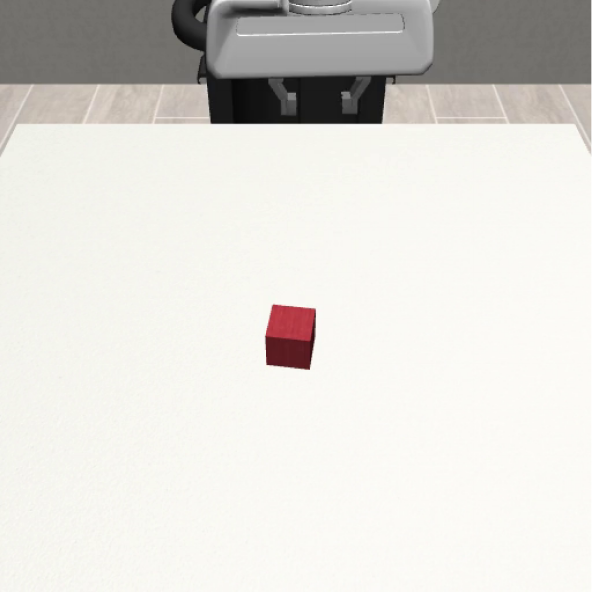}
      \includegraphics[width=0.24\textwidth]{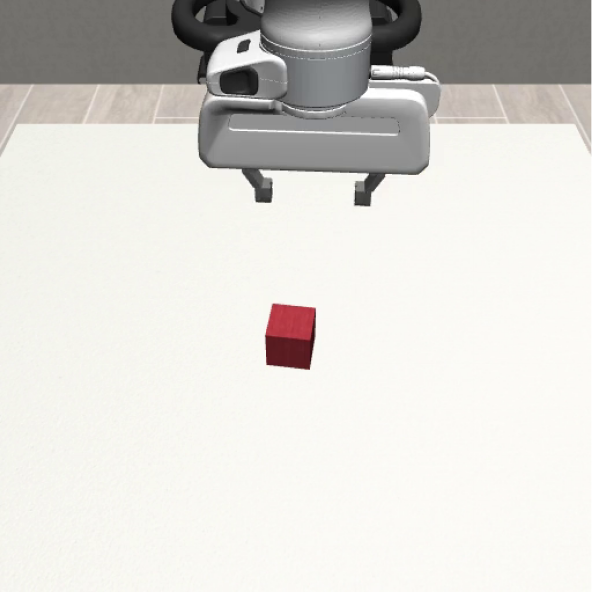}
      \includegraphics[width=0.24\textwidth]{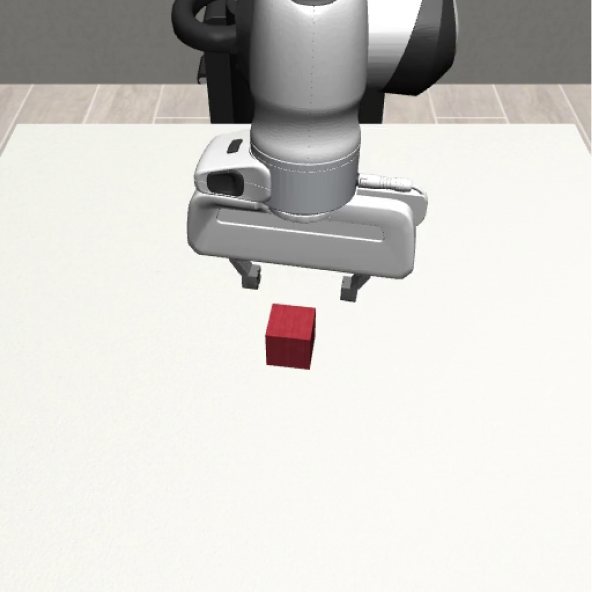}
      \includegraphics[width=0.24\textwidth]{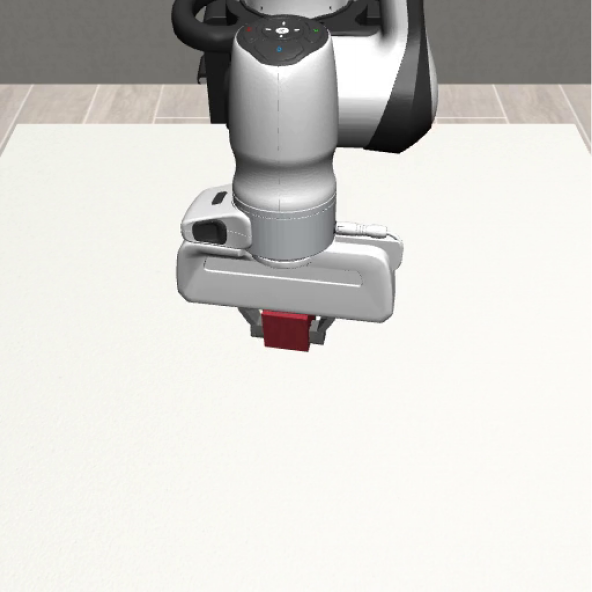}
    \caption{Snapshot of LC task rollout}
    \label{fig:task_applift}
\end{figure}

\begin{figure}[h]
    \centering
    \includegraphics[width=0.49\textwidth]{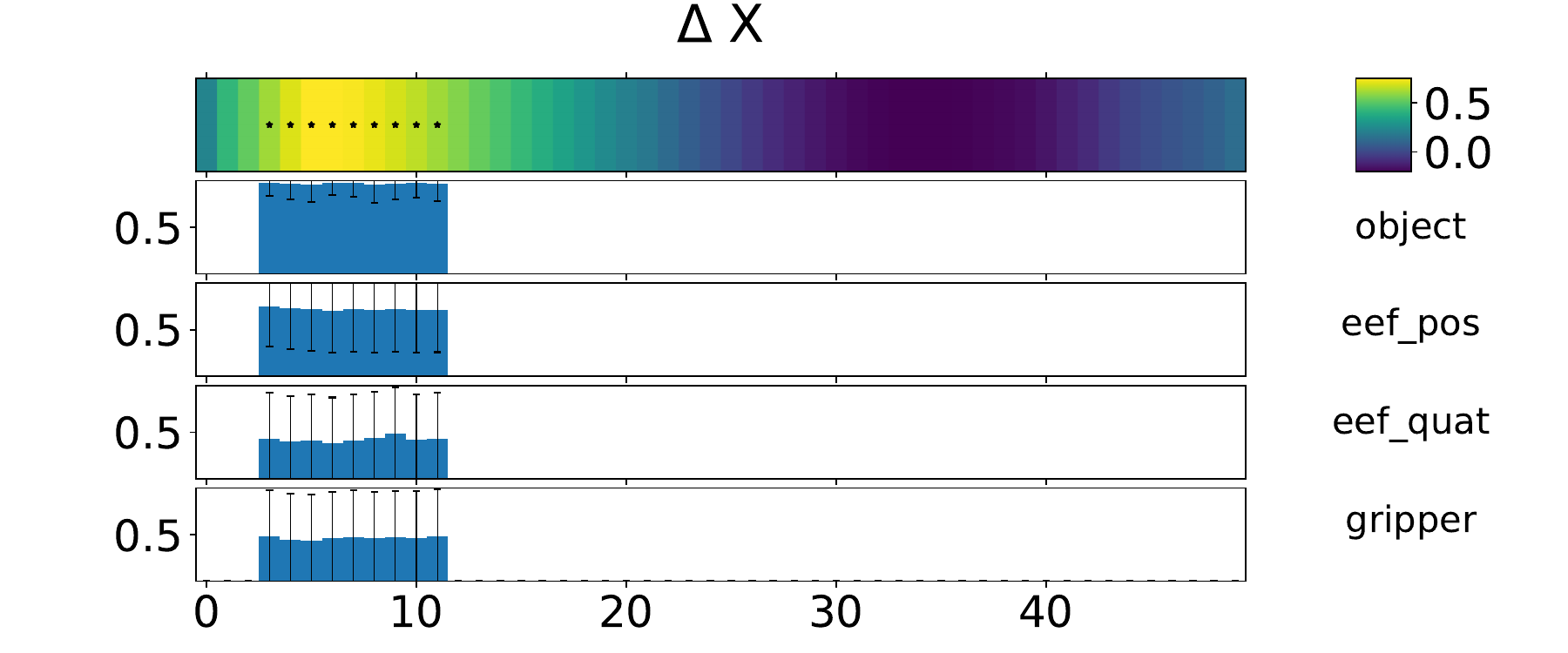}
    \includegraphics[width=0.49\textwidth]{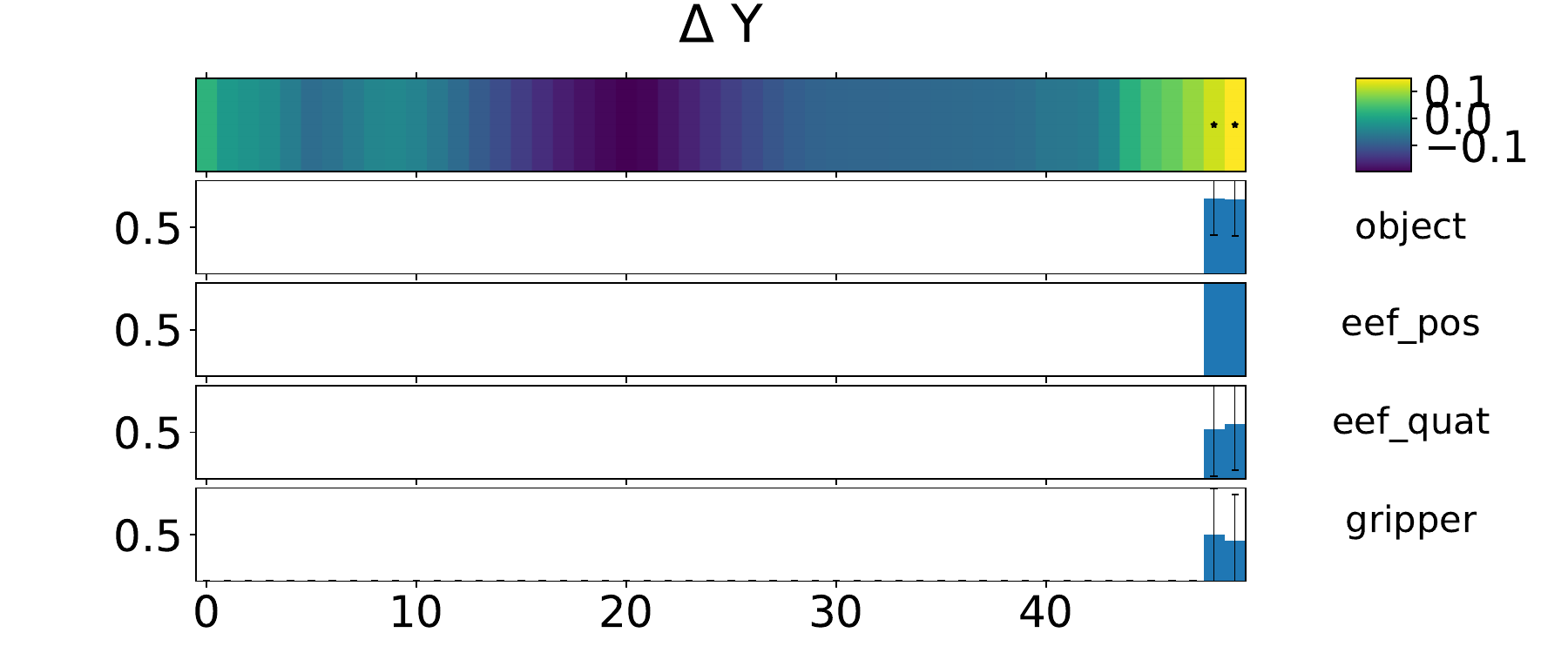}
    \includegraphics[width=0.49\textwidth]{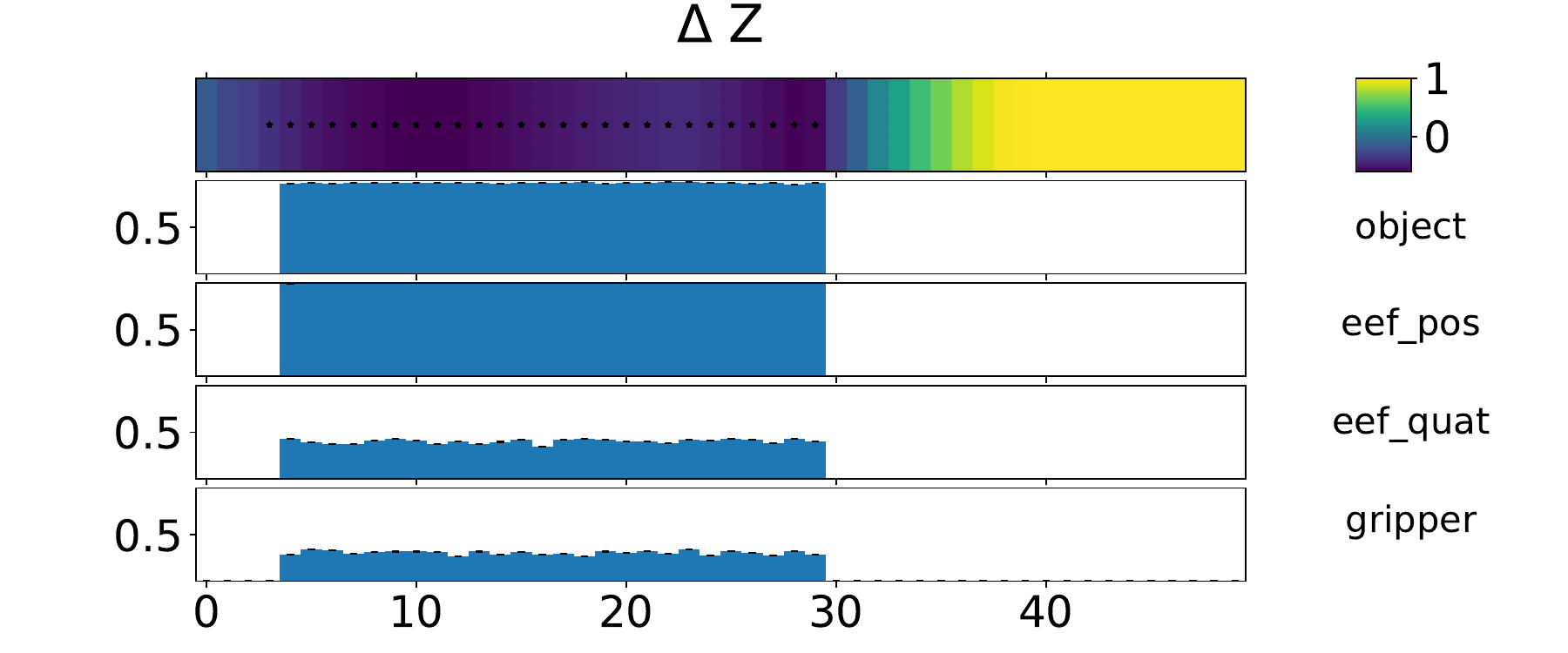}
    \includegraphics[width=0.49\textwidth]{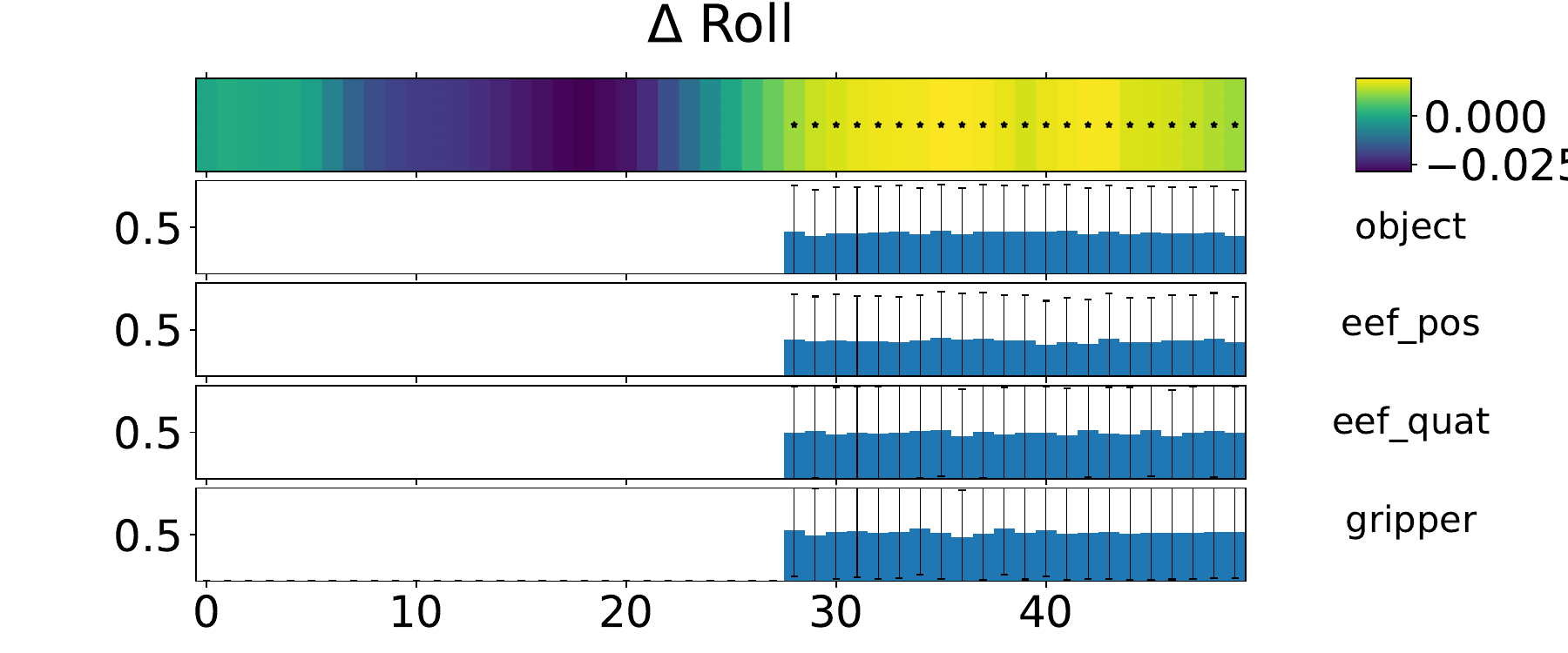}
    \includegraphics[width=0.49\textwidth]{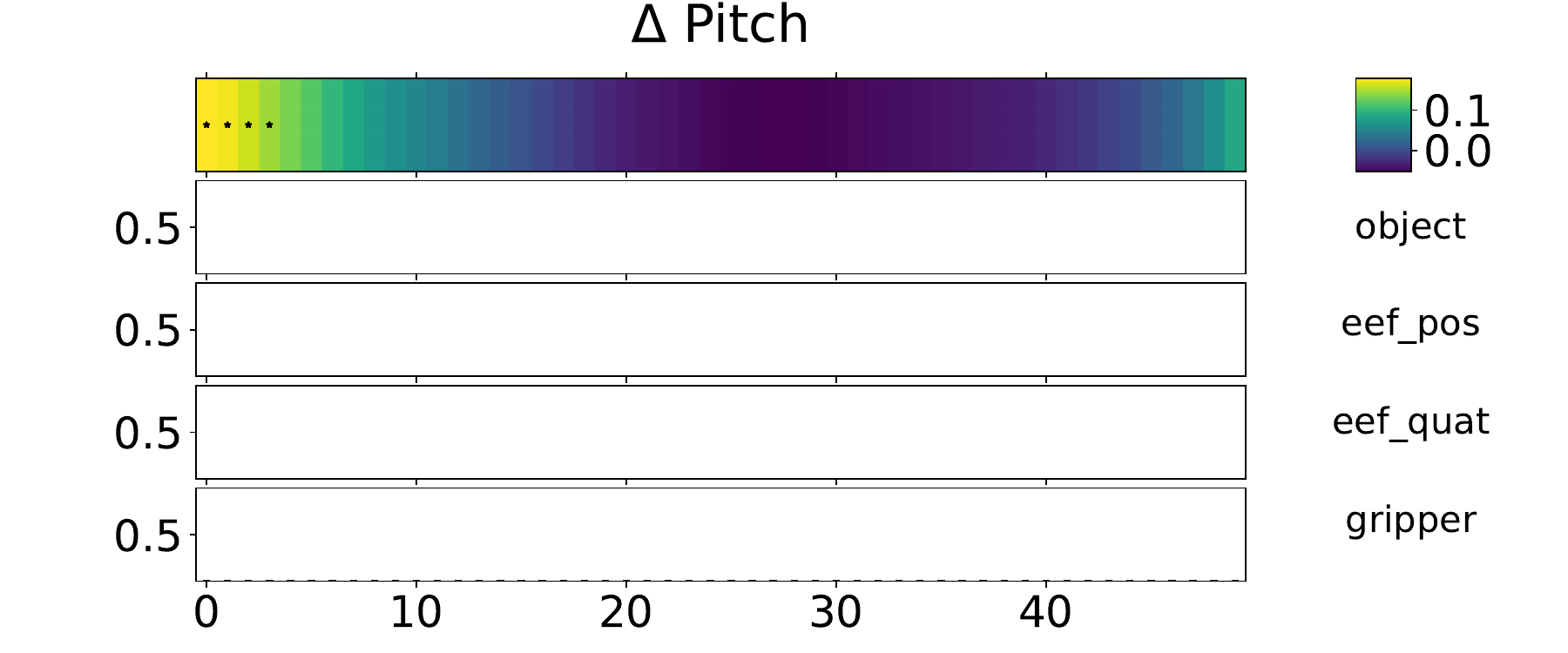}
    \includegraphics[width=0.49\textwidth]{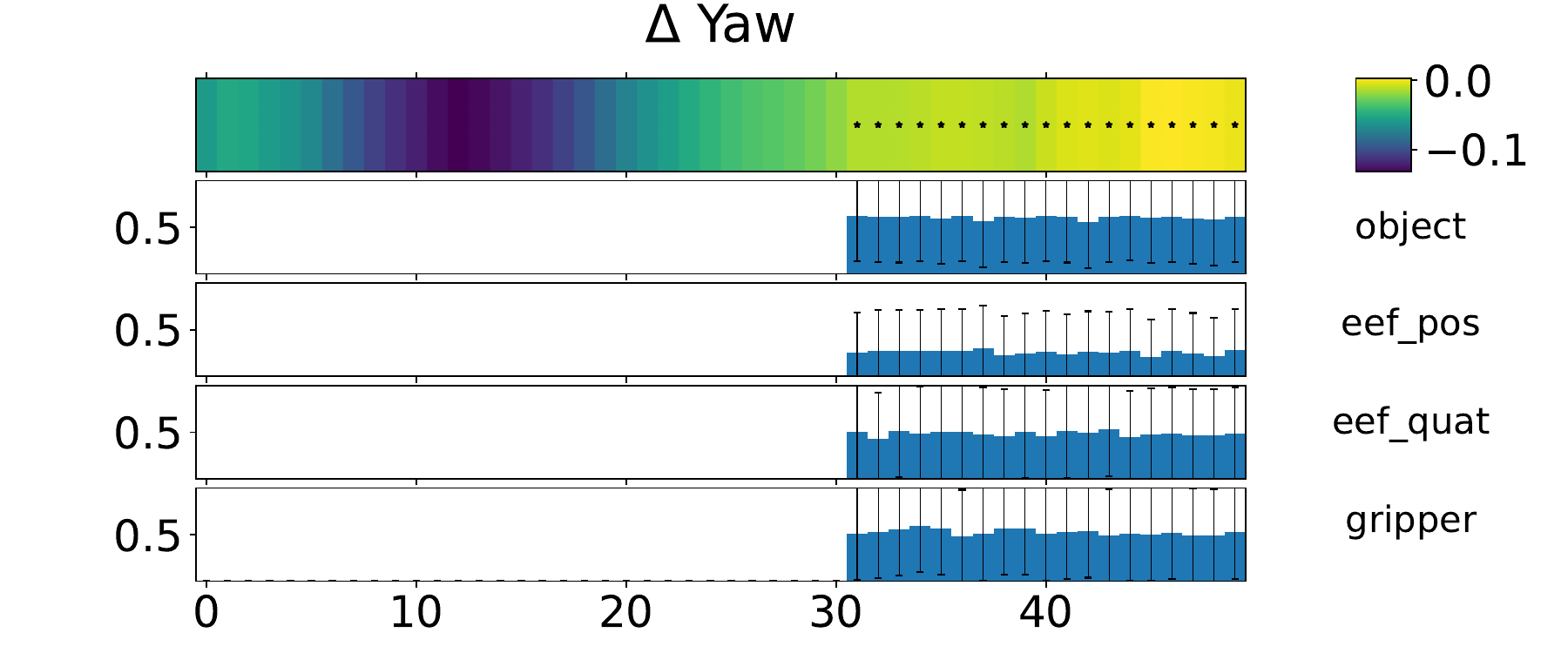}
    \includegraphics[width=0.49\textwidth]{figs/GripperLIFT_1.pdf}
    \caption{BeTCAVe across multiple $C_A$ on task LC}
    \label{fig:app-lift}
\end{figure}


\subsection{Concepts}
Due to the nature of proprioceptive sensors not using image inputs, we do not define $C_I$ as images like traditional methods. We use the inputs vectors as $C_I$ the vector input of the perticular concept would be taken from the input and rest of the dimesnion would be randomly selected for example $C_I = 'object'$ would be the vector input of object(size 10-14) + random vectors(size 9) to make up the total dimension 19. Random concept were random vectors of size 19-23.

\subsection{Additional Results}
\label{app:add_ps}

For this experiment we test BaTCAVe on the final linear layer which is just after the MLP layer in the architecture Fig~\ref{fig:Local explaination non image}(a)


\begin{table*}[ht]
    \centering
    \small
    \caption{BaTCAVe scores across different tasks, quantifying the relevance of each action concept to the task, along with uncertainty estimates. Each score measures the impact of input concepts such as the object's features, end-effector position (eef\_pos), end-effector orientation (eef\_quat in quaternion format), and gripper on task performance. (*Not Applicable)}
    \begin{tabular}{cccccc}
        \toprule
        \textbf{Task} &\textbf{Action Concept} & \textbf{object} & \textbf{eef\_pos} & \textbf{eef\_quat} & \textbf{gripper} \\
        \midrule
        \multirow{7}{*}{Pick \& place} & $\Delta$ X $(\uparrow)$ & $1 \pm 0.0$ & $0.54 \pm 0.49$ & $0.96 \pm 0.16$ & $0.09 \pm 0.09$ \\
        & $\Delta$ Y $(\uparrow)$ & $0.0 \pm 0.0$ & $0.7 \pm 0.41$ & $0.24 \pm 0.43$ & $0.0 \pm 0.0$ \\
        & $\Delta$ Z $(\uparrow)$ & $0.9 \pm 0.17$ & $0.85 \pm 0.35$ & $0.77 \pm 0.41$ & $0.91 \pm 0.27$ \\
        & $\Delta$ Roll $(\uparrow)$ & $0.44 \pm 0.49$ & $0.53 \pm 0.49$ & $0.43 \pm 0.49$ & $0.49 \pm 0.49$ \\
        & $\Delta$ Pitch $(\uparrow)$ & $0.86 \pm 0.33$ & $0.55 \pm 0.49$ & $0.45 \pm 0.49$ & $0.46 \pm 0.49$ \\
        & $\Delta$ Yaw $(\uparrow)$ & $0.84 \pm 0.35$ & $0.51 \pm 0.49$ & $0.50 \pm 0.49$ & $0.58 \pm 0.49$ \\
        & Gripper $(\uparrow)$ & $0.1 \pm 0.0$ & $0.64 \pm 0.46$ & $0.56 \pm 0.48$ & N.A* \\
        \midrule
        \multirow{7}{*}{Nut assembly} & $\Delta$ X $(\uparrow)$ & $0.64 \pm 0.47$ & $0.59 \pm 0.49$ & $0.44 \pm 0.49$ & $0.87 \pm 0.32$ \\
        & $\Delta$ Y $(\uparrow)$ & $0.87 \pm 0.32$ & $0.59 \pm 0.48$ & $0.46 \pm 0.49$ & $0.37 \pm 0.48$ \\
        & $\Delta$ Z $(\uparrow)$ & $0.005 \pm 0.07$ & $0.54 \pm 0.49$ & $0.51 \pm 0.49$ & $0.85 \pm 0.35$ \\
        & $\Delta$ Roll $(\uparrow)$ & $0.52 \pm 0.49$ & $0.48 \pm 0.49$ & $0.48 \pm 0.49$ & $0.45 \pm 0.49$ \\
        & $\Delta$ Pitch $(\uparrow)$ & $0.66 \pm 0.46$ & $0.51 \pm 0.49$ & $0.45 \pm 0.49$ & $0.56 \pm 0.49$ \\
        & $\Delta$ Yaw $(\uparrow)$ & $0.58 \pm 0.49$ & $0.44 \pm 0.49$ & $0.45 \pm 0.49$ & $0.52 \pm 0.49$ \\
        & Gripper $(\uparrow)$ & $0.1 \pm 0.0$ & $0.88 \pm 0.31$ & $0.53 \pm 0.47$ & N.A* \\
        \midrule
        \multirow{7}{*}{Lift} & $\Delta$ X $(\uparrow)$ & $0.84 \pm 0.31$ & $0.42 \pm 0.40$ & $0.42 \pm 0.49$ & $0.42 \pm 0.49$ \\
        & $\Delta$ Y $(\uparrow)$ & $0.85 \pm 0.32$ & $0.99 \pm 0.07$ & $0.56 \pm 0.49$ & $0.49 \pm 0.40$ \\
        & $\Delta$ Z $(\uparrow)$ & $0.98 \pm 0.10$ & $0.99 \pm 0.04$ & $0.42 \pm 0.49$ & $0.39 \pm 0.48$ \\
        & $\Delta$ Roll $(\uparrow)$ & $0.46 \pm 0.49$ & $0.41 \pm 0.49$ & $0.48 \pm 0.48$ & $0.51 \pm 0.49$ \\
        & $\Delta$ Pitch $(\uparrow)$ & $0.54 \pm 0.49$ & $0.28 \pm 0.73$ & $0.43 \pm 0.47$ & $0.48 \pm 0.49$ \\
        & $\Delta$ Yaw $(\uparrow)$ & $0.73 \pm 0.39$ & $0.28 \pm 0.45$ & $0.47 \pm 0.49$ & $0.48 \pm 0.49$ \\
        & Gripper $(\uparrow)$ & $0.95 \pm 0.18$ & $0.004 \pm 0.06$ & $0.31 \pm 0.39$ & N.A* \\
        \bottomrule
    \end{tabular}
    \label{table:com_val_2}
\end{table*}

\section{Experiment 3: Cube Lifting with Vision-Language Inputs}
\label{app:exp3}

\subsection{Concepts}
The model used in experiment 3 takes in both image and language input, We keep one input constant to test concepts of the other input. Random concepts were samples from ImageNet.
\begin{enumerate}
    \item Image concept: We choose concepts from the input image we blur out the rest of the concept in the input to represent a concept as shown in Fig~\ref{fig:img_concept}. BaTCAVe shows high variance but consistent score across all $C_A$ when $C_I = 'Image'$.
    \item Language concept: We use proper instructions, gibberish instructions and just verbs as language concepts. Table~\ref{tab:lang_values} shows the chosen 3 concepts for the eperiment. Fig~\ref{fig:all_language_concept} shows BaTCAVe tested across all $C_A$ in LC task with language used as $C_I$.
\end{enumerate}

\begin{figure}[h]
    \centering
    \includegraphics[width=0.23\textwidth]{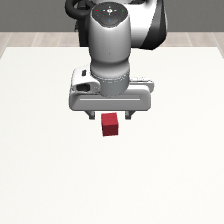}
    \includegraphics[width=0.23\textwidth]{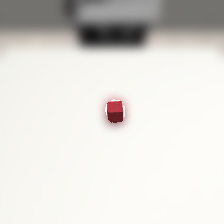}
    \includegraphics[width=0.23\textwidth]{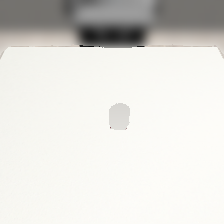}
    \includegraphics[width=0.23\textwidth]{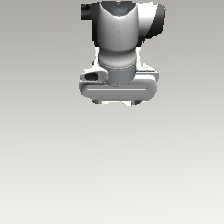}
    \caption{a) Input b) Red c) Table d) End effector}
    \label{fig:img_concept}
\end{figure}

\begin{table}[h!]
\centering
\caption{Language concepts}
\label{tab:lang_values}
\begin{tabular}{clll}
\toprule
\textbf{Index} & \textbf{Standard Instructions} & \textbf{Gibberish} & \textbf{Verb} \\ \midrule
1 & Lift the box    & skdfj 12asj 5893 2467* & Lift     \\
2 & Grab the box    & fjdkl c33kd 3940 8175  & Grab     \\
3 & Take the box    & qpwie b99fs 1295 375476 & Take     \\
4 & Move the box    & zxcvb n66gh 5421 983613 & Move     \\
5 & Collect the box & plmok u55wr 7864 2319 & Collect  \\
6 & Retrieve the box& akyse l44qs 6572 048756 & Retrieve \\
7 & Hoist the box   & bvgfr t22vp 3187 7695*( & Hoist    \\
8 & Handle the box  & nmjqw o11lm 9538 65 & Handle   \\
9 & Carry the box   & xswed k88ht 2409 186428 & Carry    \\
10& Raise the box   & ecrvt f77yr 4812 690391 & Raise    \\ \bottomrule
\end{tabular}
\end{table}
\begin{figure}[h]
    \centering
    \includegraphics[width=0.49\textwidth]{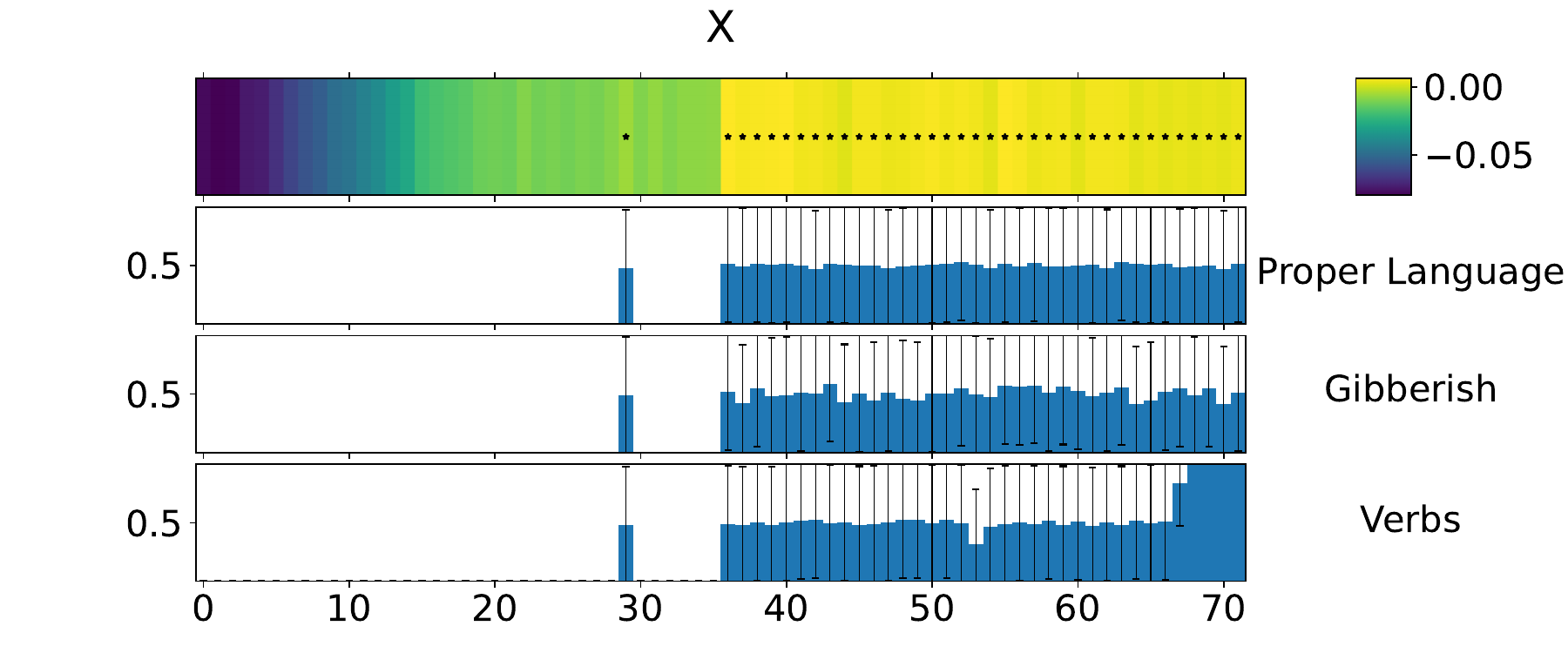}
    \includegraphics[width=0.49\textwidth]{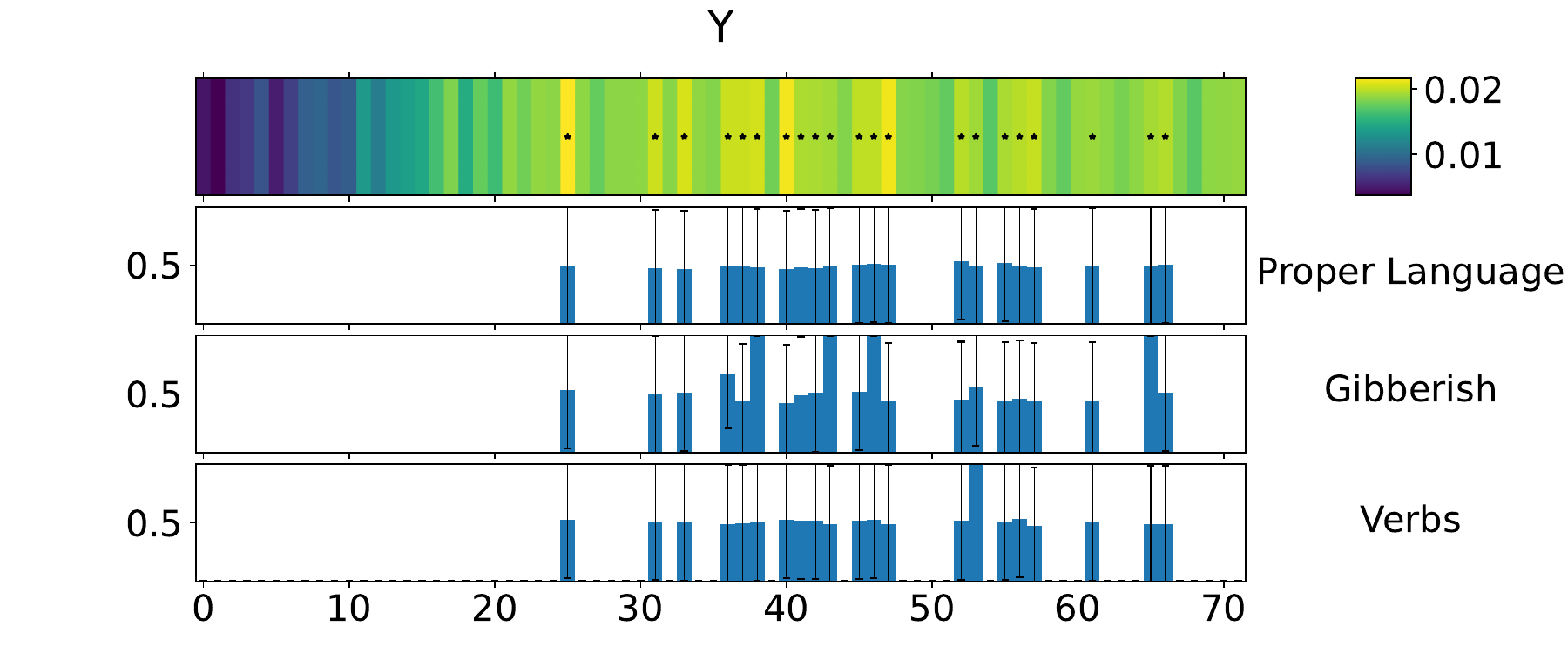}
    \includegraphics[width=0.49\textwidth]{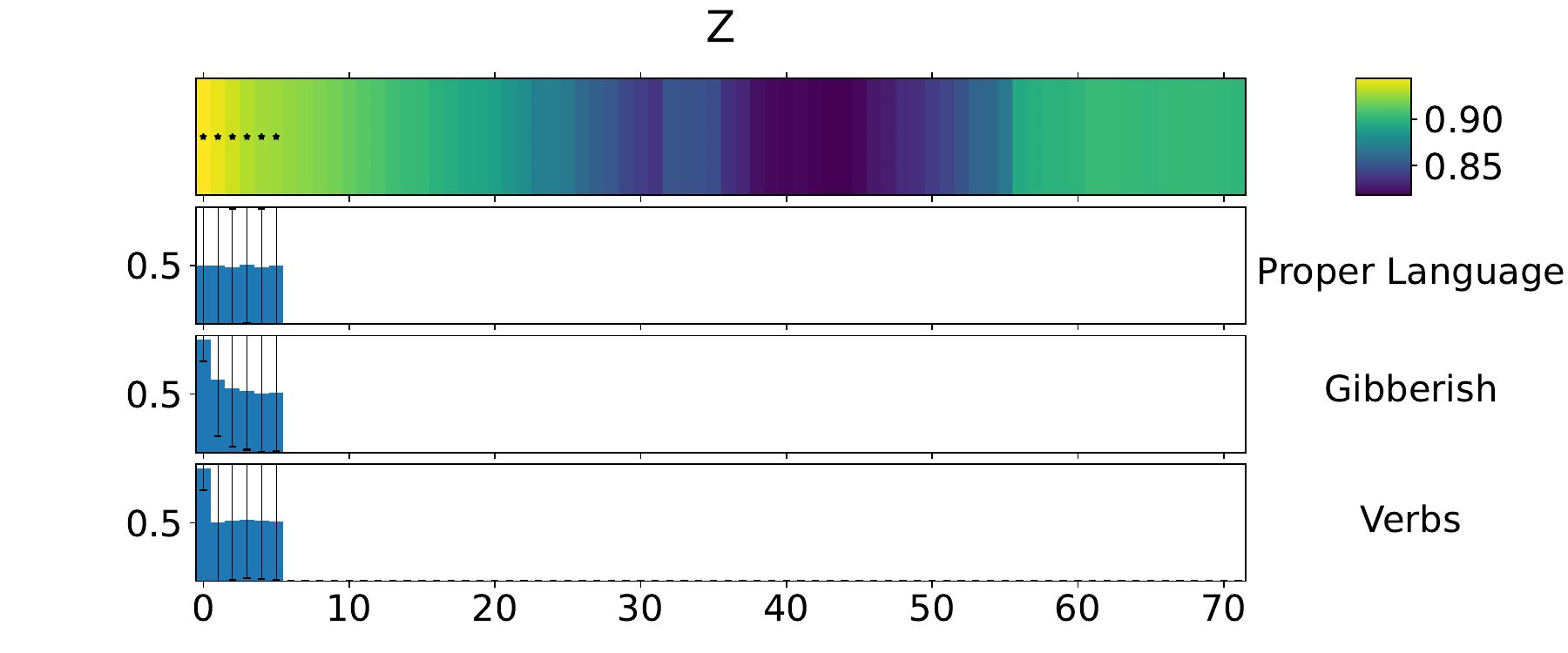}
    \includegraphics[width=0.49\textwidth]{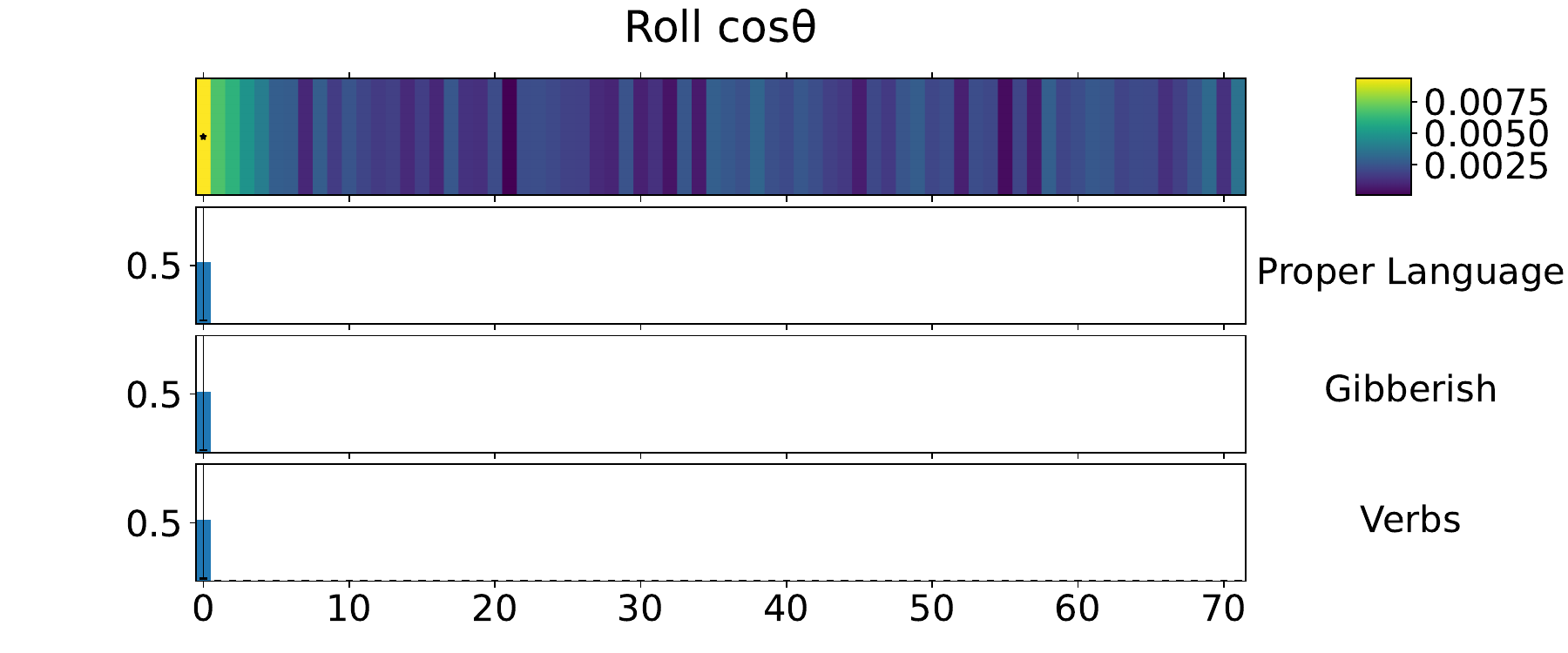}
    \includegraphics[width=0.49\textwidth]{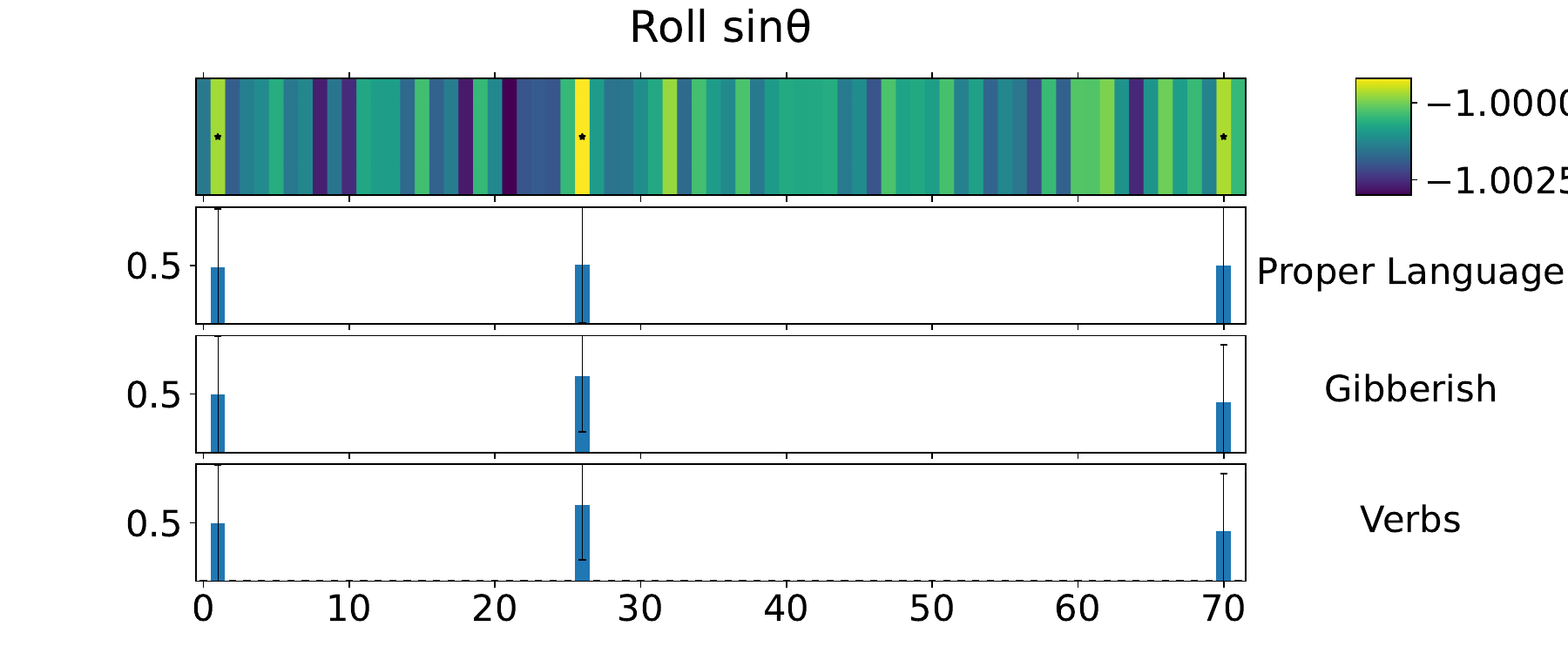}
    \includegraphics[width=0.49\textwidth]{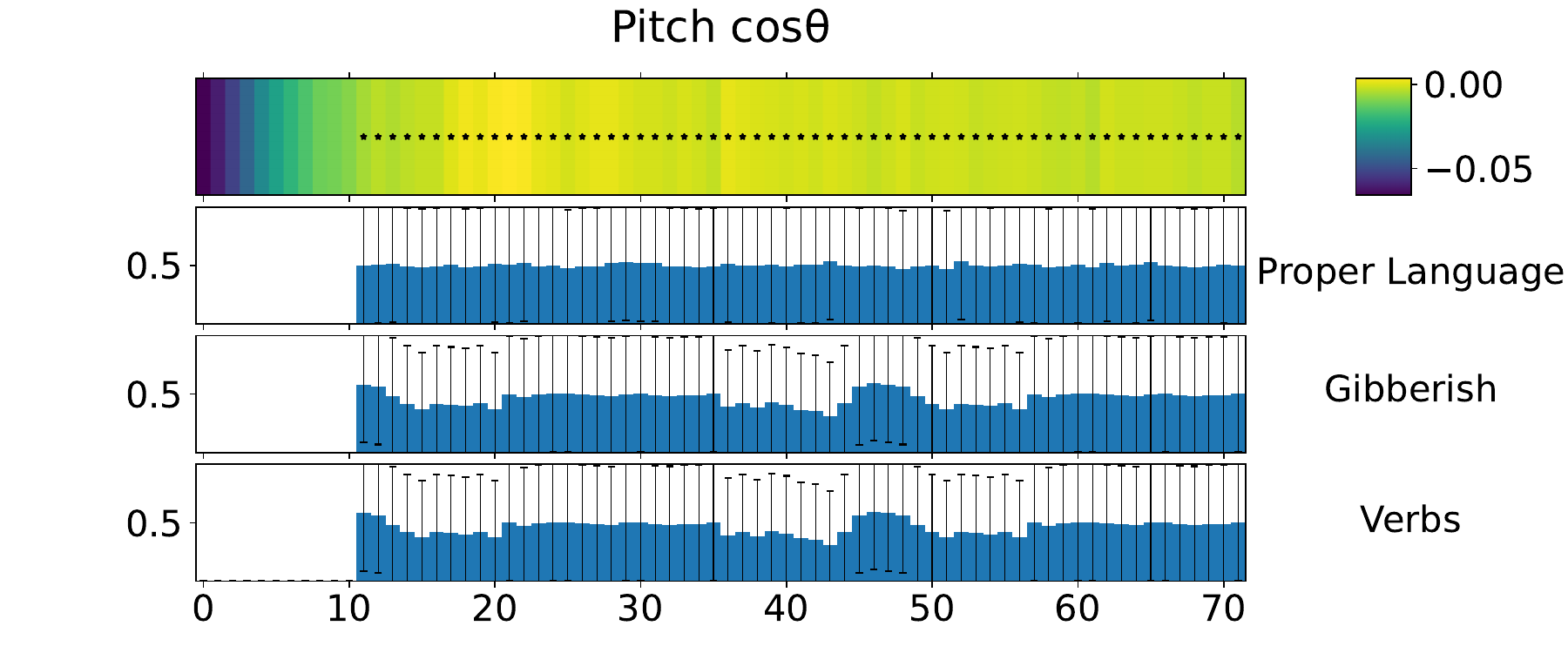}
    \includegraphics[width=0.49\textwidth]{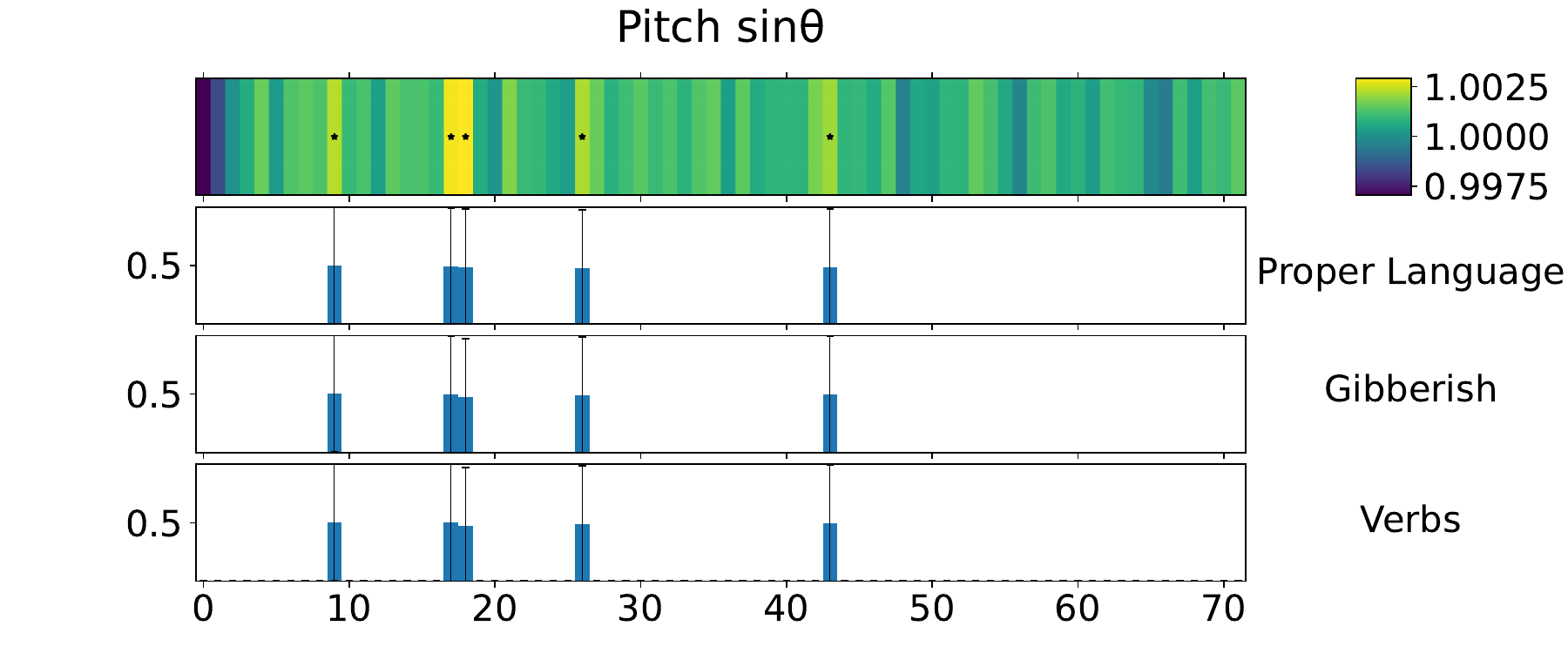}
    \includegraphics[width=0.49\textwidth]{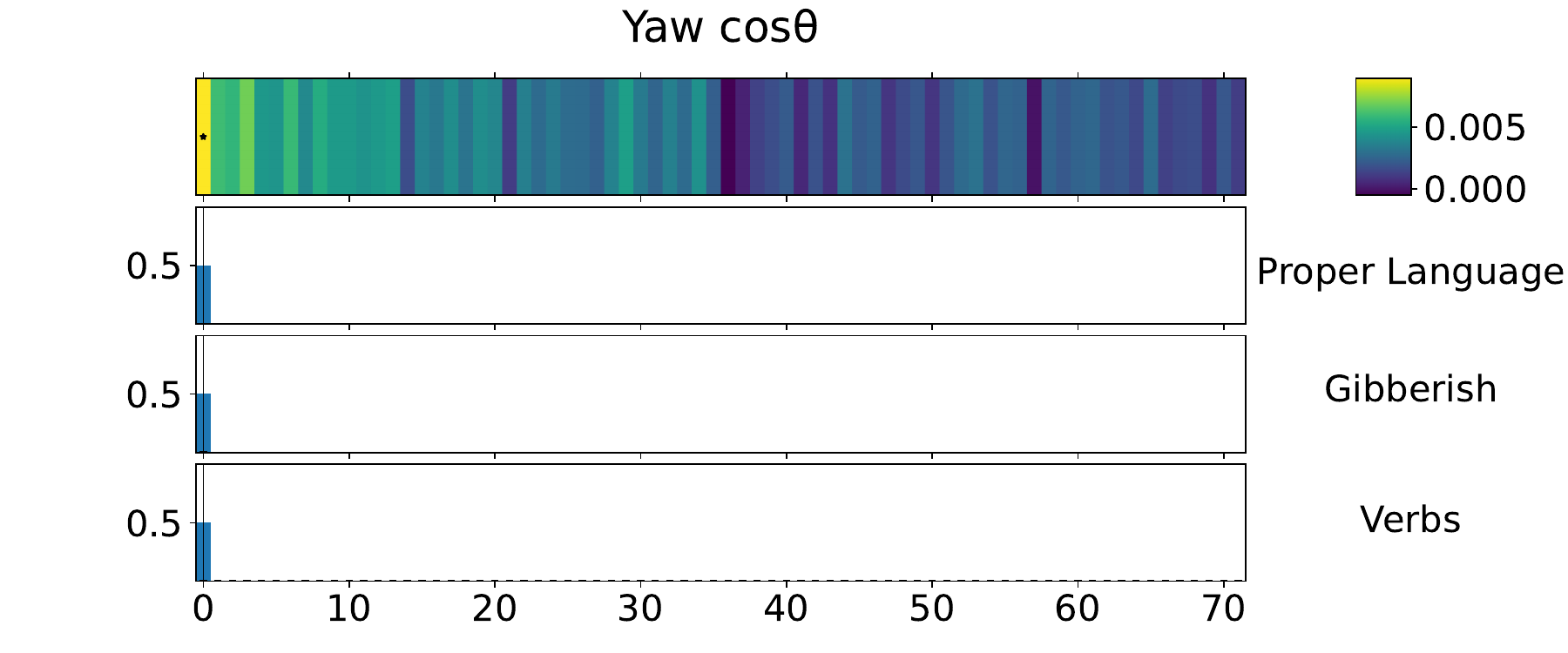}
    \includegraphics[width=0.49\textwidth]{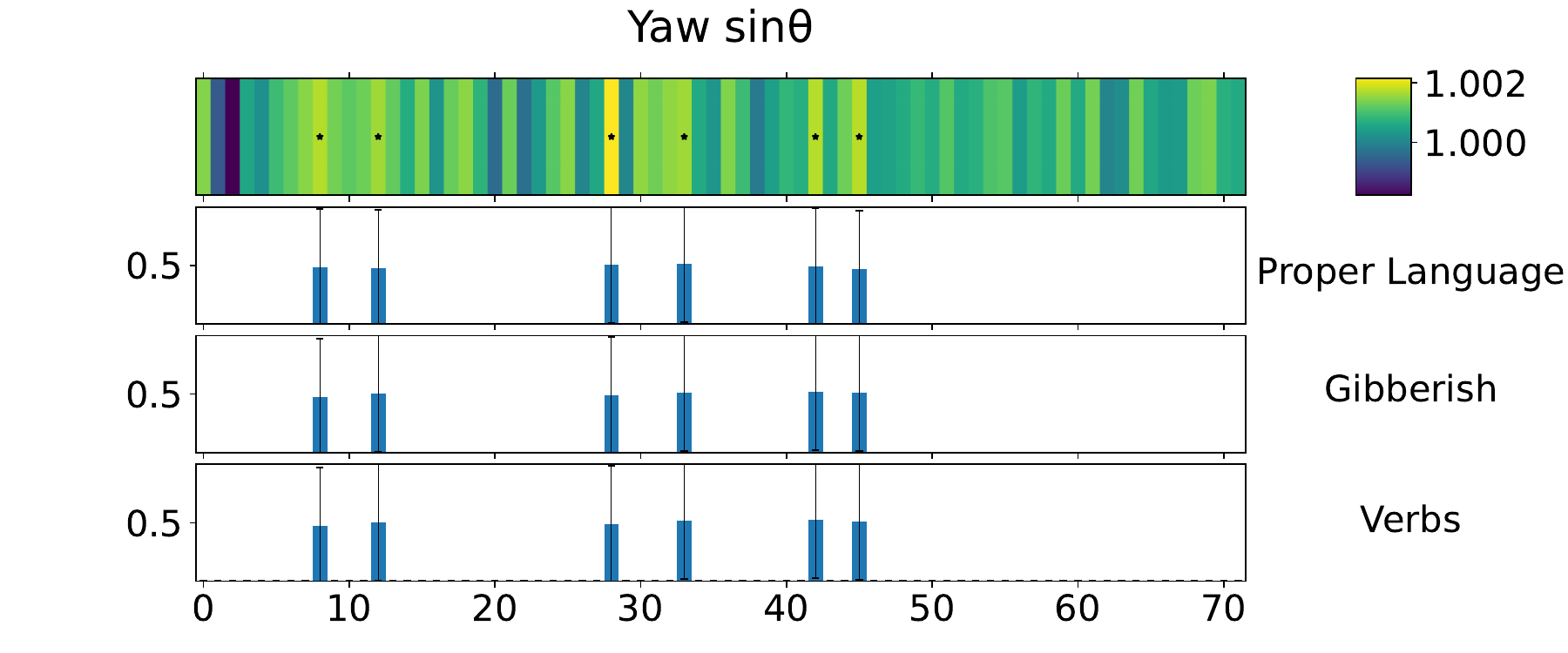}
    \includegraphics[width=0.49\textwidth]{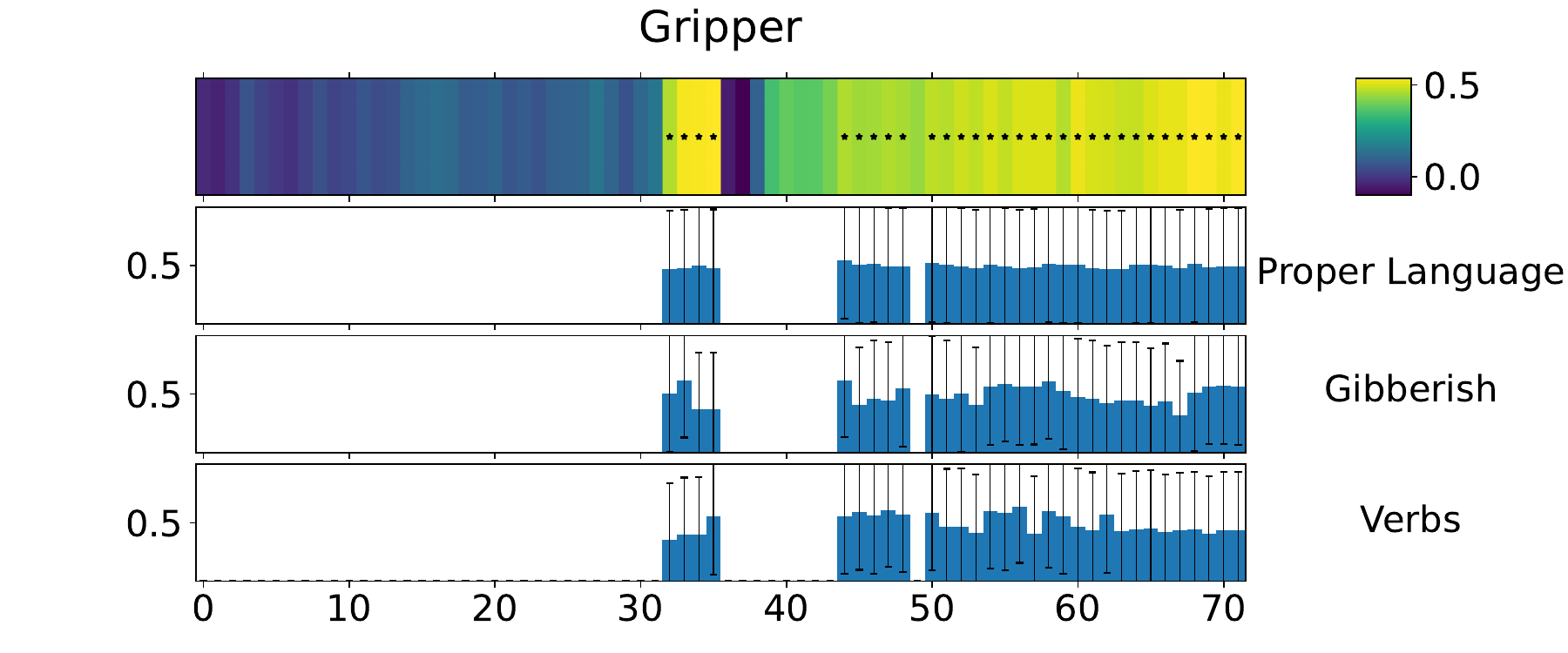}
    \caption{BeTCAVe across multiple $C_A$ on task LC with language concept }
    \label{fig:all_language_concept}
\end{figure}



\subsection{Model details}
\label{app:exp3_MT}

\textbf{Architecture}: The network outputs a 360-dimensional action vector via a final linear layer, mapping the robot's projected movements over the next 36 timesteps. This output includes the first 108 nodes for end-effector position (x, y, z), the subsequent 108 for orientation (roll, pitch, yaw), and the final 36 for gripper.

The main components are Visual Encoder, Visual Narrower, Task ID Encoder, Controllers

\begin{enumerate}
    \item Visual Encoder: This part of the model uses a ResNet (Residual Network) architecture for visual encoding.
    \item Visual Narrower: A linear transformation that reduces the feature dimension from the output of the ResNet (512 features) down to a lower-dimensional space (256 features).
    \item Task ID Encoder(CLIP):  Vision transformer for encoding task IDs based on both textual 
    \item Controllers
    \begin{enumerate}
        \item XYZ Controller: Controls the position in 3D space.
        \item RPY Controller: Controls rotation around the roll, pitch, and yaw axes.
        \item Grip Controller: Manages the actions related to opening and closing a robotic gripper.
    \end{enumerate}   
\end{enumerate}

\textbf{Training}: The Model is trained based on a behavior cloning (BC) policy tailored to handle dual inputs: high-resolution images (224x224x3) from the camera and verbal instructions given to the robot. We train the model for 100,000 epochs on 300 demonstrations with a batch size of 64 using a Huber loss function and Adam optimizer.

\section{Experiment 4: Vision-based Autonomous Driving}
\label{app:exp4}

\subsection{Model details}
\textbf{Architecture}: The policy model architecture comprises a sequence of convolutional layers followed by fully connected layers.  The input to the model is an RGB image, representing the current state of the environment. After the convolutional layers, a flattening operation is applied, transforming the 3D feature maps into a 1D feature vector. This vector serves as the input to the fully connected layers. The flattened vector is fed into a dense (fully connected) layer comprising 512 units. This layer integrates the features extracted by the convolutional layers to form a high-level representation of the input.

\textbf{Training}: The policy network was trained on PPO algorithm using Stable Baselines3 library. The parameters set for training included a learning rate of 0.0003, a rollout of 2048 steps per update, and a batch size of 64. Discount factor (gamma) was set at 0.99 with a Generalized Advantage Estimation (GAE) lambda of 0.95, which helps in balancing bias and variance. The policy clipping range was set to 0.2, and advantages were normalized to stabilize the training. Additionally, the value function coefficient was set at 0.5, and the maximum gradient norm was capped at 0.5 to prevent exploding gradients. 

\subsection{Concepts}

\begin{figure}[h]
    \centering
    \includegraphics[width=0.9\textwidth]{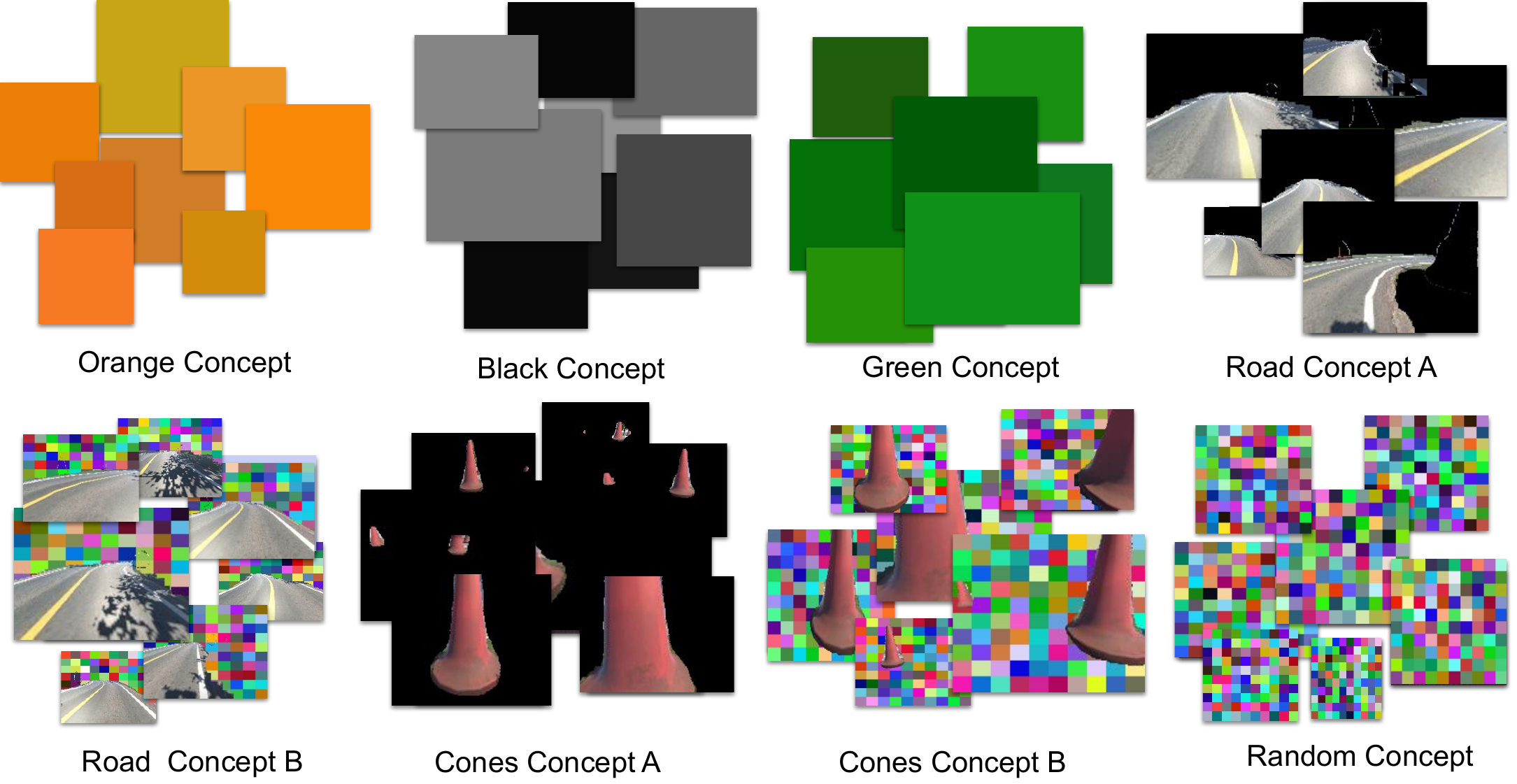}
    \caption{Different concepts used in BaTCAVe for vision-based autonomous driving task.}
    \label{fig:concept-donkeygym}
\end{figure}

One of the limitations are when the concepts are small, they might not provide a strong enough signal to distinguish from a random concept class. For instance, Cone Concept B and Random Concept in Fig.~\ref{fig:concept-donkeygym} are largely similar. 


\end{document}